\documentclass[letterpaper, oneside]{memoir}
\setulmarginsandblock{3cm}{3cm}{*}
\setlrmarginsandblock{3cm}{3cm}{*}
\checkandfixthelayout

\usepackage[numbers]{natbib}
\usepackage{graphicx}
\usepackage{amsmath}
\usepackage{xcolor}

\usepackage{hyperref}
\usepackage{tikz}
\usepackage{fontawesome5}
\usetikzlibrary{positioning}

\makeatletter
\newcommand{\chapterauthor}[1]{%
  {\parindent0pt\vspace*{-25pt}%
  \linespread{1.1}\large\scshape#1%
  \par\nobreak\vspace*{35pt}}
  \@afterheading%
}
\makeatother

\title{A Comprehensive Survey and Guide to Multimodal Large Language Models in Vision-Language Tasks}

\author{
    Chia Xin Liang \\ \texttt{cxldun@gmail.com} \\ JTB Technology Corp. \\
    \and
    Pu Tian \\ \texttt{pu.tian@stockton.edu} \\ Stockton University \\
    \and
    Caitlyn Heqi Yin \\ \texttt{hyin66@wisc.edu} \\ University of Wisconsin-Madison \\
    \and
    Yao Yua \\ \texttt{yaoyuanou@gmail.com} \\ AppCubic USA \\
    \and
    Wei An-Hou \\ \texttt{weian312@gmail.com} \\ Nomad Sustaintech LTD \\
    \and
    Li Ming \\ \texttt{mli694@gatech.edu} \\ Georgia Institute of Technology \\
    \and
    Xinyuan Song \\ \texttt{xsong69@emory.edu} \\ Emory University \\
    \and
    Tianyang Wang \\ \texttt{tianyangwang0305@gmail.com} \\ University of Liverpool \\
    \and
    Ziqian Bi \\
    \texttt{bizi@iu.edu} \\ Indiana University \\
    \and
    Ming Liu \\
        \texttt{mliu3083@purdue.edu} \\ Purdue University 
}

\begin{document}

\date{}
\maketitle

\begin{abstract}
    This survey and application guide to multimodal large language models(MLLMs) explores the rapidly developing field of MLLMs, examining their architectures, applications, and impact on AI and Generative Models. Starting with foundational concepts, we delve into how MLLMs integrate various data types, including text, images, video and audio, to enable complex AI systems for cross-modal understanding and generation. It covers essential topics such as training methods, architectural components, and practical applications in various fields, from visual storytelling to enhanced accessibility. Through detailed case studies and technical analysis, the text examines prominent MLLM implementations while addressing key challenges in scalability, robustness, and cross-modal learning. Concluding with a discussion of ethical considerations, responsible AI development, and future directions, this authoritative resource provides both theoretical frameworks and practical insights. It offers a balanced perspective on the opportunities and challenges in the development and deployment of MLLMs, and is highly valuable for researchers, practitioners, and students interested in the intersection of natural language processing and computer vision.
\end{abstract}
\newpage

\tableofcontents
\setsecnumdepth{subsection}
\settocdepth{subsection}

\chapter{Introduction to Multimodal Large Language Models (MLLMs)}

\section{Definition and Importance of MLLMs}

Multimodal Large Language Models (MLLMs) represent a significant evolution in artificial intelligence (AI), enabling the integration and understanding of various input types such as text, images, audio, and video. Unlike unimodal models restricted to a single input type, MLLMs process multiple modalities simultaneously, providing a more comprehensive understanding that reflects real-world interactions.

The key features and importance of MLLMs include:

\textbf{Cross-Modal Learning:} MLLMs are trained on extensive datasets encompassing textual, visual, auditory, and sometimes sensory data. This capability allows them to create connections between different modalities, enabling tasks that require comprehension and generation of content across diverse data types. For example:

\begin{itemize}
    \item \textbf{Text-to-Image Generation:} MLLMs can generate detailed images from textual descriptions, revolutionizing creative industries like graphic design and advertising. Imagine describing a "futuristic cityscape at sunset" and having an AI generate a corresponding image.

    \item \textbf{Visual Question Answering:} These models can analyze images and provide accurate answers to natural language questions, enhancing educational tools and accessibility technologies. For instance, an MLLM could answer questions about the contents of a photograph, such as "What breed of dog is in this image?"
    \item \textbf{Multimodal Content Creation:} MLLMs facilitate the creation of content that integrates text, visuals, and audio, such as illustrated stories or multimedia presentations. This could involve generating a coherent story with matching illustrations based on a brief prompt.
\end{itemize}

\textbf{Unified Representation:} MLLMs achieve integrated representations of multimodal data through unified codebooks and joint embedding spaces, enabling seamless processing across different modalities. This architectural approach offers several key capabilities:

\begin{itemize}
    \item Seamless translation between modalities (e.g., describing a photograph or generating an image from text).
    \item Cross-modal retrieval, where the model can find relevant images based on text queries or match sounds with visual content.
    \item More natural and intuitive interactions between humans and AI systems.
\end{itemize}

To understand unified representation, imagine a library where books, images, and audio recordings are all cataloged using the same system, allowing you to easily find related items across different media types.

\textbf{Enhanced Contextual Understanding:} By integrating multiple modalities, MLLMs generate more accurate and context-aware responses. This capability is particularly valuable in fields such as:

\begin{itemize}
    \item Healthcare: Analyzing medical images alongside patient records and physician notes for more precise diagnoses. For example, an MLLM could combine a patient's X-ray, medical history, and symptoms to suggest potential diagnoses.
    \item Security: Interpreting surveillance footage in conjunction with audio data for comprehensive situational awareness. This could involve analyzing video feeds and audio recordings to detect potential security threats.
    \item E-commerce: Enhancing product searches by understanding both textual queries and visual product attributes. An MLLM could help a customer find a "blue floral summer dress" by understanding both the text description and visual characteristics of available products.
\end{itemize}

\textbf{Generalization Across Modalities:} MLLMs demonstrate flexibility in handling various tasks across different modalities, including:

\begin{itemize}
    \item Image captioning and visual question answering.
    \item Cross-modal retrieval and content generation.
    \item Audio-visual integration for tasks like video subtitling or lip-syncing.
    \item Multimodal translation, such as converting a video into a textual summary.
    \item Enhanced human-computer interaction through simultaneous interpretation of gestures, facial expressions, speech, and text.
\end{itemize}

\textbf{Advancements in Robotics and Embodied AI:} In robotics, MLLMs contribute to systems that can perceive and interact with their environment more effectively. By processing visual, auditory, and sensory data, robots powered by MLLMs can perform complex tasks such as object manipulation, navigation, and human-robot interaction. For instance, a household robot could understand and execute a verbal command like "Please bring me the red mug from the kitchen counter," by combining language understanding with visual recognition and spatial navigation.

\textbf{Real-World Application Potential:} The ability of MLLMs to process diverse data types makes them valuable for real-world applications where information comes in various forms. For instance:

\begin{itemize}
    \item In autonomous vehicles, these models can integrate visual data from cameras with textual information from maps and traffic reports, enhancing navigation and safety features. An MLLM could help a self-driving car understand a road sign, interpret its meaning, and adjust the vehicle's behavior accordingly.
    \item In scientific research, MLLMs can analyze molecular structures, research papers, and experimental data simultaneously to identify potential new compounds for drug discovery. This could accelerate the process of finding new treatments by identifying patterns across diverse datasets that human researchers might miss.
\end{itemize}

\textbf{Bridging the Gap Between AI and Human Cognition:} MLLMs' ability to process multiple modalities mirrors human cognitive processes more closely than unimodal models. This alignment with human cognition can lead to AI systems that are more intuitive to use and better at understanding complex, context-dependent situations. For example, an MLLM-powered virtual assistant could understand and respond to a user's mood based on their tone of voice, facial expression, and choice of words, much like a human would.

\section{The Convergence of Natural Language Processing (NLP) and Computer Vision: The Emergence of MLLMs}

The fusion of natural language processing (NLP) and computer vision has been a game-changer in AI, giving rise to MLLMs. This convergence allows machines to reason across different modalities, offering a more comprehensive understanding of the world.

\textbf{Key Historical Milestones:}
\begin{itemize}
    \item \textbf{Image Captioning (2015-Present):} Early models like Show, Attend, and Tell combined Convolutional Neural Networks (CNNs) for image analysis with Recurrent Neural Networks (RNNs) for text generation. This marked the beginning of machines being able to "describe" what they "see".
    \item \textbf{Visual Question Answering (VQA):} These tasks required models to combine visual and textual inputs to generate meaningful answers. For example, a model might be asked, "What color is the car?" while being shown an image of a red car.
    \item \textbf{Vision-Language Transformers (2019-Present):} Models like ViLBERT, CLIP, and DALL-E demonstrated that transformer architectures could be extended to multimodal applications. These models can perform tasks like generating images from text descriptions or finding the most relevant image for a given text query.
\end{itemize}

\begin{center}
\begin{tikzpicture}[scale=1, transform shape]
    \draw[thick] (0,0) -- (14,0);

    \node[fill=blue!20, rounded corners, text width=3cm, align=center, above=1cm] at (2,0) {\textbf{Image Captioning}\\(2015-Present)};
    \node[fill=gray!10, text width=4.5cm, rounded corners, below=1.5cm] at (2,0) {\faCameraRetro\quad CNNs and RNNs allow models to ``describe" images};
    \draw[thick] (2, -0.3) -- (2, 0.3); 
    \node[below=0.7cm] at (2, 0) {2015};

    \node[fill=green!20, rounded corners, text width=3cm, align=center, above=1cm] at (7,0) {\textbf{Visual Question Answering}};
    \node[fill=gray!10, text width=4.5cm, rounded corners, below=1.5cm] at (7,0) {\faQuestion\quad Models answer questions like ``What color is the car?"};
    \draw[thick] (7, -0.3) -- (7, 0.3); 
    \node[below=0.7cm] at (7, 0) {2016 - 2018};

    \node[fill=orange!20, rounded corners, text width=3.5cm, align=center, above=1cm] at (12,0) {\textbf{Vision-Language Transformers}\\(2019-Present)};
    \node[fill=gray!10, text width=4.5cm, rounded corners, below=1.5cm] at (12,0) {\faRobot\quad Transformers enable multimodal tasks like text-to-image generation};
    \draw[thick] (12, -0.3) -- (12, 0.3); 
    \node[below=0.7cm] at (12, 0) {2019};

\end{tikzpicture}
\end{center}

\textbf{Theoretical Foundations:}
The convergence of NLP and computer vision is built on several key theoretical foundations:

\begin{itemize}
    \item \textbf{Representation Learning:} This allows MLLMs to create joint embeddings that capture semantic relationships across modalities. In simpler terms, it enables the model to understand how concepts in language relate to visual elements. For example, the model learns that the word "cat" is associated with certain visual features like whiskers, pointed ears, and a furry body.
    
    \item \textbf{Transfer Learning:} This technique allows models to apply knowledge gained from one task to new, related tasks. For MLLMs, this means they can leverage general knowledge acquired from large datasets to perform well on specific tasks with minimal additional training. It's like how a human who knows how to ride a bicycle can quickly learn to ride a motorcycle, applying their balance and coordination skills to the new task.
    
    \item \textbf{Attention Mechanisms:} Originally developed for NLP, attention mechanisms allow models to focus on relevant parts of inputs. In MLLMs, this extends to focusing on relevant aspects across different modalities, enabling more effective processing of multimodal data. You can think of this as similar to how humans focus on a speaker's lips when trying to understand speech in a noisy environment.
\end{itemize}

\textbf{Key Component of AI Theory} 
\begin{figure}
    \centering
    \includegraphics[width=0.6\linewidth]{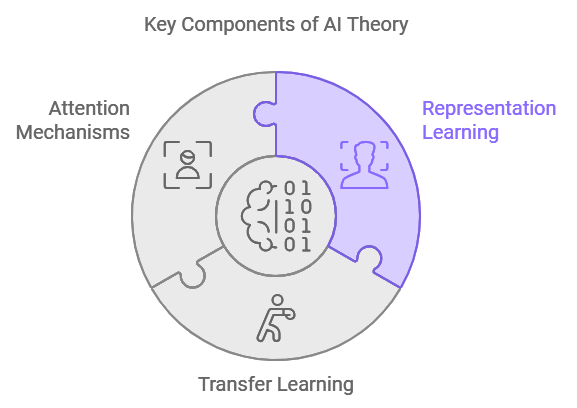}
    \caption{Key Component of AI Theory}
    \label{fig:transformer}
\end{figure}

\textbf{Architectural Innovations:}
Several key architectural innovations have enabled the development of MLLMs:

\begin{itemize}
    \item \textbf{Encoder-Decoder Frameworks:} These architectures, used in models like DALL-E, allow for mapping between text and image domains. The encoder processes the input (e.g., text), while the decoder generates the output (e.g., an image). It's like having a translator who can convert a written story into a painting.
    
    \item \textbf{Cross-Modal Transformers:} These use separate transformers for each modality, with cross-modal attention layers to fuse information. This allows the model to process text and images separately at first, then combine the information. It's similar to how humans might read a book and look at illustrations separately, then combine that information for a fuller understanding.
    
    \item \textbf{Vision Transformers (ViT):} These apply transformer architectures directly to image patches, enabling more seamless integration of vision and language models. Instead of processing an image as a whole, ViT breaks it down into smaller patches and processes them sequentially, much like how transformers process words in a sentence.
\end{itemize}

\textbf{Impact on AI Applications:}
The convergence of NLP and computer vision through MLLMs has enabled new capabilities in various AI applications:

\begin{itemize}
    \item Multimodal chatbots that understand and generate both text and images. For example, a customer service bot that can understand product images and respond with both text explanations and visual aids.
    \item Content moderation systems that analyze text and images together, providing more context-aware filtering of inappropriate content on social media platforms.
    \item Accessibility tools that generate image descriptions for visually impaired users, allowing them to "see" images through detailed textual descriptions.
    \item Enhanced human-vehicle interaction in autonomous driving systems, where the vehicle can understand both verbal commands and visual cues from the environment.
\end{itemize}

\textbf{Challenges and Future Directions:}
While MLLMs have made significant progress, several challenges remain:

\begin{itemize}
    \item \textbf{Bias and Fairness:} MLLMs can perpetuate or amplify biases present in training data across both textual and visual domains. For example, they might disproportionately misidentify individuals in images due to imbalanced training datasets. Addressing this requires careful dataset curation, diverse representation in training data, and ongoing monitoring and adjustment of model outputs. Researchers are exploring techniques like adversarial debiasing and fairness-aware learning to mitigate these issues.
    
    \item \textbf{Interpretability:} Understanding how MLLMs make decisions across modalities is crucial for building trust and improving these systems. This involves developing techniques to explain model decisions that involve both textual and visual inputs, and creating visualization tools that can effectively represent the interplay between different modalities in the model's reasoning process. Techniques like attention visualization and saliency mapping are being adapted for multimodal contexts to provide insights into model decision-making.
    
    \item \textbf{Efficiency:} Current MLLMs often require substantial computational resources. Developing more efficient architectures and training methods is an active area of research. Potential solutions include:
    \begin{itemize}
        \item Model pruning: removing unnecessary parameters to create smaller, faster models without significant loss in performance.
        \item Knowledge distillation: creating smaller models that mimic the behavior of larger ones, like a student learning from a teacher.
        \item Quantization: reducing the precision of model parameters to decrease memory usage and computational requirements.
    \end{itemize}
    
    \item \textbf{Ethical Considerations:} As MLLMs become more powerful, several ethical challenges arise:
    \begin{itemize}
        \item Privacy concerns related to the processing and potential misuse of multimodal personal data. Researchers are exploring privacy-preserving techniques like federated learning and differential privacy to address these concerns.
        \item The need for transparent decision-making processes, especially in critical applications like healthcare or autonomous systems. This involves developing explainable AI techniques that can provide clear rationales for MLLM decisions.
        \item Potential misuse for creating deepfakes or other misleading content that combines manipulated text and images. Efforts are being made to develop robust detection systems for synthetic media and to establish ethical guidelines for the use of MLLMs in content creation.
    \end{itemize}
    
    \item \textbf{Cross-modal Consistency:} Ensuring consistency across different modalities presents a significant challenge. This includes developing methods to maintain semantic consistency between generated text and images, and addressing potential conflicts when integrating information from multiple modalities. Researchers are exploring techniques like consistency regularization and multi-task learning to improve cross-modal coherence in MLLM outputs.
\end{itemize}

As research in this field progresses, we can expect MLLMs to become even more capable of understanding and generating content across diverse modalities, potentially leading to AI systems with more human-like comprehension of the world. The ongoing advancements in MLLMs continue to push the boundaries of what's possible in artificial intelligence, opening up new avenues for innovation and application across various domains.

\section{Conclusion and Future Prospects}

MLLMs represent a significant leap forward in AI technology, bridging the gap between different modes of information processing and bringing us closer to AI systems that can understand and interact with the world in ways that more closely resemble human cognition. Their ability to integrate and process multiple types of data simultaneously opens up a wide range of applications across various industries and domains.

As we look to the future, the potential impact of MLLMs is vast and transformative:

\begin{itemize}
    \item In healthcare, MLLMs could revolutionize diagnostics and treatment planning by integrating visual medical data with textual patient histories and the latest research findings. For instance, an MLLM could analyze a patient's MRI scans, medical history, and recent medical literature to suggest personalized treatment plans.
    
    \item In education, these models could create more engaging and personalized learning experiences by adapting content based on a student's multimodal interactions. An MLLM-powered tutoring system could adjust its teaching style based on a student's verbal responses, facial expressions, and performance on visual tasks.
    
    \item In scientific research, MLLMs could accelerate discoveries by analyzing complex, multimodal datasets and identifying patterns that might be missed by human researchers. For example, in climate science, an MLLM could integrate satellite imagery, weather data, and scientific papers to identify new patterns in climate change.
    
    \item In creative industries, MLLMs could become powerful tools for content creation, enabling new forms of interactive and immersive storytelling. Imagine a video game that generates unique storylines and visual content based on a player's actions and preferences.
\end{itemize}

However, as we embrace the potential of MLLMs, we must also remain vigilant about the challenges they present. Addressing issues of bias, ensuring ethical use, improving efficiency, and enhancing interpretability will be crucial in realizing the full potential of these powerful models.

\textbf{Call to Action for Researchers and Practitioners:}
\begin{itemize}
    \item Develop robust techniques for mitigating bias in multimodal datasets and model outputs.
    \item Create more efficient MLLM architectures to reduce computational requirements and environmental impact.
    \item Explore new methods for improving cross-modal consistency and coherence in MLLM outputs.
    \item Investigate the integration of MLLMs with other emerging technologies, such as augmented reality and the Internet of Things.
    \item Establish ethical guidelines and best practices for the development and deployment of MLLMs across various industries.
\end{itemize}

The development of MLLMs is not just a technological advancement; it represents a fundamental shift in how we approach artificial intelligence. By mimicking the human ability to process and integrate multiple types of information, MLLMs are bringing us closer to creating truly intelligent systems that can understand and interact with the world in more nuanced and comprehensive ways.

As research in this field continues to evolve, we can anticipate even more sophisticated MLLMs that push the boundaries of what's possible in AI. The journey ahead is filled with exciting possibilities and challenges, and the continued development of MLLMs will undoubtedly play a crucial role in shaping the future of artificial intelligence and its impact on society. It is up to researchers, practitioners, and policymakers to guide this development responsibly, ensuring that the benefits of MLLMs are realized while mitigating potential risks and ethical concerns.
  
\chapter{Foundations of Multimodal Large Language Models (MLLMs)}

The purpose of this chapter is to provide an overview of the foundational elements that have led to the development of Multimodal Large Language Models (MLLMs). It covers the evolution of Natural Language Processing (NLP) into Large Language Models (LLMs), explains the unique architecture of MLLMs, discusses the training methodologies and data requirements, and explores the capabilities of MLLMs in cross-modal understanding and visual reasoning.

Multimodal Large Language Models (MLLMs) represent a groundbreaking advancement in the field of artificial intelligence(AI), combining the power of language understanding with visual perception. These sophisticated AI systems are designed to process and comprehend multiple types of data inputs, primarily focusing on text and images, but potentially extending to audio and video as well. To fully grasp the concept of MLLMs, it's essential to break down its components and understand their significance.

At their core, MLLMs are an evolution of Large Language Models (LLMs), which are AI systems trained to understand and generate human language. The term "large" in MLLMs refers to the scale of these models, often containing billions of parameters, which allows them to capture intricate patterns and relationships in data. The "multimodal" aspect is what sets MLLMs apart from traditional LLMs. It indicates the ability to process multiple types of data or "modalities," primarily text and images in the current context.

The significance of MLLMs lies in their ability to bridge the gap between language and vision, much like humans do. When we see a picture, we can easily describe it in words, and conversely, we can visualize a scene based on a textual description. MLLMs aim to replicate this ability, enabling a more comprehensive understanding of given contexts. For instance, when analyzing a news article with accompanying images, an MLLM can provide more accurate insights by considering both the textual content and visual elements.

This enhanced understanding opens up a wide range of applications for MLLMs. They can be used for advanced image captioning, where the model generates detailed descriptions of visual content. Visual question answering is another key application, where MLLMs can respond to queries about specific elements within an image. Some advanced MLLMs are even capable of generating images based on textual descriptions, though this capability is still evolving.

To illustrate the capabilities of MLLMs, let's consider a specific example of visual question answering. Imagine an MLLM is presented with an image of a bustling city street at night, filled with neon signs and people walking. A user might ask, "What time of day is it in this image?" The MLLM, analyzing the visual content, would respond with something like, "It is nighttime in this image. The dark sky and illuminated neon signs indicate that it's evening or night." If asked, "What kind of businesses can you see?" the model might answer, "Based on the neon signs visible in the image, I can see various businesses such as restaurants, bars, and possibly some retail shops. The bright, colorful signs are typical of entertainment districts in cities."

In this example, the MLLM demonstrates several key capabilities. Firstly, it shows an understanding of the visual content of the image, recognizing that it is a night scene in a city. Secondly, it interprets specific elements within the image, such as neon signs, people, and the overall ambiance. Third, it responds to natural language questions about the image content, showcasing its language processing abilities. Finally, it provides detailed, context-aware answers that combine visual understanding with language generation.

The development of MLLMs represents a crucial step towards more human-like AI systems. By processing both text and images, these models can gain a more comprehensive understanding of a given context, much like humans do when interpreting information from multiple sources. This capability enhances human-AI interaction, paving the way for more natural and intuitive exchanges between people and AI systems.

MLLMs possess several key capabilities that set them apart from traditional language models. They can analyze and interpret the content of images, identifying objects, scenes, actions, and emotions described in images. Having a deep understanding of the relationship between textual descriptions and accompanying visuals, they are adept at aligning text and images. Both visual and linguistic information can be comprehended by their visual question-answering abilities. Additionally, MLLMs can generate descriptive captions for images, effectively translating visual information into natural language.

As MLLMs continue to evolve, they promise to revolutionize how we interact with AI systems and how machines understand and interpret the world around us. Their potential applications span various fields, including content creation, education, accessibility, and more. For instance, in education, MLLMs could enhance learning materials by providing detailed explanations of complex diagrams or historical images. In the field of accessibility, they could assist visually impaired individuals by describing the content of images or navigating visual interfaces.

However, it is important to note that the development and application of MLLMs also come with challenges and ethical considerations. These include ensuring the accuracy and fairness of the model's interpretations, addressing potential biases in training data, and considering privacy concerns when processing visual information.

In conclusion, multimodal large language models are a significant step forward in AI technology, bridging the gap between language understanding and visual perception. In the coming years, we can expect to see even more sophisticated applications that utilize MLLMs to enhance our interaction with digital content.

\section{From NLP to LLMs: A Brief Overview}

The history of Multimodal Large Language Models (MLLMs) is deeply rooted in the evolution of Natural Language Processing (NLP). Understanding this progression from traditional NLP methods to modern Large Language Models (LLMs) provides critical insights into how MLLMs were developed and sheds light on their current capabilities and potential future directions.

Early NLP techniques relied heavily on rule-based systems and statistical models. Rule-based systems used handcrafted rules to process and analyze text, which were effective for specific tasks but lacked flexibility and struggled with the complexity of natural language. Statistical models, such as n-grams and Hidden Markov Models (HMMs), introduced probabilistic methods to NLP, allowing for better handling of language variability. However, these models still fell short in capturing deeper linguistic structures.

The integration of machine learning into NLP marked a significant shift, enabling models to learn from data rather than relying solely on predefined rules. Support Vector Machines (SVMs) were used for text classification tasks, leveraging the ability to find optimal decision boundaries in high-dimensional spaces. Naive Bayes, a probabilistic classifier, was popular for tasks like spam detection, utilizing the Bayes theorem to make predictions based on word frequencies.

The rise of deep learning brought transformative changes to NLP, with neural networks enabling more sophisticated language models. Techniques like Word2Vec and GloVe represented words as dense vectors in continuous space, capturing semantic relationships and improving the performance of downstream tasks. Recurrent Neural Networks (RNNs), and their variants like Long Short-Term Memory (LSTM) networks, excelled at processing sequential data, making them suitable for tasks such as language modeling and machine translation. The introduction of attention mechanisms allowed models to focus on relevant parts of the input sequence, enhancing the performance of tasks like translation and summarization.

The development of the Transformer architecture revolutionized NLP, leading to the creation of powerful pre-trained models. The Transformer model, introduced in the paper \cite{vaswani2017attention} utilized self-attention mechanisms to capture dependencies between words, enabling parallel processing and improving scalability. Bidirectional Encoder Representations from Transformers (BERT) leveraged the Transformer architecture to create contextualized word embeddings, achieving state-of-the-art results on various NLP benchmarks. Generative Pre-trained Transformers (GPT) focused on language generation, with models like GPT-3 demonstrating remarkable capabilities in generating coherent and contextually relevant text.

The evolution from LLMs to MLLMs involved integrating visual data with textual data, enabling models to process and understand multiple modalities. Techniques like VisualBERT and VL-BERT extended the BERT architecture to handle both text and images, pre-training on large-scale multimodal datasets to learn joint representations. Cross-modal attention mechanisms allowed models to align and integrate information from different modalities, enhancing their ability to perform tasks like image captioning and visual question answering.

The integration of multimodal data has opened up new possibilities for AI applications. MLLMs can generate detailed descriptions of images, providing valuable assistance in fields like accessibility and content creation. These models can answer questions about images, demonstrating their ability to understand and reason about visual content. MLLMs also enable the creation of rich multimedia content, combining text, images, and audio to produce engaging and informative outputs.

The field of NLP has undergone significant transformations over the past few decades, driven by advancements in computational power, the availability of large-scale datasets, and breakthroughs in machine learning algorithms. This journey can be broadly categorized into several key phases:

\begin{enumerate}
    \item \textbf{Rule-based systems (1950s-1980s):} Early NLP approaches relied heavily on hand-crafted rules and linguistic knowledge bases. While these systems could perform well in narrow domains, they struggled with the complexity and ambiguity of natural language.

    These were handcrafted algorithms that relied on predefined grammatical rules to process text. Though effective in limited contexts, they were inflexible and incapable of handling complex linguistic nuances.

These handcrafted algorithms relied heavily on predefined grammatical rules to process text. Though effective in limited contexts, they were inflexible and incapable of handling complex linguistic nuances. This rigidity hindered their adaptability to diverse scenarios where language context and meaning varied significantly. As a result, these systems struggled with ambiguous or highly contextual inputs, which are common in natural language.

The shortcomings of such rule-based systems, particularly in Natural Language Processing (NLP), highlighted the need for more sophisticated approaches. Rule-based systems were successful in solving specific, well-defined tasks like discourse segmentation or parsing sentences based on syntactic structures, as seen in early attempts like the Linguistic Discourse Model (LDM). These systems used symbolic, syntactic, and semantic rules to process language but required meticulous construction and could only manage relatively straightforward linguistic structures \cite{polanyi2004rule}. Similarly, conventional rule-based expert systems were developed to encode human expertise into conditional rules to address complex decision-making tasks across various domains. These systems were designed with “if-then” logic, enabling machines to simulate human problem-solving \cite{abraham2005rule}.

However, the limitations of these early systems in managing real-world complexity soon became evident. They lacked the flexibility to deal with incomplete or noisy data and were often constrained by the size and scope of their rule base. Despite their limitations, rule-based systems laid the foundation for modern AI, particularly in providing a framework for the development of inference mechanisms like forward and backward chaining. These mechanisms played a crucial role in expert systems, where the aim was to mimic human decision-making processes \cite{abraham2005rule}.

The development of Large Language Models (LLMs) and other machine learning techniques gradually overcame these limitations by using vast amounts of data and statistical methods, allowing for a more nuanced understanding of language and context.

    \item \textbf{Statistical methods (1980s-2000s):} The introduction of statistical techniques, such as Hidden Markov Models and n-gram language models, allowed for more data-driven approaches. These methods could learn patterns from large text corpora, improving performance on tasks like machine translation and speech recognition. This marked a significant departure from rule-based systems, where linguistic expertise and predefined rules dominated.

With statistical models, the focus shifted from handcrafted rules to probabilistic reasoning. For example, HMMs became instrumental in tagging and parsing, where they modeled sequences of observations, such as words or parts of speech, and the probabilities of transitions between states \cite{koehn2009statistical}, \cite{charniak1997statistical}. These models could infer the most likely sequence of syntactic structures or translations based on observed data, making them more adaptable to unseen examples than rule-based systems.

In addition, n-gram models, which represent the probability of a word given its previous words, found success in tasks such as language modeling for speech recognition and machine translation. These models utilized large amounts of data to estimate the likelihood of word sequences, enabling them to predict future words in a sentence with remarkable accuracy \cite{koehn2009statistical}. By calculating probabilities over sequences of tokens, n-gram models helped achieve more fluent and coherent outputs in machine translation \cite{charniak1997statistical}.

Despite their progress, early statistical models had limitations. They were prone to issues like sparse data, where rare word combinations were poorly represented, leading to less accurate predictions. Techniques like smoothing were introduced to mitigate this by assigning small probabilities to unseen word sequences \cite{charniak1997statistical}. Nevertheless, these models laid the groundwork for the more advanced machine learning techniques that would follow, which incorporated deeper linguistic understanding and more sophisticated representations.
    
    \item \textbf{Machine learning era (2000s-2010s):} The rise of machine learning algorithms, particularly supervised learning techniques, improved various NLP tasks significantly. Support Vector Machines (SVM), Decision Trees, and early neural networks became popular for tasks like text classification and named entity recognition. These approaches allowed models to learn from labeled data, where patterns and relationships could be automatically extracted from examples rather than being manually coded as in rule-based systems.

According to \cite{mi2016supervised}, SVMs, for instance, worked by finding hyperplanes that best separated data into different classes, excelling in high-dimensional spaces common in NLP applications. They were especially useful for binary classification tasks, such as determining whether a document belongs to a certain category, like spam detection in emails. Decision Trees, on the other hand, divided data into increasingly smaller subsets based on feature values, creating a tree-like structure that could be easily interpreted and adapted to a variety of tasks.

Early neural networks also began to gain traction during this time, laying the groundwork for the deep learning revolution to follow. These networks were particularly adept at handling complex, nonlinear relationships in data, which was essential for more challenging tasks like machine translation, where simple linear models struggled \cite{conneau2017supervised}. While these early machine learning models marked a significant leap forward compared to rule-based systems, they still faced challenges in generalization and scalability. However, their data-driven nature allowed them to learn from diverse sources and provided a more flexible alternative to the rigid frameworks of earlier approaches.

This transition from manually crafted rules to models capable of learning from data formed the basis for modern NLP approaches, which now rely on even more sophisticated algorithms and architectures, like neural networks with attention mechanisms, to achieve state-of-the-art results \cite{mi2016supervised}, \cite{conneau2017supervised}.
    
    \item \textbf{Deep learning revolution (2010s-present):} The advent of deep learning techniques, especially neural networks with multiple layers, marked a paradigm shift in NLP. This era saw the development of word embeddings, recurrent neural networks (RNNs), and later, transformer-based models, which dramatically improved performance across a wide range of NLP tasks.

    The advent of deep learning techniques, especially neural networks with multiple layers, marked a paradigm shift in NLP. This era saw the development of word embeddings, recurrent neural networks (RNNs), and later, transformer-based models, which dramatically improved performance across a wide range of NLP tasks. Word embeddings, such as Word2Vec and GloVe, allowed models to represent words as dense vectors in continuous space, capturing semantic similarities between words that previous methods like one-hot encoding failed to do \cite{lauriola2022introduction}, \cite{henderson2020unstoppable}. These embeddings enabled models to generalize better and capture context more effectively, leading to significant breakthroughs in tasks like sentiment analysis, machine translation, and text classification.

Recurrent Neural Networks (RNNs), especially their variants like Long Short-Term Memory (LSTM) networks, were introduced to address the challenge of capturing sequential dependencies in language. These models were able to maintain a memory of previous inputs, making them particularly effective for tasks involving temporal or sequential data, such as language modeling, speech recognition, and machine translation \cite{peng2022survey}, \cite{lauriola2022introduction}. However, RNNs faced limitations in handling long-term dependencies due to issues like vanishing gradients.

The limitations of RNNs were later addressed with the development of Transformer-based models, which relied entirely on self-attention mechanisms. The Transformer model, introduced in 2017, revolutionized NLP by allowing for parallelization in training and handling long-range dependencies more effectively than RNNs. This architecture formed the foundation for models like BERT (Bidirectional Encoder Representations from Transformers) and GPT (Generative Pre-trained Transformer), which became the cornerstone of modern NLP applications. These models could be pre-trained on large corpora and fine-tuned for specific tasks, leading to state-of-the-art results in machine translation, question answering, and summarization \cite{lauriola2022introduction}, \cite{henderson2020unstoppable}.
\end{enumerate}

\begin{center}
\begin{tikzpicture}[scale=1, transform shape]
    \draw[thick] (0,0) -- (14,0);

    \node[fill=blue!20, rounded corners, text width=3cm, align=center, above=1cm] at (2,0) {\textbf{Rule-based Systems}\\(1950s-1980s)};
    \node[fill=gray!10, text width=3.5cm, rounded corners, below=1.5cm] at (2,0) {
        Handcrafted rules for text processing, limited in adaptability};
    \draw[thick] (2, -0.3) -- (2, 0.3); 
    \node[below=0cm] at (2, -0.7) {1950s - 1980s};

    \node[fill=green!20, rounded corners, text width=3cm, align=center, above=1cm] at (6,0) {\textbf{Statistical Methods}\\(1980s-2000s)};
    \node[fill=gray!10, text width=3.5cm, rounded corners, below=1.5cm] at (6,0) {
        Probabilistic models (e.g., HMMs, n-grams) for text pattern recognition};
    \draw[thick] (6, -0.3) -- (6, 0.3); 
    \node[below=0cm] at (6, -0.7) {1980s - 2000s};

    \node[fill=orange!20, rounded corners, text width=3cm, align=center, above=1cm] at (10,0) {\textbf{Machine Learning Era}\\(2000s-2010s)};
    \node[fill=gray!10, text width=3.5cm, rounded corners, below=1.5cm] at (10,0) {
        SVMs, Decision Trees, and early neural networks for NLP tasks};
    \draw[thick] (10, -0.3) -- (10, 0.3); 
    \node[below=0cm] at (10, -0.7) {2000s - 2010s};

    \node[fill=red!20, rounded corners, text width=3cm, align=center, above=1cm] at (14,0) {\textbf{Deep Learning Revolution}\\(2010s - present)};
    \node[fill=gray!10, text width=3.5cm, rounded corners, below=1.5cm] at (14,0) {
        Word embeddings, RNNs, Transformers drive NLP advances};
    \draw[thick] (14, -0.3) -- (14, 0.3); 
    \node[below=0cm] at (14, -0.7) {2010s - present};

\end{tikzpicture}
\end{center}

The transition from traditional NLP methods to Large Language Models (LLMs) represents a significant leap in the field's capabilities. LLMs, such as GPT (Generative Pre-trained Transformer) and BERT (Bidirectional Encoder Representations from Transformers), leverage massive amounts of text data and advanced neural network architectures to capture complex linguistic patterns and generate human-like text.

This evolution set the stage for the development of MLLMs, which extend the capabilities of LLMs to handle multiple modalities, particularly integrating vision and language. By understanding the historical context and technological advancements that led to LLMs, we can better appreciate the challenges and opportunities presented by MLLMs in combining textual and visual information processing.

\subsection{Traditional NLP Methods}
Traditional NLP methods focused on rule-based systems and statistical models. These early approaches included:

\textbf{Rule-Based Systems:} These were handcrafted algorithms that relied on predefined grammatical rules to process text. Although effective in limited contexts, they were inflexible and incapable of handling complex linguistic nuances. Despite their limitations, rule-based systems played a crucial role in the evolution of Natural Language Processing and laid the groundwork for more advanced techniques. These systems contributed significantly to our understanding of language structure and the challenges inherent in computational linguistics.

One notable application of rule-based systems was in the development of early machine translation systems. The Georgetown-IBM experiment in 1954, for instance, utilized a rule-based approach to translate Russian sentences into English. While limited in scope, this experiment demonstrated the potential for automated language processing and sparked further research in the field. Rule-based systems also found applications in information extraction tasks. Systems like FASTUS (Finite State Automaton Text Understanding System) employed cascaded finite-state transducers to extract specific information from text. This approach proved effective for well-defined extraction tasks in domains with structured information. Moreover, rule-based systems contributed to the development of formal grammars and parsing techniques. The work on context-free grammars and parsing algorithms, such as the CYK algorithm, provided a foundation for understanding sentence structure and laid the groundwork for more sophisticated natural language understanding systems. The transition from rule-based systems to statistical and machine learning approaches in NLP was gradual. Hybrid systems that combined rules with statistical methods emerged as an intermediate step. These systems attempted to leverage the strengths of both approaches, using rules to capture linguistic knowledge and statistical methods to handle variability and ambiguity in language.

While modern Large Language Models have largely superseded traditional rule-based systems in many NLP tasks, the insights gained from developing and refining rule-based approaches continue to inform current research. The explicit encoding of linguistic knowledge in rules has influenced the design of neural network architectures and the development of linguistically-informed learning objectives in contemporary NLP models.

\textbf{Bag-of-Words and TF-IDF:} These models represented text as a collection of words (bag-of-words), neglecting grammar or word order, but capturing word frequency to identify relevant terms. TF-IDF (Term Frequency-Inverse Document Frequency) improved on this by down-weighting common words and highlighting unique ones. These models, while simplistic in their approach, laid important groundwork for text representation in NLP tasks.

\subsubsection{Bag-of-Words (BoW)}
The Bag-of-Words (BoW) model is a fundamental text representation technique that disregards grammar and word order, instead focusing on the occurrence of words within a document. In this model, each document is represented as a ``bag'' of its words, typically encoded as a vector where each element corresponds to a word in the vocabulary. The value of each element can be binary (indicating presence or absence), the count of occurrences, or some other weighting scheme. Despite its simplicity, BoW has proven effective in various applications, including document classification and information retrieval.

However, the BoW model has limitations. It fails to capture semantic relationships between words and can be biased towards frequently occurring terms that may not be particularly informative. To address these issues, the Term Frequency-Inverse Document Frequency (TF-IDF) model was developed.

\subsubsection{Term Frequency-Inverse Document Frequency (TF-IDF)}
TF-IDF is a statistical measure used to evaluate the importance of a word in a document within a corpus. It comprises two components:

\begin{enumerate}
    \item \textbf{Term Frequency (TF):} This measures how frequently a term appears in a document. It is typically calculated as:
    
    \begin{equation}
        \text{TF}(t,d) = \frac{\text{Number of times term } t \text{ appears in document } d}{\text{Total number of terms in document } d}
    \end{equation}

    \item \textbf{Inverse Document Frequency (IDF):} This measures how important a term is across the entire corpus. It is calculated as:
    
    \begin{equation}
        \text{IDF}(t) = \log\left(\frac{\text{Total number of documents}}{\text{Number of documents containing term } t}\right)
    \end{equation}
\end{enumerate}

The TF-IDF score is then computed by multiplying TF and IDF:

\begin{equation}
    \text{TF-IDF}(t,d) = \text{TF}(t,d) \times \text{IDF}(t)
\end{equation}

This weighting scheme effectively reduces the impact of common words that appear frequently across many documents (like ``the'' or ``and'') while emphasizing words that are more unique to specific documents. As a result, TF-IDF provides a more nuanced representation of document content compared to simple word counts.

\paragraph{Example: Calculating TF-IDF}

Consider a corpus containing three documents:
\begin{itemize}
    \item Document 1: ``apple banana apple apple orange"
    \item Document 2: ``banana apple orange orange"
    \item Document 3: ``banana banana apple"
\end{itemize}

We want to calculate the TF-IDF score for the term ``apple" in Document 1.

\begin{enumerate}
    \item \textbf{Calculate Term Frequency (TF):}
    
    In Document 1, the term ``apple" appears 3 times, and there are 5 terms in total. Therefore:
    \[
    \text{TF}(\text{apple}, \text{Document 1}) = \frac{3}{5} = 0.6
    \]

    \item \textbf{Calculate Inverse Document Frequency (IDF):}
    
    The term ``apple" appears in all three documents in the corpus. Therefore, the IDF for ``apple" is:
    \[
    \text{IDF}(\text{apple}) = \log\left(\frac{\text{Total number of documents}}{\text{Number of documents containing ``apple"}}\right) = \log\left(\frac{3}{3}\right) = 0
    \]

    \item \textbf{Compute TF-IDF:}
    
    Finally, the TF-IDF score for ``apple" in Document 1 is:
    \[
    \text{TF-IDF}(\text{apple}, \text{Document 1}) = \text{TF}(\text{apple}, \text{Document 1}) \times \text{IDF}(\text{apple}) = 0.6 \times 0 = 0
    \]
\end{enumerate}

In this case, because ``apple" appears in every document in the corpus, its IDF is zero, resulting in a TF-IDF score of zero. This demonstrates how common terms across the corpus receive lower TF-IDF scores.

Both BoW and TF-IDF models have been widely used in various NLP tasks, including document classification, information retrieval, and as features for more complex machine learning models. While they have largely been superseded by more advanced techniques in many applications, their simplicity and interpretability ensure their continued relevance in certain contexts, particularly in scenarios with limited computational resources or where model explainability is crucial.

\textbf{Early Machine Learning Models:} 

Early machine learning models revolutionized Natural Language Processing (NLP), introducing statistical approaches that could learn patterns from data. These models, while still limited in their ability to capture deep linguistic context, represented a crucial step forward in the evolution of NLP techniques. Notable among these early models were Naive Bayes classifiers, which applied probabilistic methods to text classification; Support Vector Machines (SVMs), which excelled in high-dimensional spaces typical of text data; and Hidden Markov Models (HMMs), which proved particularly effective for sequential data processing in tasks such as part-of-speech tagging. These approaches laid the groundwork for more advanced techniques, demonstrating the potential of statistical methods in language processing and paving the way for the deep learning revolution in NLP.

\subsection{Rise of Large Language Models (LLMs)}
To overcome the limitations of traditional methods, Large Language Models (LLMs) were developed, which leveraged deep learning architectures such as neural networks. In NLP, LLMs represent a paradigm shift from traditional rule-based and statistical approaches to more sophisticated, data-driven approaches. This evolution was catalyzed by advancements in deep learning architectures, increased computational capabilities, and the availability of vast amounts of textual data.

\subsubsection{Neural Network Foundations}
The foundation for LLMs was laid with the advent of neural network-based models in NLP. These models, particularly those utilizing deep learning techniques, demonstrated an unprecedented ability to capture complex linguistic patterns and relationships. Unlike their predecessors, neural models could automatically learn hierarchical representations of language, reducing the need for manual feature engineering.

\subsubsection{Emergence of Transformer Architecture}
A pivotal moment in the development of LLMs was the introduction of the Transformer architecture by Vaswani et al. (2017) in their seminal paper "Attention is All You Need". The Transformer's self-attention mechanism allowed models to process input sequences in parallel, capturing long-range dependencies more effectively than previous sequential models like Recurrent Neural Networks (RNNs) and Long Short-Term Memory (LSTM) networks.

\subsubsection{Pre-training and Transfer Learning}
The concept of pre-training on large corpora of unlabeled text data, followed by fine-tuning on specific tasks, became a cornerstone in the development of LLMs. This approach, exemplified by models like BERT (Bidirectional Encoder Representations from Transformers) by Devlin et al. (2018), allowed models to acquire general language understanding that could be transferred to a wide range of NLP tasks with minimal task-specific training.

\subsubsection{Scaling Up: From BERT to GPT}
The progression from BERT to models like GPT (Generative Pre-trained Transformer) by OpenAI marked a significant increase in model size and capability. GPT-3, introduced by Brown et al. (2020), with its 175 billion parameters, demonstrated remarkable few-shot learning abilities across various language tasks, approaching human-level performance in some cases.

\subsubsection{Multimodal Extensions}
As LLMs continued to evolve, researchers began exploring ways to extend their capabilities beyond text, leading to the development of multimodal models. These models, capable of processing both textual and visual information, represent the next frontier in AI, bridging the gap between language understanding and visual perception.

\subsubsection{Ethical and Computational Considerations}
The rise of LLMs has also brought forth important discussions regarding the ethical implications of these powerful models, including issues of bias, privacy, and the environmental impact of training such large-scale systems. Additionally, the computational resources required for training and deploying LLMs have spurred research into more efficient architectures and training methodologies.


\textbf{Word Embeddings:} Techniques such as Word2Vec and GloVe introduced the idea of embedding words in a continuous vector space, where the distance between words reflects their semantic relationships. This allowed for a more nuanced understanding of language. Word embeddings represent a significant advancement in the field of NLP, offering a sophisticated method for representing words as dense vectors in a continuous vector space. This approach, pioneered by techniques such as Word2Vec (\cite{mikolov2013efficient}) and GloVe (\cite{pennington2014glove}), revolutionized the way machines process and understand language by capturing semantic relationships between words in a numerical format.

The fundamental principle behind word embeddings is the distributional hypothesis, which posits that words appearing in similar contexts tend to have similar meanings. By leveraging this hypothesis, word embedding models analyze large corpora of text to learn vector representations of words. These vectors typically have dimensions ranging from 50 to 300, allowing for a rich representation of semantic and syntactic properties.

One of the most notable characteristics of word embeddings is their ability to capture semantic relationships in vector space. For instance, in a well-trained embedding space, the vector operation "king" - "man" + "woman" would result in a vector close to "queen". This property enables machines to perform analogical reasoning and understand complex linguistic relationships.

Word embeddings have several advantages over traditional one-hot encoding methods:
\begin{enumerate}
    \item \textbf{Dimensionality Reduction:} They provide a dense representation of words, significantly reducing the dimensionality compared to sparse representations like one-hot encoding.
    \item \textbf{Semantic Similarity:} The cosine similarity between word vectors correlates with semantic similarity, allowing for meaningful comparisons between words.
    \item \textbf{Generalization:} Word embeddings enable models to generalize better to unseen words by leveraging the semantic relationships learned during training.
    \item \textbf{Transfer Learning:} Pre-trained word embeddings can be used as input features for various NLP tasks, facilitating transfer learning and improving performance on downstream tasks.
\end{enumerate}

Despite their advantages, word embeddings also have limitations. They struggle with polysemy (words with multiple meanings) and require large amounts of training data to learn accurate representations. Additionally, they are static, meaning the same word always has the same embedding regardless of context.

Subsequent developments, such as contextual embeddings (e.g., ELMo, BERT), have addressed some of these limitations by generating dynamic word representations based on the surrounding context. Nevertheless, the introduction of word embeddings marked a pivotal moment in NLP, laying the groundwork for many of the advanced language models we see today.

\textbf{Recurrent Neural Networks (RNNs) and Long Short-Term Memory (LSTM):} These models improved sequential data processing by retaining information over longer time steps, making them useful for tasks like language translation and text generation.

Recurrent Neural Networks (RNNs) and Long Short-Term Memory (LSTM) networks represent significant advancements in sequential data processing, particularly in the domain of Natural Language Processing (NLP). These architectures addressed the limitations of traditional feedforward neural networks by introducing mechanisms to retain information over extended sequences, making them particularly well-suited for tasks such as language translation, text generation, and sentiment analysis.

\textbf{Recurrent Neural Networks (RNNs):}
RNNs introduced a novel approach to processing sequential data by incorporating feedback connections, allowing information to persist through time steps. This recurrent structure enables the network to maintain a form of 'memory' about previous inputs, which is crucial for understanding context in language processing tasks. The basic RNN architecture consists of a hidden state that is updated at each time step, taking into account both the current input and the previous hidden state.

\begin{center}
\textbf{Recurrent Neural Network (RNN) Structure}
\begin{tikzpicture}[scale=1, transform shape, node distance=1.8cm]
    \tikzstyle{layer} = [rectangle, draw=black, rounded corners, minimum width=1.5cm, minimum height=0.8cm, align=center, fill=blue!10]
    \tikzstyle{arrow} = [->, thick]

    \node[layer, fill=green!20] (x1) at (0, 0) {$x_1$};
    \node[layer, fill=green!20, right=2.5cm of x1] (x2) {$x_2$};
    \node[layer, fill=green!20, right=2.5cm of x2] (x3) {$x_3$};

    \node[layer, fill=blue!20, above=1.5cm of x1] (h1) {$h_1$};
    \node[layer, fill=blue!20, right=2.5cm of h1] (h2) {$h_2$};
    \node[layer, fill=blue!20, right=2.5cm of h2] (h3) {$h_3$};

    \node[layer, fill=red!20, above=1.5cm of h1] (y1) {$y_1$};
    \node[layer, fill=red!20, right=2.5cm of y1] (y2) {$y_2$};
    \node[layer, fill=red!20, right=2.5cm of y2] (y3) {$y_3$};

    \draw[arrow] (x1) -- (h1);
    \draw[arrow] (x2) -- (h2);
    \draw[arrow] (x3) -- (h3);

    \draw[arrow] (h1) -- (y1);
    \draw[arrow] (h2) -- (y2);
    \draw[arrow] (h3) -- (y3);

    \draw[arrow] (h1) -- (h2);
    \draw[arrow] (h2) -- (h3);

    \node[below=0.2cm] at (x1.south) {Input $x_1$};
    \node[below=0.2cm] at (x2.south) {Input $x_2$};
    \node[below=0.2cm] at (x3.south) {Input $x_3$};

    \node[above=0.2cm] at (y1.north) {Output $y_1$};
    \node[above=0.2cm] at (y2.north) {Output $y_2$};
    \node[above=0.2cm] at (y3.north) {Output $y_3$};

    \node[left=0.2cm] at (h1.west) {Hidden $h_1$};
    \node[left=0.2cm] at (h2.west) {Hidden $h_2$};
    \node[left=0.2cm] at (h3.west) {Hidden $h_3$};

\end{tikzpicture}
\end{center}

Mathematically, the hidden state $h_t$ at time $t$ is computed as:
\begin{equation}
    h_t = f(W_{hh}h_{t-1} + W_{xh}x_t + b_h)
\end{equation}

Where $W_{hh}$ and $W_{xh}$ are weight matrices, $x_t$ is the input at time $t$, $b_h$ is a bias term, and $f$ is typically a non-linear activation function such as tanh or ReLU.

Despite their ability to capture temporal dependencies, traditional RNNs suffer from the vanishing gradient problem, which limits their effectiveness in learning long-range dependencies. This limitation led to the development of more sophisticated architectures, notably the Long Short-Term Memory networks.

\textbf{Long Short-Term Memory (LSTM):}
LSTM networks, introduced by \cite{hochreiter1997long}, were designed to mitigate the vanishing gradient problem inherent in standard RNNs. LSTMs incorporate a more complex internal structure, featuring a memory cell and three gating mechanisms: input gate, forget gate, and output gate. These gates regulate the flow of information into, out of, and within the memory cell, allowing the network to selectively remember or forget information over long sequences.

\begin{center}
\textbf{Long Short-Term Memory (LSTM) Network Structure}
\begin{tikzpicture}[scale=1, transform shape]
    \tikzstyle{cell} = [rectangle, rounded corners, draw=black, fill=blue!10, minimum width=1.8cm, minimum height=0.8cm]
    \tikzstyle{gate} = [circle, draw=black, fill=orange!20, minimum width=0.8cm]
    \tikzstyle{arrow} = [->, thick]

    \node[cell, fill=green!20] (c_prev) at (0, 2) {$c_{t-1}$};
    \node[above] at (c_prev.north) {Previous Cell State};

    \node[gate] (forget) at (3.5, 2) {$f_{t}$};
    \node[above] at (forget.north) {Forget Gate};

    \node[gate] (input) at (3.5, -1) {$i_{t}$};
    \node[below] at (input.south) {Input Gate};

    \node[cell, fill=yellow!20, minimum width=1.6cm] (c_tilde) at (6, -1) {$\tilde{c}_{t}$};
    \node[below] at (c_tilde.south) {Candidate Cell};

    \node[cell, fill=green!20, minimum width=1.8cm, minimum height=1.5cm] (cell) at (8.5, 0.5) {$c_{t}$};
    \node[above] at (cell.north) {Cell State};

    \node[gate] (output) at (11, 0.5) {$o_{t}$};
    \node[above] at (output.north) {Output Gate};

    \node[cell, fill=red!20, minimum width=1.6cm] (h) at (13, 0.5) {$h_{t}$};
    \node[above] at (h.north) {Hidden State};

    \node[cell, fill=gray!20, minimum width=1.6cm] (x_t) at (3.5, -3) {$x_{t}$};
    \node[below] at (x_t.south) {Input};

    \draw[arrow] (c_prev) -- (forget);
    \draw[arrow] (forget) -- node[midway, above] {$f_{t} \cdot c_{t-1}$} (cell);
    \draw[arrow] (x_t) -- (input);
    \draw[arrow] (x_t) -| (c_tilde);
    \draw[arrow] (input) -- node[midway, above] {$i_{t} \cdot \tilde{c}_{t}$} (c_tilde);
    \draw[arrow] (c_tilde) -| (cell);
    \draw[arrow] (cell) -- (output);
    \draw[arrow] (output) -- (h);

    \node at (13, -0.5) {Output $h_{t}$};

\end{tikzpicture}
\end{center}

The LSTM update equations are as follows:

\begin{align}
f_t &= \sigma(W_f \cdot [h_{t-1}, x_t] + b_f) \\
i_t &= \sigma(W_i \cdot [h_{t-1}, x_t] + b_i) \\
\tilde{C}_t &= \tanh(W_C \cdot [h_{t-1}, x_t] + b_C) \\
C_t &= f_t * C_{t-1} + i_t * \tilde{C}_t \\
o_t &= \sigma(W_o \cdot [h_{t-1}, x_t] + b_o) \\
h_t &= o_t * \tanh(C_t)
\end{align}

Where $\sigma$ denotes the sigmoid function, $*$ represents element-wise multiplication, $[h_{t-1}, x_t]$ is the concatenation of the previous hidden state and current input, and $W_f$, $W_i$, $W_C$, $W_o$ are weight matrices for the forget gate, input gate, cell state, and output gate, respectively.

The sophisticated gating mechanism of LSTMs allows them to capture long-term dependencies more effectively than standard RNNs. This capability has made LSTMs particularly successful in various NLP tasks, including machine translation, speech recognition, and text summarization.

Both RNNs and LSTMs have played pivotal roles in advancing the field of NLP, serving as foundational architectures for many state-of-the-art models. Their ability to process sequential data and capture temporal dependencies has been instrumental in improving performance on a wide range of language-related tasks. While more recent architectures like Transformers have surpassed RNNs and LSTMs in many applications, the principles underlying these recurrent models continue to influence the design of modern NLP systems.

\textbf{Transformers and BERT:} The introduction of the Transformer architecture in the paper Attention is All You Need revolutionized NLP. Transformers enabled models like BERT (Bidirectional Encoder Representations from Transformers) to perform well across various NLP tasks by utilizing self-attention mechanisms to capture contextual relationships between words.

The Transformer architecture, introduced by Vaswani et al. (2017) in their seminal paper "Attention is All You Need," marked a paradigm shift in natural language processing. This model eschews recurrence and convolutions entirely in favor of self-attention mechanisms, allowing for more efficient parallel processing and improved modeling of long-range dependencies in sequential data.

\begin{center}
\textbf{Self-Attention Mechanism}
\begin{tikzpicture}[scale=1, transform shape]
    \tikzstyle{matrix} = [rectangle, draw=black, minimum width=1.5cm, minimum height=0.8cm, align=center, fill=blue!10]
    \tikzstyle{arrow} = [->, thick]

    \node[matrix, fill=green!20] (Q) at (0, 0) {$Q$};
    \node[below=0.3cm of Q] {Query};

    \node[matrix, fill=orange!20] (K) at (0, -2) {$K$};
    \node[below=0.3cm of K] {Key};

    \node[matrix, fill=yellow!20] (V) at (0, -4) {$V$};
    \node[below=0.3cm of V] {Value};

    \node[matrix, minimum width=2cm, fill=gray!20] (QK) at (3, -1) {$QK^T$};
    \node[below=0.3cm of QK] {Similarity Scores};

    \node[matrix, minimum width=2cm, fill=gray!20] (scale) at (5.5, -1) {$\frac{QK^T}{\sqrt{d_k}}$};
    \node[below=0.3cm of scale] {Scaling};

    \node[matrix, fill=blue!20] (softmax) at (8, -1) {softmax};
    \node[below=0.3cm of softmax] {Attention Weights};

    \node[matrix, fill=red!20] (output) at (10.5, -2.5) {Output};
    \node[below=0.3cm of output] {Weighted Sum};

    \draw[arrow] (Q) -- (QK);
    \draw[arrow] (K) -- (QK);
    \draw[arrow] (QK) -- (scale) node[midway, above=0.4cm] {Scaling};
    \draw[arrow] (scale) -- (softmax) node[midway, above=0.4cm] {Softmax};
    \draw[arrow] (softmax) -- (output) node[midway, above right] {Weights $\cdot V$};
    \draw[arrow] (V) -- (output);

\end{tikzpicture}
\end{center}

At the core of the Transformer architecture is the multi-head attention mechanism. This mechanism allows the model to jointly attend to information from different representation subspaces at different positions. Mathematically, for a given query Q, key K, and value V, the attention function is computed as:
\begin{equation}
    Attention(Q, K, V) = softmax(\frac{QK^T}{\sqrt{d_k}})V
\end{equation}

Where $d_k$ is the dimension of the key vectors. The multi-head attention extends this by linearly projecting the queries, keys, and values h times with different learned projections, applying attention to each projection in parallel, and concatenating the results.

The Transformer architecture consists of an encoder and a decoder, each composed of a stack of identical layers. Each layer in the encoder contains two sub-layers: a multi-head self-attention mechanism and a position-wise fully connected feed-forward network. The decoder is similar but includes an additional sub-layer that performs multi-head attention over the output of the encoder stack. Residual connections and layer normalization are employed around each sub-layer to facilitate training of deep models.

Building upon the Transformer architecture, Bidirectional Encoder Representations from Transformers (BERT), introduced by Devlin et al. (2018), further revolutionized NLP by introducing a powerful pre-training approach. BERT is designed to pre-train deep bidirectional representations from unlabeled text by jointly conditioning on both left and right context in all layers.

BERT's pre-training involves two unsupervised tasks:

\begin{enumerate}
    \item \textbf{Masked Language Model (MLM):} A certain percentage of input tokens are masked, and the model is trained to predict these masked tokens based on the context provided by the non-masked tokens.
    \item \textbf{Next Sentence Prediction (NSP):} The model is trained to predict whether a given sentence follows another in the original text, helping it understand the relationship between sentences.
\end{enumerate}

The pre-trained BERT model can then be fine-tuned with just one additional output layer to create state-of-the-art models for a wide range of NLP tasks, such as question answering, sentiment analysis, and named entity recognition, without substantial task-specific architecture modifications.

BERT's bidirectional nature, enabled by the Transformer's self-attention mechanism, allows it to capture context from both directions, leading to a more nuanced understanding of language compared to previous unidirectional models. This bidirectional context is particularly crucial for tasks that require a deep understanding of language semantics and context.

The success of BERT has spawned numerous variants and improvements, such as RoBERTa, ALBERT, and T5, each pushing the boundaries of what's possible in NLP. These models have not only advanced the state-of-the-art in various NLP benchmarks but have also paved the way for more efficient and effective transfer learning in language understanding tasks.

\textbf{GPT and the Rise of Generative Models:} Generative Pre-trained Transformers (GPT) introduced a new paradigm for language generation, where pre-trained models could be fine-tuned on specific tasks. This paved the way for GPT-3 and similar models, which exhibited unprecedented language generation capabilities. LLMs like these are foundational to MLLMs, which extend similar architectures to multimodal tasks. The Generative Pre-trained Transformer (GPT) series, introduced by OpenAI, represents a significant milestone in the evolution of language models and generative AI. These models, built upon the Transformer architecture, have progressively pushed the boundaries of natural language processing and generation capabilities.

In 2018, GPT demonstrated the effectiveness of unsupervised pre-training on large corpora of text, followed by fine-tuning. With this approach, the model acquired a broad understanding of language structure and content, allowing it to be adapted to various downstream tasks with minimal task-specific training. The subsequent iterations, GPT-2 and GPT-3, marked substantial advancements in scale and capability. GPT-3, in particular, with its 175 billion parameters, exhibited remarkable few-shot and zero-shot learning abilities across a wide range of language tasks. This model demonstrated an unprecedented capacity for task generalization, often performing competitively with fine-tuned models without any task-specific training.

Several years after the success of the GPT series, NLP applications and research have witnessed a paradigm shift. Key aspects of this shift include:
\begin{enumerate}
    \item \textbf{Scale as a Driver of Capability:} The GPT models demonstrated that increasing model size and training data volume could lead to qualitative improvements in language understanding and generation.
    \item \textbf{Emergence of In-context Learning:} Large language models like GPT-3 exhibited the ability to perform tasks based solely on instructions or examples provided in the input prompt, without fine-tuning.
    \item \textbf{Versatility and Adaptability:} These models showed proficiency in a diverse array of tasks, from translation and summarization to code generation and creative writing.
    \item \textbf{Ethical and Societal Implications:} The powerful capabilities of these models raised important questions about their potential impact on society, including concerns about misinformation, bias, and the future of human-AI interaction.
\end{enumerate}

With the advent of GPT and similar models, multimodal systems and language models have become even more advanced. The line between different types of AI systems has further blurred with subsequent developments like GPT-4, which incorporates multimodal capabilities.

These advancements in generative models have profound implications for MLLMs. The principles of large-scale pre-training, the ability to understand and generate human-like text, and the capacity for few-shot learning are all crucial elements that MLLMs build upon. By extending these capabilities to include visual inputs, MLLMs represent the next frontier in AI's ability to understand and interact with the world in a more human-like manner.

\section{Architecture of MLLMs}


The architecture of Multimodal Large Language Models (MLLMs) represents a significant advancement in artificial intelligence, combining the strengths of language models with the ability to process and understand visual information. This integration allows MLLMs to perform complex tasks that require reasoning across both textual and visual domains.

At its core, the architecture of MLLMs is designed to bridge the gap between language understanding and visual perception. This is achieved through a carefully crafted combination of neural network components, each specialized for different aspects of multimodal processing. The key components of MLLM architecture typically include several interconnected elements working in harmony.

A crucial part of the MLLM architecture is the visual encoder, responsible for processing and encoding visual inputs. This component often utilizes convolutional neural networks (CNNs) or more advanced architectures like Vision Transformers (ViT) to extract meaningful features from images or video frames. Working alongside the visual encoder is the language encoder, similar to traditional language models. This component processes textual inputs and is typically based on the Transformer architecture, allowing for contextual understanding of text.

Central to the MLLM's ability to handle multimodal data is the multimodal fusion module. This critical component integrates the information from both visual and textual modalities, creating a unified representation that captures the relationships between visual and linguistic features. Complementing the fusion module are cross-modal attention mechanisms, which enable the model to attend to relevant parts of one modality while processing the other. For instance, when answering a question about an image, the model can focus on specific image regions that are most relevant to the question.

Finally, the decoder generates outputs based on the fused multimodal representations. Depending on the task, this could involve generating text, classifying images, or producing other forms of multimodal output. The seamless integration of these components allows MLLMs to perform a wide range of tasks that require joint understanding of visual and textual information. This architectural design enables MLLMs to not only process each modality independently but also to reason about the relationships between them, leading to more sophisticated and human-like understanding of multimodal inputs.
\\
\textbf{Transformer Backbone:} 
\begin{figure}
    \centering
    \includegraphics[width=0.6\linewidth]{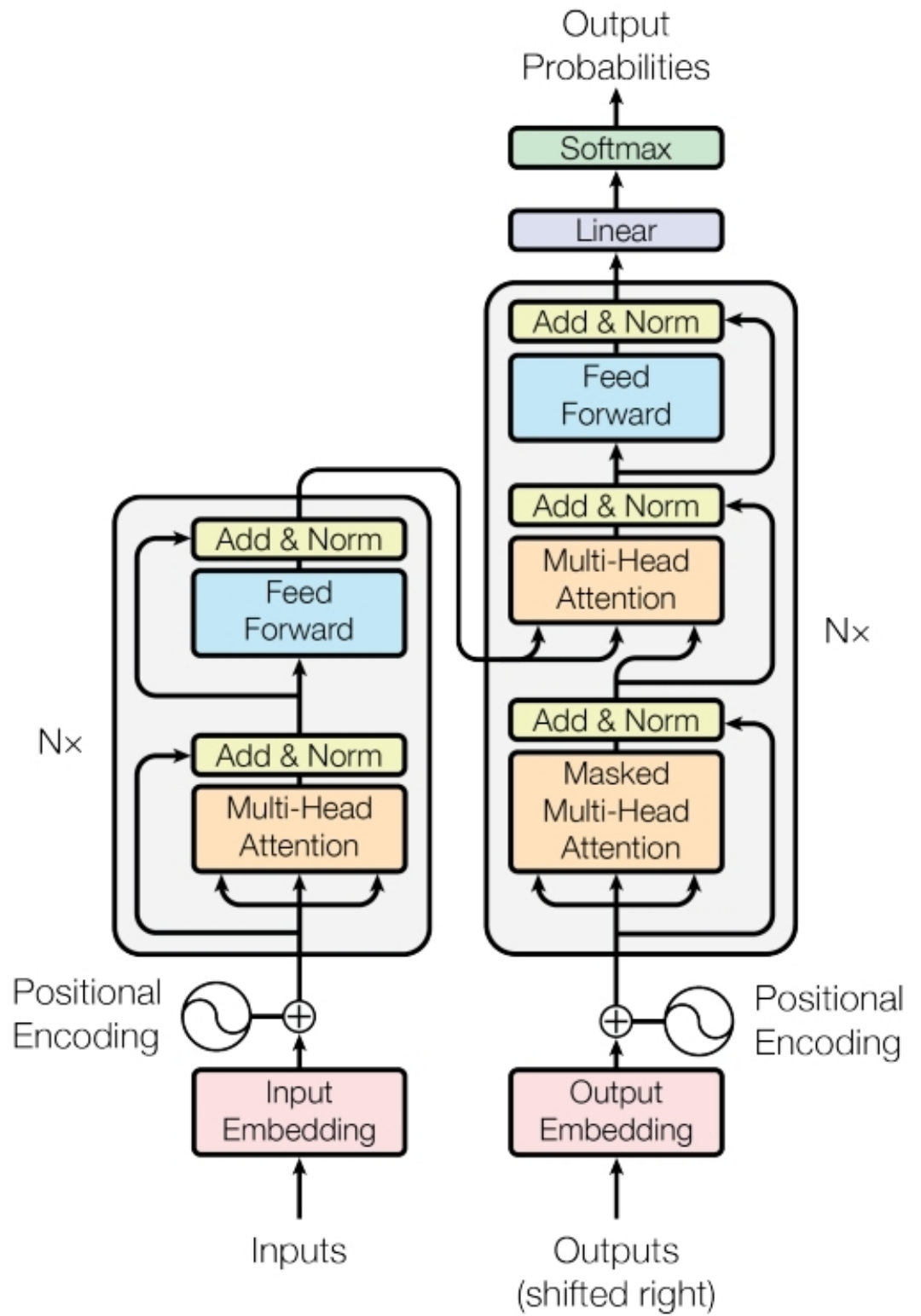}
    \caption{Transformer Backbone}
    \label{fig:transformer1}
\end{figure}
\\
MLLMs typically rely on the Transformer architecture, which uses self-attention to understand the relationships between data elements, whether they are words or image patches. The Transformer model’s ability to handle sequential and parallel information makes it ideal for multimodal tasks.

The Transformer architecture, introduced by \cite{vaswani2017attention}, serves as the fundamental backbone for MLLMs, providing a powerful mechanism for processing both textual and visual information. This architecture's core strength lies in its self-attention mechanism, which allows the model to dynamically focus on relevant parts of the input, regardless of their position in the sequence.

MLLMs benefit greatly from the Transformer's versatility. Multimodal tasks benefit from its ability to handle both sequential and parallel data (like text). In order to understand the nuanced relationships between visual and textual elements, the self-attention mechanism allows the model to capture long-range dependencies and complex relationships.

The Transformer's architecture typically consists of multiple layers of self-attention and feed-forward neural networks. In MLLMs, this structure is often adapted to include:
\begin{enumerate}
    \item \textbf{Encoder layers:} These process the input data, whether it's text tokens or visual features, and create contextualized representations.
    \item \textbf{Decoder layers:} These generate output based on the encoded representations, often incorporating cross-attention mechanisms to attend to relevant parts of the input.
    \item \textbf{Multi-head attention:} This allows the model to attend to different aspects of the input simultaneously, enhancing its ability to capture diverse relationships in multimodal data.
    \item \textbf{Position encodings:} These are crucial for maintaining spatial or sequential information, especially important when dealing with image patches or text sequences.
\end{enumerate}

Scalability is another factor that contributes to the Transformer's popularity among MLLMs. As demonstrated by models like GPT-3 and CLIP, increasing the size of Transformer-based models often leads to improved performance and generalization capabilities. This scalability allows MLLMs to leverage vast amounts of multimodal data effectively, leading to more robust and versatile models.

Furthermore, the Transformer's architecture facilitates efficient parallel processing, making it well-suited for handling the computational demands of large-scale multimodal tasks. This efficiency is crucial when processing high-dimensional inputs like images alongside text.

\textbf{Multimodal Embedding:} 
\begin{figure}
    \centering
    \includegraphics[width=1.0\linewidth]{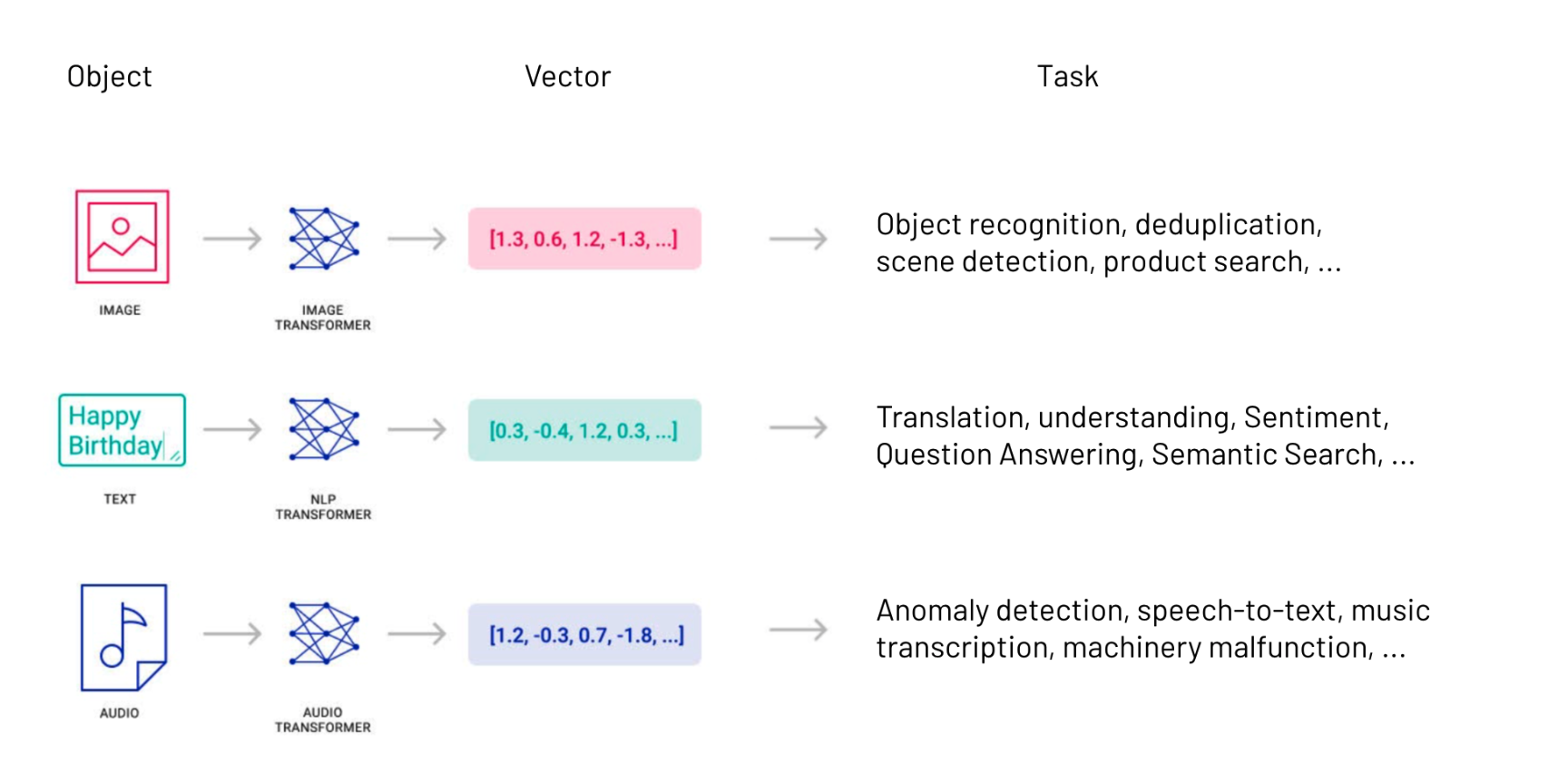}
    \caption{Multimodal Embeddings}
    Source: \url{https://www.twelvelabs.io/blog/multimodal-embeddings}
    \label{fig:multimodal_embeddings}
\end{figure}
One of the key features of MLLMs is their ability to embed both text and visual information into a unified space. This allows the model to reason about both modalities in a consistent manner, ensuring that relationships between visual and textual data can be leveraged effectively.

Multimodal embedding is a crucial component in the architecture of MLLMs, enabling these models to represent and reason about diverse types of information within a unified computational framework. This technique involves projecting data from different modalities—primarily text and images in the context of MLLMs—into a shared, high-dimensional vector space. The resulting embeddings capture semantic relationships not only within each modality but also across modalities, facilitating more sophisticated cross-modal reasoning and analysis.

The process of creating multimodal embeddings typically involves several key steps:
\begin{enumerate}
    \item \textbf{Modality-Specific Encoding:} Initially, each input modality is processed through specialized encoders. For textual data, this often involves tokenization followed by embedding through techniques like word2vec or more advanced contextual embedding methods. Visual data is typically encoded using convolutional neural networks (CNNs) or Vision Transformers (ViT) to extract salient features.
    \item \textbf{Dimensionality Alignment:} The embeddings from different modalities are often of different dimensionalities. A crucial step is to project these embeddings into a common dimensional space, usually through learnable linear transformations or more complex neural network layers.
    \item \textbf{Joint Representation Learning:} The aligned embeddings are then further processed to create a truly joint representation. This often involves attention mechanisms or fusion layers that allow the model to learn complex interactions between the modalities.
    \item \textbf{Contrastive Learning:} Many state-of-the-art MLLMs employ contrastive learning techniques during training. This approach encourages the model to produce similar embeddings for semantically related text-image pairs while pushing apart unrelated pairs in the embedding space.
    \item \textbf{Fine-tuning for Downstream Tasks:} The joint embeddings are then fine-tuned on specific downstream tasks, allowing the model to adapt its representations for particular applications while retaining the general cross-modal understanding gained during pretraining.
\end{enumerate}

\textbf{Create Multimodal Embeddings} 
\begin{figure}
    \centering
    \includegraphics[width=0.25\linewidth]{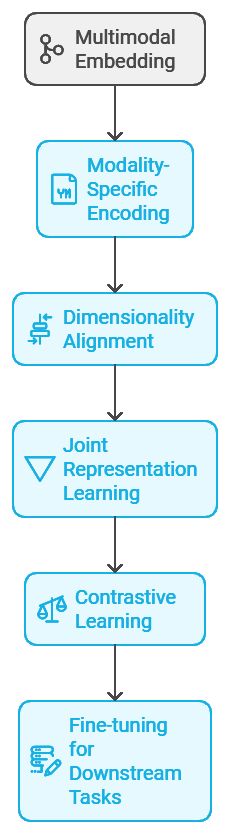}
    \caption{Create Multimodal Embeddings}
    \label{fig:Embeddings}
\end{figure}

The effectiveness of multimodal embeddings in MLLMs is evidenced by their performance on various tasks requiring cross-modal understanding. For instance, in image-text retrieval tasks, these embeddings enable the model to find semantically relevant images given a textual query, and vice versa. In visual question answering, the joint embedding space allows the model to reason about textual questions in the context of visual information.

Moreover, recent research has shown that multimodal embeddings can capture nuanced relationships between concepts across modalities. For example, they can represent abstract concepts that are difficult to visualize directly but are often associated with certain visual patterns or contexts. This capability enables MLLMs to perform more human-like reasoning tasks, such as visual commonsense inference or generating creative descriptions of images.

However, creating effective multimodal embeddings also presents several challenges. These include dealing with the semantic gap between modalities, handling the different statistical properties of visual and textual data, and ensuring that the embeddings generalize well across diverse tasks and domains. Ongoing research in this area focuses on developing more sophisticated embedding techniques, exploring ways to incorporate additional modalities, and improving the interpretability of these high-dimensional representations.

\textbf{Cross-Attention Layers:} To enable interaction between text and images, MLLMs often employ cross-attention mechanisms. These layers allow the model to focus on relevant parts of an image when processing a piece of text (or vice versa), enhancing the ability to understand the relationships between the two.

Cross-attention layers are a fundamental component of MLLMs, facilitating the intricate interplay between textual and visual modalities. These layers enable the model to dynamically focus on relevant aspects of one modality while processing information from the other, thereby enhancing the model's capacity to understand and reason about multimodal inputs.

The mechanism of cross-attention is derived from the self-attention concept introduced in the Transformer architecture. However, unlike self-attention, which operates within a single modality, cross-attention allows the model to attend to information across different modalities. In the context of MLLMs, this typically involves attention between textual and visual representations.

The cross-attention process can be formalized as follows:

Let Q represent the query vectors from one modality (e.g., text), and K and V represent the key and value vectors from another modality (e.g., image). The cross-attention operation can be expressed as:
\begin{equation}
    Attention(Q, K, V) = softmax(Q{K^T} / \sqrt{d_k})V.
\end{equation}
Where $d_k$ is the dimensionality of the key vectors, and the softmax operation is applied row-wise.

This formulation allows the model to compute attention weights that determine the relevance of each element in one modality to each element in the other modality. For instance, when processing a textual query about an image, the cross-attention layer enables the model to focus on specific regions of the image that are most pertinent to the query.

The benefits of cross-attention layers in MLLMs are multifold:
\begin{enumerate}
    \item \textbf{Fine-grained multimodal alignment:} Cross-attention facilitates precise alignment between elements of different modalities, allowing the model to capture nuanced relationships between specific words and image regions.
    \item \textbf{Contextual understanding:} By attending to relevant parts of one modality while processing the other, the model can develop a more contextual and holistic understanding of the multimodal input.
    \item \textbf{Flexibility in handling varying input sizes:} Cross-attention can naturally handle inputs of different lengths or sizes, making it suitable for processing variable-length text and images of different resolutions.
    \item \textbf{Improved interpretability:} The attention weights produced by cross-attention layers can be visualized, providing insights into which parts of an image the model focuses on when processing specific textual inputs, and vice versa.
\end{enumerate}

\textbf{Benefits of Cross-attention Layer} 
\begin{figure}
    \centering
    \includegraphics[width=0.6\linewidth]{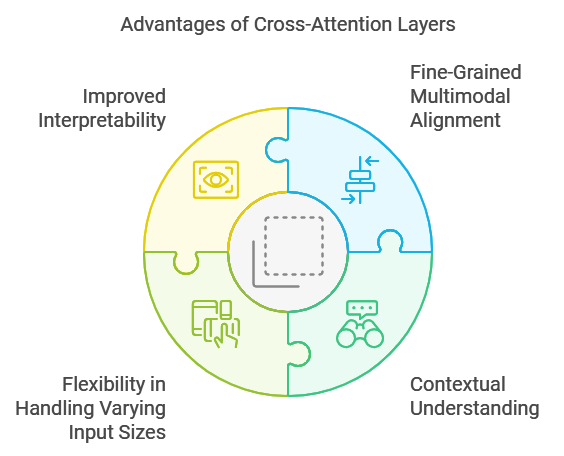}
    \caption{Benefits of Cross-attention Layer}
    \label{fig:Embeddings1}
\end{figure}

Recent advancements in cross-attention mechanisms for MLLMs include the development of more efficient attention computations to handle large-scale inputs, the incorporation of multi-head cross-attention for capturing diverse cross-modal relationships, and the exploration of hierarchical cross-attention structures to model interactions at different levels of abstraction.

\textbf{Vision Encoders and Language Decoders:} MLLMs typically include separate encoders for visual inputs (often based on CNNs or Vision Transformers) and text inputs. These encoders transform each input modality into a shared latent space, which is then processed by a unified model to perform tasks like captioning, question answering, or retrieval.

Vision Encoders and Language Decoders are fundamental components of MLLMs, serving as the interface between raw input data and the model's internal representations. These specialized modules are designed to process and transform visual and textual information, respectively, enabling the model to operate on a unified representation space.

Vision Encoders are responsible for extracting salient features from visual inputs. Traditionally, Convolutional Neural Networks (CNNs) have been the predominant architecture for this task. CNNs excel at capturing hierarchical visual features, from low-level edges and textures to high-level semantic concepts. Notable CNN architectures employed in MLLMs include ResNet, Inception, and EfficientNet. These networks typically consist of multiple convolutional layers, pooling operations, and non-linear activations, culminating in a dense representation of the input image.

More recently, Vision Transformers (ViT) have emerged as a powerful alternative to CNNs for visual encoding. Introduced by \cite{dosovitskiy2020image}, ViTs adapt the transformer architecture, originally designed for sequence modeling, to image processing. In a ViT, an image is divided into fixed-size patches, which are linearly embedded and treated as a sequence of tokens. This approach leverages the self-attention mechanism to capture global dependencies in the image, potentially offering advantages over the local receptive fields of CNNs.

The output of the Vision Encoder is typically a set of feature vectors or a single aggregated vector representing the entire image. This representation is then projected into a shared embedding space that is compatible with the textual representations.

Language Decoders, on the other hand, are responsible for generating coherent and contextually appropriate textual outputs based on the model's internal representations. In MLLMs, these decoders often employ transformer-based architectures, leveraging the power of self-attention and cross-attention mechanisms.

The Language Decoder operates autoregressively, generating text one token at a time. At each step, it attends to both the previously generated tokens and the visual representations provided by the Vision Encoder. This cross-modal attention allows the decoder to ground its textual outputs in the visual context, enabling tasks such as image captioning or visual question answering.

A key feature of modern Language Decoders in MLLMs is their ability to handle prompts or queries. By conditioning the generation process on a given prompt, these decoders can produce responses that are not only relevant to the visual input but also tailored to specific instructions or questions.

The interplay between Vision Encoders and Language Decoders is crucial for the performance of MLLMs. The quality of visual features extracted by the encoder directly impacts the decoder's ability to generate accurate and relevant textual outputs. Similarly, the decoder's capacity to effectively utilize these visual features determines the model's overall cross-modal understanding and generation capabilities.

Recent advancements in this domain include the development of more efficient encoder-decoder architectures, such as the use of sparse attention mechanisms to handle longer sequences, and the exploration of alternative visual encoding strategies like implicit neural representations. These innovations aim to enhance the model's ability to capture fine-grained visual details while maintaining computational efficiency.

\section{Training Methodologies and Data Requirements}
Training Methodologies and Data Requirements for Multimodal Large Language Models (MLLMs) encompass a complex and multifaceted process that demands careful consideration of various factors. This section delves into the intricacies of training these sophisticated models, elucidating the methodologies employed and the stringent data requirements necessary for their development.

The pretraining phase is crucial for MLLMs, as it lays the foundation for the model's ability to understand and generate multimodal content. Several strategies have been developed to optimize this process. Contrastive learning has emerged as a powerful technique, where the model learns to distinguish between related and unrelated image-text pairs. This approach enables the development of a nuanced understanding of the relationship between visual and textual modalities. Masked language modeling, adapted from text-only models, involves randomly masking tokens in the input text and training the model to predict these masked tokens. In the multimodal context, this technique is extended to incorporate visual information, allowing the model to leverage visual cues in predicting masked textual content. Additionally, image-text matching tasks encourage the model to develop a holistic understanding of both modalities and their interrelations.

After pretraining, MLLMs undergo fine-tuning to adapt to specific downstream tasks. This process involves several considerations, including task-specific adaptation tailored to the requirements of the target task. For instance, fine-tuning for visual question answering may involve training the model on question-answer pairs associated with images, while image captioning fine-tuning focuses on generating descriptive text for given images. Recent advancements have explored few-shot and zero-shot learning capabilities, aiming to minimize the need for extensive task-specific fine-tuning by leveraging the model's pretrained knowledge to perform new tasks with minimal or no additional training examples.

The quality and scale of training data significantly impact the performance of MLLMs. These models typically require massive datasets comprising millions to billions of image-text pairs. For instance, the CLIP model was trained on a dataset of 400 million image-text pairs, while more recent models have utilized even larger datasets. To ensure robust performance across various domains and tasks, training data must exhibit substantial diversity in both visual and textual content. This includes diversity in visual elements such as different object types, scenes, and artistic styles, as well as textual diversity in languages, writing styles, and topic areas.

The accuracy of image-text alignments in the training data is paramount. Misaligned pairs can introduce noise into the training process, potentially leading to erroneous associations between visual and textual elements. Consequently, significant effort is invested in data cleaning and validation processes to ensure high-quality, well-aligned datasets.

Training MLLMs is computationally intensive, often requiring distributed training across multiple high-performance GPUs or TPUs. The scale of computational resources needed has implications for both the environmental impact of AI research and the accessibility of MLLM development to different research groups.

Ethical considerations must also be accounted for in the training process of MLLMs. Steps must be taken to identify and mitigate biases present in the training data, which could otherwise be perpetuated or amplified by the model. When training on large-scale datasets scraped from the internet, care must be taken to respect privacy rights and comply with data protection regulations.

Training MLLMs is a computationally intensive process, and several factors must be considered to ensure effective learning:

\textbf{Pretraining on Multimodal Data:} MLLMs are often pretrained on large-scale multimodal datasets that include aligned pairs of text and images. This allows the model to learn correlations between words and visual features, a crucial aspect for downstream tasks.

Pretraining on multimodal data is a critical phase in the development of MLLMs, serving as the foundation for their ability to understand and generate content across different modalities. This process involves exposing the model to vast amounts of paired visual and textual data, enabling it to learn rich, joint representations of both modalities.

The pretraining process typically employs several key strategies:
\begin{enumerate}
    \item \textbf{Contrastive Learning:} This approach involves training the model to distinguish between related and unrelated image-text pairs. The model learns to maximize the similarity between matching pairs while minimizing it for non-matching ones. This technique helps the model develop a nuanced understanding of the relationships between visual and textual elements.
    \item \textbf{Masked Language Modeling (MLM):} Adapted from text-only models, MLM involves randomly masking tokens in the input text and training the model to predict these masked tokens. In the multimodal context, this technique is extended to incorporate visual information, allowing the model to leverage visual cues in predicting masked textual content.
    \item \textbf{Masked Image Modeling (MIM):} Similar to MLM, this technique involves masking portions of the input image and training the model to reconstruct or predict the masked regions. This encourages the model to develop a deeper understanding of visual structures and their relationships to textual descriptions.
    \item \textbf{Image-Text Matching:} The model is trained to determine whether a given image-text pair is matching or not. This task promotes the development of a holistic understanding of both modalities and their interrelations.
    \item \textbf{Visual Grounding:} This involves training the model to locate specific objects or regions in an image based on textual descriptions. This technique enhances the model's ability to align visual and textual information at a fine-grained level.
\end{enumerate}

The choice of pretraining dataset is crucial and often includes large-scale, web-scraped collections such as Conceptual Captions, LAION-5B, or JFT-300M. These datasets provide diverse, real-world examples of image-text pairs, enabling the model to learn robust, generalizable representations.

There are several challenges associated with pretraining on multimodal data, including optimizing the processing of high-dimensional visual data, handling potential misalignments in image-text pairs, and balancing learning objectives across modes. Recent advancements have focused on developing more efficient architectures and training strategies to address these challenges, such as the use of vision transformers for improved visual processing and the development of more sophisticated loss functions that better capture cross-modal relationships.


\textbf{Fine-Tuning for Specific Tasks:} After pretraining, MLLMs are fine-tuned on task-specific datasets. For example, models may be fine-tuned on datasets for image captioning, visual question answering, or cross-modal retrieval. This process ensures that the model can transfer its learned multimodal representations to practical applications.

Fine-tuning for specific tasks is a crucial phase in the development of MLLMs, allowing these models to adapt their pretrained knowledge to particular downstream applications. This process involves several key considerations and techniques:
\begin{enumerate}
    \item \textbf{Task-Specific Adaptation:} Fine-tuning tailors the model's capabilities to the requirements of the target task. For instance, fine-tuning for visual question answering (VQA) involves training the model on question-answer pairs associated with images, while image captioning fine-tuning focuses on generating descriptive text for given images.
    \item \textbf{Transfer Learning:} Fine-tuning leverages the knowledge acquired during pretraining, allowing the model to transfer its learned representations to new tasks. This approach significantly reduces the amount of task-specific data required and accelerates the learning process.
    \item \textbf{Architectural Modifications:} Depending on the task, the model architecture may be slightly modified during fine-tuning. For example, additional task-specific layers might be added to the output of the pretrained model to facilitate task-specific predictions.
    \item \textbf{Hyperparameter Optimization:} Fine-tuning often involves careful tuning of hyperparameters such as learning rate, batch size, and number of training epochs. These parameters can significantly impact the model's performance on the specific task.
    \item \textbf{Few-Shot and Zero-Shot Learning:} Recent advancements have explored fine-tuning techniques that minimize the need for extensive task-specific data. Few-shot learning aims to adapt the model using only a small number of examples, while zero-shot learning leverages the model's pretrained knowledge to perform new tasks without any additional training examples.
    \item \textbf{Continual Learning:} Some fine-tuning approaches focus on enabling the model to learn new tasks without forgetting previously learned information, a challenge known as catastrophic forgetting.
    \item \textbf{Multi-Task Fine-Tuning:} In some cases, models are fine-tuned on multiple related tasks simultaneously, which can lead to improved performance across all tasks due to shared learning.
    \item \textbf{Domain Adaptation:} When the target task involves a domain significantly different from the pretraining data, fine-tuning may include techniques to bridge this domain gap, such as gradual fine-tuning or domain-adversarial training.
\end{enumerate}

The effectiveness of fine-tuning is often evaluated through careful benchmarking on task-specific datasets and comparison with specialized models. As research progresses, more sophisticated fine-tuning techniques are being developed to enhance the adaptability and performance of MLLMs across a wide range of multimodal tasks.

\textbf{Data Scale and Quality:} The success of MLLMs hinges on the availability of large, high-quality datasets. The most advanced models are trained on datasets like Common Crawl (for text) and Conceptual Captions (for images), which contain billions of image-text pairs. However, curating and annotating such datasets is an enormous challenge.

The scale and quality of data used in training MLLMs are paramount to their performance and capabilities. These factors significantly influence the model's ability to learn robust representations and generalize across diverse tasks.

\textbf{Data Scale:} The sheer volume of data required for training MLLMs is staggering. State-of-the-art models often utilize datasets comprising billions of image-text pairs. This massive scale is necessary to capture the complexity and diversity of multimodal relationships. Large-scale datasets like Common Crawl for text and Conceptual Captions for images have become standard in the field. The scale of data serves several crucial purposes:
\begin{enumerate}
    \item \textbf{Comprehensive Coverage:} Vast datasets increase the likelihood of exposing the model to a wide range of concepts, contexts, and relationships between visual and textual elements.
    \item \textbf{Improved Generalization:} Larger datasets help mitigate overfitting and enhance the model's ability to generalize to unseen examples and tasks.
    \item \textbf{Rare Instance Learning:} With increased scale, the model has more opportunities to encounter and learn from rare or uncommon instances, which is crucial for robust performance in real-world applications.
\end{enumerate}

\textbf{Data Quality:} While scale is important, the quality of the data is equally critical. High-quality data ensures that the model learns meaningful and accurate representations. Several aspects of data quality are considered:
\begin{enumerate}
    \item \textbf{Relevance:} The data should be relevant to the intended applications of the MLLM. This often necessitates careful curation of web-scraped datasets to remove irrelevant or inappropriate content.
    \item \textbf{Accuracy:} The alignment between image and text pairs must be accurate. Misaligned pairs can introduce noise into the learning process and hinder the model's performance.
    \item \textbf{Diversity:} The dataset should represent a diverse range of concepts, styles, and domains to ensure the model's versatility.
    \item \textbf{Cleanliness:} Data cleaning processes are essential to remove duplicates, handle missing values, and correct errors that could negatively impact training.
    \item \textbf{Ethical Considerations:} High-quality datasets must also adhere to ethical standards, respecting privacy, copyright, and avoiding biases that could lead to unfair or discriminatory model behavior.
\end{enumerate}

\textbf{Challenges in Data Curation:} Despite the critical importance of data scale and quality, curating suitable datasets for MLLMs presents significant challenges:

\begin{enumerate}
    \item \textbf{Resource Intensity:} Collecting and processing billions of image-text pairs requires substantial computational and human resources.
    \item \textbf{Annotation Complexity:} For tasks requiring fine-grained annotations, such as object detection or segmentation, the annotation process becomes increasingly complex and time-consuming.
    \item \textbf{Quality Control:} Maintaining consistent quality across a large-scale dataset is challenging, often requiring multiple rounds of validation and refinement.
    \item \textbf{Legal and Ethical Considerations:} Navigating the legal landscape of data usage, especially for web-scraped content, presents ongoing challenges. Ensuring ethical data collection and usage practices adds another layer of complexity.
    \item \textbf{Domain Specificity:} For certain applications, general web-scraped datasets may not suffice, necessitating the creation of domain-specific datasets, which can be particularly resource-intensive.
\end{enumerate}

In conclusion, while the scale and quality of data are crucial for the success of MLLMs, they also present significant challenges in terms of collection, curation, and maintenance. Ongoing research in this area focuses on developing more efficient data collection methodologies, improving data quality assessment techniques, and exploring ways to maximize the utility of available data resources.

\textbf{Challenges of Data Alignment:} One of the biggest challenges in training MLLMs is aligning the text and visual data correctly. Mismatches between image and text pairs can hinder learning and lead to poor model performance.

The alignment of text and visual data in the training of Multimodal Large Language Models (MLLMs) presents a complex and multifaceted challenge that significantly impacts the efficacy and reliability of these models. This challenge encompasses several critical aspects:

\begin{enumerate}
    \item \textbf{Semantic Coherence:} Ensuring that the textual descriptions accurately reflect the content of the corresponding images is paramount. Misalignments in this regard can lead to the model learning incorrect associations, potentially compromising its ability to generate accurate cross-modal representations.
    \item \textbf{Contextual Relevance:} The textual data must not only describe the image accurately but also capture the relevant context. This is particularly challenging when dealing with images that contain multiple elements or complex scenes, where the importance of different components may vary based on the intended application of the MLLM.
    \item \textbf{Temporal Consistency:} In cases where the training data includes temporal sequences (e.g., video frames with accompanying narration), maintaining alignment across time becomes crucial. Ensuring that the textual descriptions synchronize with the visual content as it evolves presents additional complexities.
    \item \textbf{Cultural and Linguistic Nuances:} The alignment challenge is further compounded when dealing with data from diverse cultural and linguistic backgrounds. Nuances in language and cultural interpretations of visual content can lead to subtle misalignments that are difficult to detect and correct at scale.
    \item \textbf{Ambiguity and Subjectivity:} Many images can be interpreted in multiple ways, and textual descriptions may reflect subjective perspectives. Balancing these subjective elements while maintaining overall alignment is a delicate task that requires careful consideration.
    \item \textbf{Scale-induced Errors:} As the scale of the dataset increases, the probability of alignment errors also grows. Detecting and correcting these errors in massive datasets becomes increasingly challenging and resource-intensive.
    \item \textbf{Automated Alignment Techniques:} While automated methods for aligning text and images (e.g., using computer vision and natural language processing techniques) can help scale the process, they are not infallible. These methods may introduce systematic biases or fail to capture nuanced relationships between visual and textual elements.
    \item \textbf{Quality-Quantity Trade-off:} There is often a tension between the desire for large-scale datasets and the need for high-quality alignments. Striking the right balance between these competing factors is crucial for developing robust MLLMs.
    \item \textbf{Domain-specific Challenges:} Different domains (e.g., medical imaging, satellite imagery) may present unique alignment challenges due to specialized vocabulary or the need for expert knowledge to accurately describe visual content.
\end{enumerate}

Addressing these challenges requires a multifaceted approach, combining advanced machine learning techniques, human expertise, and rigorous quality control processes. Ongoing research in this area focuses on developing more sophisticated alignment algorithms, improving data curation methodologies, and exploring ways to leverage semi-supervised and unsupervised learning techniques to mitigate the impact of alignment errors.

\section{Cross-Modal Understanding and Visual Reasoning}

MLLMs are particularly adept at cross-modal understanding, where the model integrates knowledge from both language and vision to perform tasks that require reasoning across modalities. Visual reasoning is a key capability that allows MLLMs to interpret images in the context of textual queries or instructions.

The cross-modal understanding and visual reasoning capabilities of MLLMs represent a significant advancement in artificial intelligence, bridging the gap between language and vision in ways that were previously challenging for machines. These models demonstrate an impressive ability to not only process but also integrate and reason across different modalities, particularly text and images. This integration allows for a more holistic understanding of complex scenarios, mirroring human-like cognitive processes.

Key aspects of cross-modal understanding and visual reasoning in MLLMs include:
\begin{enumerate}
    \item \textbf{Semantic Integration:} MLLMs excel at combining semantic information from both textual and visual inputs. This allows them to form a comprehensive understanding of a given scenario, where the meaning derived from one modality complements and enhances the interpretation of the other.
    \item \textbf{Contextual Interpretation:} These models can interpret visual elements in the context of textual information and vice versa. This contextual understanding enables more nuanced and accurate responses to queries that span both modalities.
    \item \textbf{Abstract Reasoning:} MLLMs demonstrate the ability to perform abstract reasoning tasks that require synthesizing information across modalities. This includes drawing inferences, making predictions, and understanding implicit relationships not explicitly stated in either the text or the image.
    \item \textbf{Flexible Query Handling:} These models can process and respond to a wide range of query types, from simple object identification to complex scenarios requiring multi-step reasoning. This flexibility makes them versatile tools for various applications.
    \item \textbf{Generalization Capabilities:} Well-trained MLLMs can generalize their understanding to novel combinations of visual and textual inputs, showcasing their ability to apply learned concepts in new contexts.
\end{enumerate}

The development of these capabilities in MLLMs opens up numerous possibilities for advanced applications in fields such as robotics, autonomous systems, medical diagnosis, and educational technologies, where the ability to reason across different types of data is crucial.

\textbf{Visual Question Answering (VQA):} In VQA tasks, the model must understand an image and answer questions about its content. This requires not just recognizing objects but also reasoning about their relationships and attributes based on the accompanying text.

Visual Question Answering (VQA) represents a complex and challenging task at the intersection of computer vision and natural language processing. It requires MLLMs to comprehend both visual and textual inputs, and to generate accurate, contextually relevant responses. The process involves several sophisticated steps:
\begin{enumerate}
    \item \textbf{Image Analysis:} The model must first analyze the given image, identifying objects, their attributes, spatial relationships, and other relevant visual features. This involves advanced computer vision techniques, including object detection, scene parsing, and attribute recognition.
    \item \textbf{Question Comprehension:} Simultaneously, the model must parse and understand the textual question. This involves natural language processing to discern the query's intent, identify key entities, and recognize the type of information being sought.
    \item \textbf{Cross-Modal Reasoning:} The MLLM then engages in complex reasoning that integrates the visual and textual information. This step may involve:
    \\
     (a) Aligning concepts between the two modalities.\\
   (b) Inferring implicit information not directly stated or shown.\\
   (c) Applying common sense knowledge to interpret the scene
    \item \textbf{Answer Generation:} Based on the integrated understanding and reasoning, the model generates an appropriate answer. This may be a simple word or phrase, or a more complex explanation, depending on the nature of the question.
    \item \textbf{Confidence Assessment:} Advanced VQA systems often include a mechanism to assess the confidence of the generated answer, which can be crucial in real-world applications where reliability is paramount.
\end{enumerate}

The complexity of VQA tasks can vary significantly, ranging from simple object identification questions to those requiring abstract reasoning or external knowledge. For instance:

\begin{enumerate}
    \item \textbf{Factual Questions:} "What color is the car?" requires basic object recognition and attribute identification.
    \item \textbf{Counting Questions:} "How many people are in the image?" necessitates accurate object detection and quantification.
    \item \textbf{Relational Questions:} "Is the cup on the table?" demands understanding of spatial relationships.
    \item \textbf{Inferential Questions:} "Why is the person smiling?" requires higher-level reasoning about emotions and context.
\end{enumerate}
VQA presents several unique challenges:

\begin{enumerate}
    \item \textbf{Ambiguity:} Images can be interpreted in multiple ways, and questions may have multiple valid answers.
    \item \textbf{Bias:} Training data may contain biases that affect the model's performance on certain types of questions or images.
    \item \textbf{Out-of-Distribution Generalization:} The model must handle questions and visual scenarios not encountered during training.
    \item \textbf{Explainability:} In many applications, it's crucial not just to answer correctly, but to explain the reasoning process.
\end{enumerate}

Ongoing research in VQA focuses on improving model architectures, developing more diverse and challenging datasets, and creating evaluation metrics that better capture the nuanced aspects of visual reasoning. As MLLMs continue to advance, their capacity for sophisticated VQA tasks is expected to play a crucial role in applications ranging from assistive technologies for the visually impaired to advanced human-computer interaction systems.

\textbf{Image Captioning:} In image captioning, the model generates textual descriptions of an image, demonstrating its ability to translate visual information into natural language.

Image Captioning is a sophisticated task that exemplifies the cross-modal capabilities of MLLMs, requiring the integration of computer vision and natural language processing. This task involves generating coherent and descriptive textual representations of visual content, demonstrating the model's ability to bridge the semantic gap between visual and linguistic modalities.

The process of image captioning typically encompasses several key steps:

\begin{enumerate}
    \item \textbf{Visual Feature Extraction:} The MLLM employs advanced convolutional neural networks (CNNs) or vision transformers to extract salient features from the input image. These features encapsulate information about objects, their attributes, spatial relationships, and overall scene composition.
    \item \textbf{Semantic Representation:} The extracted visual features are then mapped to a semantic space that aligns with the model's language understanding. This step is crucial for establishing a common ground between visual and textual information.
    \item \textbf{Language Generation:} Utilizing recurrent neural networks (RNNs), long short-term memory networks (LSTMs), or transformer-based architectures, the model generates a sequence of words that form a coherent caption. This process often employs attention mechanisms to focus on relevant parts of the image while generating each word.
    \item \textbf{Context Integration:} Advanced image captioning models incorporate contextual information, considering not just individual objects but also their interactions, the overall scene, and potential implications or actions depicted in the image.
    \item \textbf{Caption Refinement:} Some state-of-the-art approaches implement iterative refinement processes, where initial captions are generated and then improved through multiple passes, enhancing accuracy and descriptiveness.
\end{enumerate}

The complexity of image captioning extends beyond mere object identification, encompassing several challenging aspects:

\begin{enumerate}
    \item \textbf{Semantic Accuracy:} Captions must accurately reflect the content and context of the image, avoiding misinterpretations or omissions of crucial elements.
    \item \textbf{Linguistic Quality:} Generated captions should adhere to proper grammatical structures and exhibit natural language flow.
    \item \textbf{Relevance:} The caption should capture the most salient aspects of the image, discerning between central and peripheral elements.
    \item \textbf{Abstraction:} Advanced captioning models can generate descriptions that go beyond literal visual elements, inferring abstract concepts or potential narratives from the image.
\end{enumerate}

Evaluation of image captioning models typically employs metrics such as BLEU, METEOR, CIDEr, and SPICE, which assess both the semantic correctness and the linguistic quality of the generated captions. However, due to the subjective nature of caption quality, human evaluation often complements these automated metrics.

Recent advancements in image captioning research focus on:

\begin{enumerate}
    \item \textbf{Dense Captioning:} Generating multiple captions for different regions of an image, providing a more comprehensive description.
    \item \textbf{Stylized Captioning:} Adapting the caption style to specific tones or genres while maintaining accuracy.
    \item \textbf{Multilingual Captioning:} Extending captioning capabilities across multiple languages to enhance accessibility and global applicability.
\end{enumerate}

The development of sophisticated image captioning capabilities in MLLMs has significant implications for various applications, including assistive technologies for visually impaired individuals, content indexing and retrieval systems, and automated content description for social media platforms. As these models continue to evolve, they promise to bridge the gap between visual perception and linguistic expression in increasingly nuanced and human-like ways.

\textbf{Cross-Modal Retrieval:} MLLMs can retrieve relevant images based on a textual query or find corresponding text based on an image. This is a practical application in search engines and recommendation systems, where visual and textual content must be aligned.

Cross-Modal Retrieval is a sophisticated task that exemplifies the advanced capabilities of MLLMs in bridging the semantic gap between visual and textual modalities. This task involves the bidirectional retrieval of content across different modalities, typically text and images, demonstrating the model's ability to establish meaningful connections between diverse data types.

The process of cross-modal retrieval encompasses two primary scenarios:

\begin{enumerate}
    \item \textbf{Text-to-Image Retrieval:} In this scenario, the system retrieves relevant images based on a textual query. This requires the MLLM to comprehend the semantic content of the text and match it with the visual features of images in a database.
    \item \textbf{Image-to-Text Retrieval:} Conversely, this involves finding textual descriptions or documents that are semantically relevant to a given image query. The model must extract meaningful features from the image and align them with textual representations.
\end{enumerate}

The implementation of effective cross-modal retrieval systems involves several key components and challenges:

\begin{enumerate}
    \item \textbf{Feature Extraction and Representation:} MLLMs employ advanced techniques to extract salient features from both textual and visual inputs. For images, this often involves convolutional neural networks (CNNs) or vision transformers, while text processing may utilize transformer-based architectures like BERT or GPT.
    \item \textbf{Semantic Alignment:} A crucial aspect of cross-modal retrieval is the alignment of feature spaces across modalities. This is typically achieved through joint embedding spaces or shared latent representations that capture the semantic relationships between visual and textual content.
    \item \textbf{Similarity Measurement:} The system must define and compute similarity metrics that effectively capture the semantic relevance between items across different modalities. This often involves techniques such as cosine similarity, Euclidean distance, or learned similarity functions.
    \item \textbf{Scalability and Efficiency:} Cross-modal retrieval systems must be designed to handle large-scale datasets efficiently, often employing indexing techniques and approximate nearest neighbor search algorithms to enable fast retrieval.
    \item \textbf{Handling Semantic Ambiguity:} The model must address the inherent ambiguity in both language and visual interpretation, accounting for polysemy, context-dependent meanings, and varying levels of abstraction.
\end{enumerate}

Recent advancements in cross-modal retrieval research focus on several key areas:

\begin{enumerate}
    \item \textbf{Zero-shot and Few-shot Learning:} Developing models capable of retrieving content across modalities for categories or concepts not seen during training.
    \item \textbf{Attention Mechanisms:} Incorporating attention-based approaches to focus on the most relevant parts of images or text during the retrieval process.
    \item \textbf{Multi-modal Fusion:} Exploring advanced techniques for combining information from multiple modalities to enhance retrieval accuracy.
    \item \textbf{Explainable Retrieval:} Developing methods to provide interpretable explanations for why certain items were retrieved, enhancing user trust and system transparency.
    \item \textbf{Domain Adaptation:} Addressing the challenge of adapting cross-modal retrieval systems to new domains or languages with minimal additional training.
\end{enumerate}

The applications of cross-modal retrieval are diverse and impactful, including:

\begin{enumerate}
    \item Enhanced search engines that can retrieve images based on textual queries and vice versa.
    \item Content recommendation systems that leverage both visual and textual features to suggest relevant items.
    \item Accessibility tools that enable visually impaired users to find images based on textual descriptions.
    \item E-commerce platforms that allow users to search for products using either text or image inputs.
\end{enumerate}

\textbf{Visual Commonsense Reasoning:} This involves more advanced reasoning where the model must infer implicit information from an image, such as why an event is happening or what might happen next, based on both the visual content and a text-based query.

Visual Commonsense Reasoning (VCR) represents a sophisticated cognitive task that extends beyond mere object recognition or scene description. It requires MLLMs to demonstrate a nuanced understanding of visual scenes, incorporating contextual knowledge, causal relationships, and social dynamics. This advanced capability enables models to make inferences about implicit information, predict future events, and explain the rationale behind observed phenomena in images.

The process of Visual Commonsense Reasoning encompasses several key components:

\begin{enumerate}
    \item \textbf{Scene Understanding:} MLLMs must first comprehend the overall context of the image, including the spatial relationships between objects, the actions being performed, and the environmental setting.
    \item \textbf{Knowledge Integration:} The model needs to incorporate external knowledge about the world, including physical laws, social norms, and typical cause-effect relationships, to make informed inferences.
    \item \textbf{Temporal Reasoning:} VCR often involves understanding the temporal aspects of a scene, allowing the model to infer past events or predict future outcomes based on the current visual information.
    \item \textbf{Counterfactual Thinking:} Advanced VCR systems can engage in hypothetical reasoning, considering alternative scenarios or outcomes that deviate from the observed scene.
    \item \textbf{Multimodal Integration:} VCR tasks frequently require the seamless integration of visual and textual information, as queries or additional context may be provided in natural language.
\end{enumerate}

The implementation of VCR in MLLMs presents several challenges and areas of ongoing research:

\begin{enumerate}
    \item \textbf{Bias Mitigation:} Ensuring that the model's reasoning is not unduly influenced by dataset biases or stereotypes present in training data.
    \item \textbf{Explainability:} Developing methods to provide clear, human-interpretable explanations for the model's reasoning process and conclusions.
    \item \textbf{Abstraction and Generalization:} Enabling models to apply learned commonsense knowledge to novel situations or domains not encountered during training.
    \item \textbf{Handling Ambiguity:} Addressing scenarios where multiple interpretations or outcomes are plausible, and providing probabilistic or ranked responses.
    \item \textbf{Contextual Adaptation:} Adjusting reasoning based on cultural, geographical, or temporal contexts that may influence the interpretation of visual scenes.
\end{enumerate}

Recent advancements in VCR research focus on several key areas:

\begin{enumerate}
    \item \textbf{Neuro-symbolic Approaches:} Combining neural networks with symbolic reasoning systems to enhance logical inference capabilities.
    \item \textbf{Causal Reasoning:} Incorporating causal models to improve the understanding of cause-effect relationships in visual scenes.
    \item \textbf{Multi-task Learning:} Developing models that can perform various reasoning tasks simultaneously, leading to more robust and versatile systems.
    \item \textbf{Few-shot and Zero-shot VCR:} Enabling models to perform commonsense reasoning on novel concepts or scenarios with minimal or no specific training examples.
\end{enumerate}

The applications of Visual Commonsense Reasoning are diverse and hold significant potential across various domains:
- Autonomous Systems: Enhancing the decision-making capabilities of self-driving vehicles or robots in complex, real-world environments.
- Healthcare: Assisting in medical image analysis by inferring potential causes or outcomes based on visual symptoms.
- Education: Developing interactive learning tools that can explain complex concepts through visual reasoning.
- Security and Surveillance: Improving threat detection systems by reasoning about unusual or potentially dangerous situations.

Through Visual Commonsense Reasoning, we can bridge the gap between machine perception and human-like understanding, leading to more intelligent and contextually aware artificial intelligence systems.

\chapter{Training and Fine-Tuning Multimodal Large Language Models (MLLMs)}

Training Multimodal Large Language Models (MLLMs) involves a multifaceted approach that combines extensive pre-training with targeted fine-tuning to achieve optimal performance across various tasks \cite{yu2023scaling}. This chapter examines the strategies employed during the pre-training phase, underscores the significance of fine-tuning for specific applications, and discusses advanced methodologies such as few-shot and zero-shot learning. Furthermore, we explore instruction tuning—a contemporary technique that enhances MLLMs' proficiency in adhering to human-like instructions across diverse modalities \cite{xu2022multiinstruct, li2023otter}.

\section{Pre-Training Strategies}

MLLMs are based on pre-training. It involves training the model on vast multimodal datasets, typically consisting of paired text and image data. The goal of pre-training is to provide the model with a general understanding of language and visual representations that can later be adapted to specific tasks.

\subsection{Contrastive Learning (CLIP, ALIGN)}

Contrastive learning is a key method in training multimodal large language models (MLLMs) \cite{wang2024comprehensive}. Here are some important methods and insights related to contrastive learning in this context:

\textbf{Basic Concept}: Contrastive learning involves training models to differentiate between similar and dissimilar pairs of data. In the case of MLLMs, this often means aligning text and image pairs while distinguishing them from mismatched pairs. This approach helps in creating shared embedding spaces for different modalities, which is crucial for tasks like cross-modal retrieval \cite{cite4}.

\textbf{Hallucination Augmented Contrastive Learning}: This method introduces the concept of hallucination, where additional synthetic data points are generated to enhance the contrastive learning process. This approach aims to improve the model's robustness and generalization capabilities, especially in zero-shot scenarios where the model needs to perform tasks it hasn't been explicitly trained on \cite{cite1, cite2}.

\textbf{Img-Diff: Contrastive Data Synthesis}: This technique involves creating a novel dataset that enhances the quality of contrastive learning by synthesizing new data points. The Img-Diff dataset, for instance, focuses on improving the quality of multimodal data, which is essential for the effective training of high-performance MLLMs \cite{cite3}.

\textbf{Integration with Other Techniques}: Contrastive learning is often combined with other methods like masked language modeling and visual question answering to enhance the model's understanding of multimodal data. This integration helps in building robust models that can handle a wide range of tasks across different modalities \cite{cite4}.

These methods highlight the innovative approaches being used to enhance the capabilities of MLLMs through contrastive learning, ensuring they can effectively process and understand both text and visual information.

CLIP, developed by OpenAI, utilizes a contrastive learning method where the model learns to associate images with their corresponding text descriptions while distinguishing them from unrelated pairs. This approach allows CLIP to perform zero-shot learning, enabling it to generalize to tasks it wasn't explicitly trained on, such as recognizing objects in images without having seen labeled examples of those objects during training.

Similarly, ALIGN, developed by Google, aligns images and text by learning joint embeddings from large-scale noisy data. ALIGN is designed to handle massive datasets and is robust to noise in the training data, making it highly scalable. Like CLIP, ALIGN also demonstrates strong zero-shot performance, enabling it to perform well on a variety of tasks without specific task-based training.

Both CLIP and ALIGN exemplify the power of contrastive learning in multimodal AI systems, effectively bridging the gap between textual and visual data through shared embedding spaces.

\subsection{Masked Language Modeling (MLM) in Multimodal Large Language Models}

Masked Language Modeling (MLM) is a traditional technique where the model is trained to predict missing words in a sentence using the surrounding context. In Multimodal Large Language Models (MLLMs), this technique is extended to \textbf{multimodal masked modeling}, where the model must predict masked words and image regions, forcing it to learn joint representations of text and images \cite{ComprehensiveSurvey2024}.

MLLMs employing MLM techniques are used in various applications, including object-centric robotic manipulation, where the model predicts the precise end-effector pose by understanding both the textual instructions and the visual context. This capability is crucial for developing embodied AI systems that can interact with and manipulate their environment effectively \cite{ManipLLM2024}.

The training of MLLMs often involves a combination of image-text contrastive learning, image-text matching, and masked language modeling. These tasks collectively help the model to align visual and textual modalities, improving its performance in tasks such as image captioning, visual question answering, and more complex multimodal interactions \cite{OverviewLMM2024}.

Ongoing research in this area focuses on enhancing the efficiency and accuracy of MLLMs by refining the MLM techniques used, exploring new architectures, and integrating more diverse datasets. These efforts aim to create models that are not only more capable of understanding multimodal inputs but also more efficient in terms of computational resources \cite{ResearchDevelopment2023}.

\subsection{Visual Question Answering (VQA) Pre-training}

Visual Question Answering (VQA) is a crucial task in the realm of multimodal models, where models are pre-trained on tasks such as VQA or image captioning. In this scenario, the model is exposed to paired questions and images, requiring it to learn how to infer relationships between text and visual content. This approach often leverages cross-attention mechanisms to align the two modalities effectively \cite{MaskedVisionLanguage2023}.

Recent advancements in VQA pre-training have shown significant success, particularly in specialized domains such as medical imaging. These models are trained using both unimodal and multimodal contrastive losses, enhancing their ability to understand complex visual and textual interactions. The use of retrieval-augmented methods has further improved the performance of VQA systems by incorporating additional contextual information from large datasets \cite{RAMMBiomedicalVQA2023}.

However, challenges remain due to the limited availability of diverse multimodal datasets, especially in niche areas like medical VQA. This scarcity necessitates innovative approaches to data augmentation and transfer learning to ensure robust model performance across various applications \cite{MedicalVQA2023}.

Ongoing research focuses on refining these pre-training techniques and expanding the dataset availability to improve the versatility and accuracy of VQA models in different contexts \cite{pengfeiliHEU2023}.

\subsection{Vision-and-Language Pretraining (VLP)}

\textbf{Vision-and-Language Pretraining (VLP)} strategies are crucial for developing robust multimodal models. These strategies involve pre-training on diverse tasks such as image-text matching, masked language modeling, and next-sentence prediction, all within a multimodal context. By engaging in multiple tasks simultaneously, the model gains a comprehensive understanding of the interactions between language and vision, enabling it to perform complex reasoning tasks \cite{VisoAI2024}.

Models like UNITER, ViLBERT, and OSCAR are prime examples of this multitask approach, which enhances multimodal reasoning by integrating cross-modal fusions within a dual-encoder architecture. This architecture allows the models to effectively process and align visual and textual data, improving their performance across various multimodal tasks \cite{FlexibleVLP2023}.

Moreover, recent advancements in VLP strategies have addressed challenges related to heterogeneity in federated learning environments, particularly in specialized domains such as biomedical applications. These improvements have been instrumental in refining the flexibility and adaptability of VLP models, making them more efficient in handling diverse datasets and tasks \cite{HeterogeneityFederatedVLP2024}.

Ongoing research continues to explore innovative pretraining strategies, aiming to further enhance the capabilities and efficiency of VLP models in understanding and reasoning about multimodal data \cite{FeedbackModalSearch2024}.

\section{Fine-Tuning for Specific Tasks}

After pre-training, Multimodal Large Language Models (MLLMs) are typically fine-tuned on specific tasks to maximize their performance in particular domains. Fine-tuning is an essential process that adapts the general knowledge gained during pre-training to the nuances of a specific task, ensuring that the model can deliver more accurate and relevant results \cite{turn0search5}.

This process involves adjusting the model’s parameters using task-specific data, which helps in aligning the model’s capabilities with the requirements of the target application. Techniques such as task-specific instruction tuning and the use of multiway adapters have been developed to make fine-tuning more efficient and less resource-intensive \cite{turn0search4}.

Moreover, fine-tuning allows MLLMs to leverage their multimodal understanding, enhancing their ability to process and integrate information from different modalities, such as text and images, which is particularly beneficial in complex tasks \cite{turn0search3}.

Ongoing research aims to further streamline fine-tuning processes, reducing the computational costs and improving the adaptability of MLLMs to a wider range of tasks and domains \cite{turn0search7}.

\subsection{Task-Specific Datasets}

Fine-tuning Multimodal Large Language Models (MLLMs) necessitates the use of task-specific datasets to tailor the model’s capabilities to particular applications. These datasets provide the model with domain-specific knowledge, enabling it to perform tasks such as image captioning, visual question answering (VQA), and cross-modal retrieval with greater accuracy and relevance.

For instance, in image captioning, datasets like the Microsoft Common Objects in Context (MS COCO) are extensively used. MS COCO comprises over 330,000 images, each annotated with multiple captions, offering a rich resource for training models to generate descriptive textual representations of visual content. The diversity and volume of this dataset help models learn to associate visual features with corresponding textual descriptions, enhancing their ability to generate accurate and contextually appropriate captions.

In the realm of Visual Question Answering, the VQA 2.0 dataset is a prominent resource. It contains over 1.1 million questions based on over 200,000 images, with each question designed to test the model’s ability to comprehend and reason about visual information in conjunction with textual queries. The dataset includes questions that require an understanding of object recognition, counting, and spatial reasoning, among other skills, thereby challenging models to develop a nuanced understanding of the interplay between visual and textual data.

Beyond these, specialized datasets cater to niche applications. For example, the TextVQA dataset focuses on questions that require reading and understanding text present within images, pushing models to integrate optical character recognition with visual reasoning. Similarly, the GQA dataset emphasizes compositional question answering, assessing a model’s ability to handle complex queries that involve multiple reasoning steps.

The selection of an appropriate dataset is crucial for effective fine-tuning. Datasets must be carefully curated to ensure they are representative of the target domain and task. High-quality annotations, diversity in data samples, and sufficient volume are key factors that influence the success of the fine-tuning process. Moreover, the alignment between the dataset’s characteristics and the intended application significantly impacts the model’s performance and generalization capabilities.

In summary, task-specific datasets are foundational to the fine-tuning of MLLMs, providing the necessary context and examples for models to adapt their pre-trained knowledge to specific tasks. The careful selection and utilization of these datasets enable MLLMs to achieve high performance across a diverse array of multimodal applications.

\subsection{Learning Rate Scheduling and Optimization}

Fine-tuning Multimodal Large Language Models (MLLMs) necessitates meticulous adjustment of the learning rate to ensure effective adaptation to specific tasks. During fine-tuning, it is common practice to employ smaller learning rates compared to those used in the pre-training phase. This approach helps preserve the general knowledge acquired during pre-training while allowing the model to refine its parameters for the target task.

The learning rate is a critical hyperparameter that dictates the step size at each iteration during the optimization process. An appropriately chosen learning rate facilitates efficient convergence to an optimal solution, whereas an excessively high learning rate can lead to instability and divergence, and an overly low learning rate may result in prolonged training times and suboptimal performance.

To manage the learning rate effectively, various scheduling strategies are employed. One common approach is the StepLR scheduler, which reduces the learning rate by a fixed factor after a specified number of epochs. For instance, the learning rate might be decayed by a factor of 0.1 every 10 epochs, allowing the model to make significant updates initially and finer adjustments as training progresses \cite{turn0search11}.

Another strategy is the ReduceLROnPlateau scheduler, which monitors a specific metric, such as validation loss, and reduces the learning rate when the metric ceases to improve. This adaptive approach enables the model to adjust the learning rate in response to its performance, potentially leading to more efficient training \cite{turn0search11}.

In addition to learning rate scheduling, the choice of optimization algorithm plays a pivotal role in fine-tuning. AdamW, an extension of the Adam optimizer that incorporates weight decay, is widely used due to its ability to handle sparse gradients and maintain a balance between adaptive learning rates and regularization. The weight decay term helps prevent overfitting by penalizing large weights, thereby promoting generalization.

Recent research has also explored the integration of learning rate scheduling with adaptive optimizers. For example, the Gradient-based Learning Rate scheduler (GLR) dynamically adjusts the learning rate based on the norm of the gradient, aiming to reduce the tuning effort and achieve competitive results across various tasks \cite{turn0search9}.

In summary, effective fine-tuning of MLLMs requires careful consideration of learning rate scheduling and optimization techniques. By employing appropriate learning rate schedules and optimizers, practitioners can enhance the model’s performance on specific tasks while preserving the valuable knowledge gained during pre-training.

\subsection{Multitask Fine-Tuning}

In certain scenarios, models are fine-tuned on multiple tasks simultaneously, a technique known as \textbf{multitask learning}. This approach enhances the model’s ability to generalize across various related tasks \cite{MFTCoder}. For example, a model might be fine-tuned concurrently on image captioning and visual question answering datasets, allowing it to perform well across both contexts \cite{PEFT}. Multitask learning leverages shared features among tasks, which improves overall performance and efficiency \cite{SPT}.

However, multitask fine-tuning also presents challenges, such as task interference and increased computational demands. To address these, researchers have developed parameter-efficient fine-tuning (PEFT) techniques. Methods like Low-Rank Adaptation (LoRA) and Quantized LoRA (QLoRA) adjust only a subset of model parameters, reducing computational costs while maintaining strong performance \cite{PEFT}. An example of a successful multitask fine-tuning application is MFTCoder, a framework designed to enhance code generation capabilities by training on multiple tasks. Experimental results demonstrate that MFTCoder achieved a pass@1 score of 74.4\% on the HumanEval benchmark, surpassing GPT-4’s performance \cite{MFTCoder}.

These studies show the potential of multitask fine-tuning in developing efficient, versatile models that can perform well in diverse tasks with fewer resource requirements.

\subsection{Cross-Modal Tasks}

Fine-tuning is essential for tasks that require the model to reason across modalities, such as cross-modal retrieval or referring expression comprehension, where the model must identify specific objects in an image based on a text description. The goal during this phase is to effectively align the visual and textual representations \cite{m2ist}. Recent studies have demonstrated that fine-tuning large pre-trained models for cross-modal tasks yields impressive results but can be computationally expensive \cite{loupe}. To address this, parameter-efficient transfer learning techniques have been developed, which update only a subset of parameters to adapt pre-trained models to downstream tasks efficiently \cite{petl}. 

For example, the M2IST framework (Multi-Modal Interactive Side-Tuning) introduces side-tuning with a mixture of multi-modal interactive side adapters, enabling better vision-language alignment and efficient fine-tuning for tasks like referring expression comprehension \cite{m2ist}. Additionally, methods such as LOUPE focus on learning fine-grained semantic alignment between visual regions and textual phrases, enhancing the model's capacity for cross-modal tasks \cite{loupe}. These approaches underscore the importance of fine-tuning in achieving effective cross-modal alignment, critical for integrating visual and textual information in complex tasks.

\section{Few-Shot and Zero-Shot Learning in Multimodal Large Language Models}

Few-shot and zero-shot learning have emerged as powerful capabilities of Multimodal Large Language Models (MLLMs), enabling them to generalize to new tasks with minimal or no task-specific data. This is particularly valuable when labeled datasets are scarce or expensive to curate. For instance, the Frozen model demonstrates that a pre-trained language model can be extended to multimodal tasks without updating its weights, effectively transferring few-shot learning abilities to settings involving both vision and language \cite{tsimpoukelli2021multimodal}. Similarly, Kosmos-1 is a Multimodal Large Language Model trained from scratch on web-scale multimodal corpora, including interleaved text and images, image-caption pairs, and text data, showcasing the model’s capacity to perform various tasks without gradient updates or fine-tuning \cite{kosmos2023}. These advancements highlight the potential of MLLMs in scenarios where traditional supervised learning is impractical, paving the way for more adaptable and efficient AI systems.

\subsection{Few-Shot Learning}

In \textbf{few-shot learning}, a model is fine-tuned on a small number of examples for a new task. For Multimodal Large Language Models (MLLMs), this means that after extensive pre-training, the model can quickly adapt to new tasks by observing just a handful of image-text pairs or task-specific examples. This capability is particularly advantageous in scenarios where labeled data is scarce or expensive to obtain.

The process of few-shot learning in MLLMs typically involves presenting the model with a few examples of the new task during the fine-tuning phase. These examples serve as a guide, enabling the model to understand the task's structure and requirements. For instance, in image captioning, the model might be shown a few images along with their corresponding captions, using this limited information to generate captions for new, unseen images.

A notable example of this approach is the "Frozen" model, which integrates a pre-trained language model with a vision encoder. In this setup, the language model's weights remain unchanged ("frozen"), while the vision encoder is trained to produce embeddings that the language model can interpret. This design allows the model to perform tasks such as visual question answering and image captioning with minimal task-specific data, effectively transferring the few-shot learning capabilities of language models to multimodal settings \cite{tsimpoukelli2021multimodal}.

Few-shot learning is especially useful for niche tasks where only a limited amount of data is available. In such cases, traditional supervised learning methods, which require large datasets, are impractical. Few-shot learning enables models to generalize from a small number of examples, making it feasible to develop AI systems for specialized applications without extensive data collection and annotation.

However, few-shot learning also presents challenges. The model must generalize effectively from limited examples, which can be difficult if the new task differs significantly from those encountered during pre-training. Additionally, the quality of the examples provided during fine-tuning is crucial; poor-quality examples can lead to suboptimal performance.

To address these challenges, researchers are exploring various strategies. One approach involves enhancing the pre-training phase by incorporating a diverse range of tasks and data modalities, equipping the model with a broader knowledge base to draw upon during few-shot learning. Another strategy focuses on developing more sophisticated fine-tuning techniques that can better leverage the limited examples available.

In summary, few-shot learning enables MLLMs to adapt to new tasks with minimal data, offering a practical solution for applications where large labeled datasets are unavailable. While challenges remain, ongoing research continues to improve the effectiveness and efficiency of few-shot learning in multimodal contexts.

\subsection{Zero-Shot Learning}

\textbf{Zero-shot learning} refers to a model’s ability to perform tasks without having seen any examples of that task during training. In the context of Multimodal Large Language Models (MLLMs), this capability enables the model to generalize across tasks and domains by leveraging its understanding of the relationships between different modalities, such as text and images.

A prominent example of this is the CLIP (Contrastive Language–Image Pretraining) model, which is trained on a vast dataset of text-image pairs. By learning to associate textual descriptions with corresponding images, CLIP develops a rich multimodal representation that allows it to perform zero-shot image classification. This means that CLIP can assign labels to images it has never encountered before, simply by interpreting the textual descriptions associated with those labels. For instance, when presented with an image of a previously unseen object, CLIP can accurately classify it by matching the image to the most relevant textual label from its training data \cite{radford2021learning}.

The success of zero-shot learning in MLLMs like CLIP is attributed to their ability to capture high-level semantic information across modalities. By aligning visual and textual representations in a shared embedding space, these models can transfer knowledge from one domain to another, facilitating tasks such as image retrieval, captioning, and classification without task-specific training data.

Recent research has further explored the potential of zero-shot learning in MLLMs. For example, the study “Large Multilingual Models Pivot Zero-Shot Multimodal Learning across Languages” demonstrates that multilingual language models can serve as pivots for zero-shot multimodal learning across different languages. By leveraging a strong multilingual large language model, multimodal models pretrained on English-only image-text data can generalize to other languages in a zero-shot manner, even surpassing models trained on image-text data in native languages \cite{hu2023large}.

Another study, “MultiInstruct: Improving Multi-Modal Zero-Shot Learning via Instruction Tuning,” investigates the enhancement of zero-shot performance on various unseen multimodal tasks through instruction tuning. The researchers fine-tuned a multimodal model on a diverse set of tasks and instructions, demonstrating strong zero-shot performance and reduced sensitivity to variations in instructions for each task \cite{xu2023multiinstruct}.

These advancements highlight the growing importance of zero-shot learning in the development of versatile and efficient AI systems. By enabling models to perform tasks without task-specific training data, zero-shot learning reduces the reliance on large labeled datasets, making it particularly valuable in scenarios where data collection is challenging or impractical.

\subsection{Transfer Learning}

Few-shot and zero-shot learning are made possible by \textbf{transfer learning}, where knowledge gained from pre-training on one set of tasks is transferred to new, unseen tasks. This is particularly effective in MLLMs because they are trained on large, diverse multimodal datasets that cover a wide range of text and visual domains, allowing for strong generalization across tasks \cite{hu2023large}.

\section{Instruction Tuning for MLLMs}

Instruction tuning is a newer technique that enhances the ability of MLLMs to follow human instructions across modalities. It involves fine-tuning the model using explicit instructions in natural language, enabling the model to perform a broader range of tasks with greater flexibility and accuracy \cite{peng2023instruction}.

\subsection{Natural Language Instructions}

Instruction tuning uses datasets where tasks are framed as natural language instructions. For instance, instead of providing just an image and asking the model to generate a caption, the model is given a prompt like, \textit{"Describe the image in detail."} This allows the model to understand human-like instructions and follow them more closely \cite{peng2023instruction}.

\subsection{Multimodal Instruction Tuning}

Instruction tuning can also be applied to multimodal tasks. In this case, the model is trained to follow multimodal prompts that involve both text and images. For example, a task might include an image and the instruction, \textit{"What is the person in the image doing?"} This helps the model learn to follow complex, human-like commands that span multiple modalities \cite{peng2023instruction}.

\subsection{Improving Generalization}

Instruction tuning is designed to improve the model’s generalization abilities. By training the model to interpret instructions in natural language, it becomes more flexible in handling new tasks without needing extensive retraining. This technique can also be combined with few-shot and zero-shot learning, further enhancing the model’s ability to generalize across tasks with minimal additional data.

\subsection{Applications of Instruction Tuning}

Instruction-tuned MLLMs are particularly useful in interactive AI systems, where users provide instructions in natural language and expect the AI to perform a task based on those instructions. This has broad applications in personal assistants, customer service bots, and even creative tasks like generating art or stories based on user prompts.

Instruction-tuned models enhance the capabilities of personal assistants by allowing them to understand and execute complex user instructions. This includes managing schedules, setting reminders, and even controlling smart home devices through natural language commands \cite{gupta2023instruction}.

\subsection{Instruction Tuning for MLLMs} 
\begin{figure}
    \centering
    \includegraphics[width=0.7\linewidth]{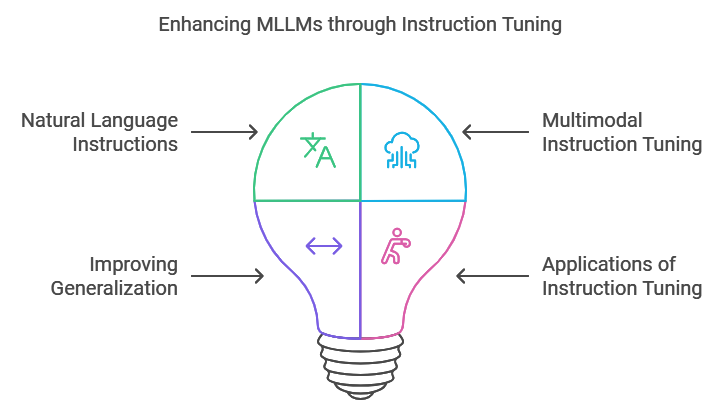}
    \caption{Instruction Tuning for MLLMs}
    \label{fig:Fine-tuning1}
\end{figure}

In customer service, instruction-tuned models can handle a wide range of queries by understanding and responding to customer instructions with high accuracy. Instruction-tuned models are also valuable in creative domains, such as generating art, stories, or music based on user prompts \cite{liu2024visual}. These models can be used in educational tools to provide personalized learning experiences. In software development, instruction-tuned models can assist in code generation and debugging by following developer instructions.

\chapter{Applications of MLLMs in Vision-Language Tasks}

\section{Image Captioning and VQA}
The integration of computer vision and natural language processing (NLP) has led to remarkable advancements in tasks like \textbf{Image Captioning} and \textbf{Visual Question Answering (VQA)}. Both fields aim to generate rich textual interpretations of visual data, requiring a deep understanding of visual elements and linguistic relationships. Historically, early models relied on hand-crafted features, but modern systems, particularly with the rise of \textbf{Multimodal Large Language Models (MLLMs)}, have revolutionized the performance and scope of these applications. Leveraging vast datasets and sophisticated architectures, MLLMs now generate captions and answer questions about images with unprecedented accuracy, making them integral to several real-world applications.

This paper surveys the progress of MLLMs in both image captioning and VQA, illustrating their technical underpinnings and applications.

\subsection{Image Captioning: Overview and Advances}
Image captioning refers to automatically generating textual descriptions for images by combining visual and linguistic processing. Early methods were limited to hand-crafted rules, but modern deep learning approaches, particularly through MLLMs, have dramatically improved performance. By training on large image-text datasets like MSCOCO and Flickr, MLLMs can generate rich and contextually accurate captions, significantly outperforming older methods \cite{icmeta2020m2transformer, icmeta2019oscar, icmeta2020densecap}.

Key advancements include techniques like:
\begin{itemize}
    \item \textbf{OSCAR (Object-Semantics Aligned Pre-training)}: Enhances captioning by aligning object tags with textual descriptions during pre-training. This leads to better object recognition and semantically rich captions \cite{icmeta2019oscar}.
    \item \textbf{VIVO (Visual Vocabulary Pre-training)}: Introduces a vocabulary of visual concepts, allowing models to caption novel objects unseen during training, crucial for real-world applications \cite{icmeta2019vivo}.
    \item \textbf{Dense Captioning}: A novel approach that generates region-specific captions for different parts of an image, useful in detailed image understanding and retrieval \cite{icmeta2020densecap}.
    \item \textbf{Generative Adversarial Networks (GANs)}: Applied to refine caption fluency and coherence by leveraging adversarial training \cite{icmeta2021gan}.
    \item \textbf{Meta-Learning Approaches}: Enable MLLMs to quickly adapt to new tasks with minimal data, improving generalization across various tasks, especially when training data is scarce \cite{icmeta2019meta}.
\end{itemize}

\subsection{Visual Question Answering (VQA): Overview and Advances}
VQA is an interdisciplinary field requiring a system to generate accurate answers to questions posed about images, combining both visual and textual reasoning \cite{vqa_survey2023}. Unlike image captioning, which describes an image holistically, VQA focuses on specific queries about an image's content.

MLLMs have made significant strides in VQA tasks by utilizing the same underlying multimodal architectures:
\begin{itemize}
    \item \textbf{MCAN (Multimodal Co-Attention Network)}: Uses co-attention mechanisms to fuse image and text features, resulting in improved understanding of image-question relationships \cite{mcan_vqa2019}.
    \item \textbf{Knowledge-Enhanced VQA Models}: Incorporate external knowledge graphs for commonsense reasoning, improving performance on complex VQA tasks \cite{lan2023improvingzeroshotvisualquestion}.
\end{itemize}

\subsection{Applications of Image Captioning and VQA}

The advancements in MLLMs have expanded their applications across diverse domains, offering solutions to various real-world problems by bridging the gap between visual data and language.

\begin{itemize}
    \item \textbf{Assistive Technologies} Both image captioning and VQA play critical roles in assistive technologies for the visually impaired. Image captioning systems convert visual scenes into real-time audio descriptions, helping users interpret their surroundings. For instance, applications like Microsoft Seeing AI use MLLM-generated captions to describe objects, people, and text in a user's environment \cite{icmeta2021assistive}. VQA enhances this by allowing users to ask specific questions about their environment, such as "What is the name on the sign?" or "Is there anyone near me?" This interactivity elevates the autonomy of visually impaired individuals, enabling more tailored assistance \cite{assistive_vqa2020}.
    \item \textbf{Autonomous Systems and Vehicles:} In autonomous vehicles, the ability to generate accurate captions and answer complex questions about the driving environment is critical for safety and decision-making. MLLMs generate captions describing road conditions, obstacles, pedestrians, and traffic signs, improving the vehicle's situational awareness \cite{icmeta2020autonomous}. VQA complements this by enabling the system to answer real-time queries like "Is there a stop sign ahead?" or "What is the speed limit?" enhancing the vehicle's capacity to navigate complex environments safely \cite{autonomous_vqa2019}.
    
    \item \textbf{Medical Imaging and Healthcare:} In medical imaging, MLLM-based captioning systems automatically generate diagnostic reports by interpreting X-rays, CT scans, and MRIs. This reduces the workload on radiologists and ensures faster report generation with high accuracy. Tools like CaptionHealth utilize these techniques to assist in real-time diagnostics \cite{icmeta2021medical}. VQA models enable clinicians to query specific aspects of medical images, such as "What abnormalities are present?" or "Is there evidence of pneumonia?" This leads to more interactive and accessible healthcare diagnostics, aiding in rapid decision-making \cite{med_vqa2019, healthcare_vqa2021}.
    
    \item \textbf{Content Moderation and Search Engines:} On social media platforms, MLLMs are used to generate captions that automatically tag images and flag inappropriate content, making content moderation more efficient. Platforms like Facebook and Instagram employ image captioning tools to scale moderation efforts by identifying objectionable or dangerous content from uploaded images \cite{icmeta2020content}. VQA further enhances content moderation by allowing moderators to query images directly, asking questions such as "Does this image contain violence?" or "Is there inappropriate text in the background?" This interactive capability allows for more nuanced and scalable moderation systems \cite{moderation_vqa2021}.
\end{itemize}

\textbf{Applications of Image Captioning and VQA} 
\begin{figure}
    \centering
    \includegraphics[width=0.7\linewidth]{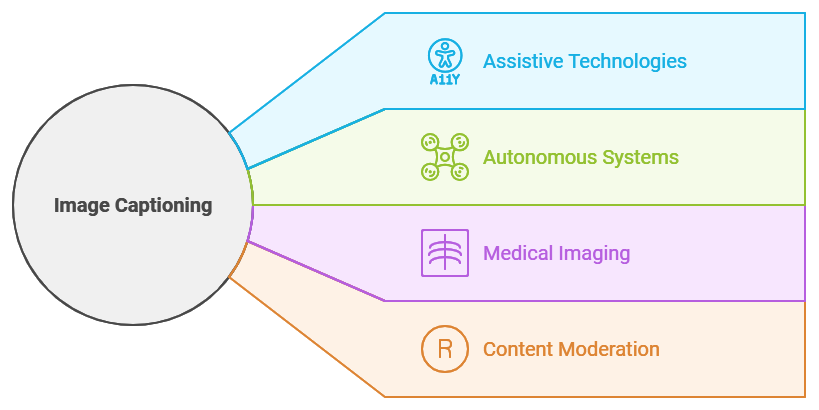}
    \caption{Applications of Image Captioning and VQA}
    \label{fig:Fine-tuning2}
\end{figure}

Multimodal Large Language Models have revolutionized the fields of image captioning and VQA, providing sophisticated tools that understand and interpret complex visual information through text. From assistive technologies to autonomous systems, healthcare to content moderation, the applications of MLLMs continue to grow, promising a future where machines can interpret and respond to visual data with human-like comprehension.

\section{Visual Storytelling and Scene Understanding}

\subsection{Introduction}

Multimodal Large Language Models (MLLMs) have profoundly impacted how AI handles tasks involving both visual and textual information. These models merge visual inputs with language processing, enabling systems to comprehend complex scenes and generate coherent narratives. This ability is especially crucial in fields like autonomous driving, interactive media, 3D modeling, and human-computer interaction \cite{vs2024li,vs2020hong}. This survey reviews recent advancements in visual storytelling and scene understanding, focusing on how MLLMs integrate visual and textual information to enable richer semantic understanding and narrative generation.

\subsection{Technologies for Visual Storytelling and Scene Understanding}

Visual storytelling, defined as generating coherent narratives from sequences of images or videos, has evolved from simpler object recognition methods to more sophisticated models that combine both visual semantics and contextual information. Early models often used scene graphs to capture relationships between objects but struggled with narrative coherence. The introduction of MLLMs has allowed for a more detailed interpretation of these relationships, integrating semantic layers for more nuanced storytelling.

For instance, the \textit{Multidimensional Semantic Augmented Network} provides a method for merging scene, object, and action semantics, enabling richer narrative construction \cite{vs2024li}. These models go beyond the static understanding of objects and scenes, instead focusing on how individual elements interact over time to form cohesive stories \cite{vs2024li}. In another example, \textit{Kosmos-1} leverages cross-modal knowledge transfer between vision and language, resulting in stories that are not only descriptive but also contextually relevant \cite{vs2024song}. By drawing on both visual and textual data, these models enable more fluent and complex storytelling compared to earlier systems.

Scene understanding, another key task, involves comprehending the spatial and relational structure of objects within a scene. Advanced models like \textit{Scene-LLM} integrate 3D visual data with textual descriptions, allowing for high-level reasoning about spatial relationships and object interactions. The use of hybrid 3D feature representations, which combine both global scene-level information and local object-centric details, enables models to reason about dynamic environments effectively \cite{vs2024rao}. This is particularly important in real-time applications, such as robotics and autonomous driving, where accurate scene understanding is critical to decision-making.

\subsection{Applications}

MLLMs are proving essential in several domains, where their ability to generate, understand, and manipulate multimodal data offers significant advantages.

In the \textit{entertainment industry}, MLLMs are used to automatically generate narratives for films, games, and other media. By analyzing sequences of images or video, these models can create storylines that evolve dynamically based on the characters' actions or environmental changes. For instance, games can now feature AI-driven storylines that adapt based on player decisions, creating an immersive and interactive experience \cite{vs2020parde}. Similarly, in content generation for streaming platforms, MLLMs are used to develop personalized narratives by analyzing viewer preferences and tailoring story elements accordingly \cite{vs2024song}.

In the realm of \textit{autonomous driving}, scene understanding is vital for real-time decision-making. MLLMs like \textit{OmniDrive} enhance autonomous systems' abilities to interpret 3D environments by accurately analyzing traffic situations, detecting potential hazards, and predicting the actions of other vehicles \cite{vs2024alvarez}. These models are trained on large datasets that include diverse driving scenarios, enabling them to generalize across different environments, including complex urban settings. By understanding the spatial and temporal relationships between objects, autonomous vehicles can make safer and more informed decisions on the road \cite{vs2024rao}.

For \textit{augmented reality (AR)} and \textit{interactive storytelling}, MLLMs offer the potential to create dynamic narratives that respond to user interactions with their environment. In AR applications, these models analyze the physical space around the user and generate stories or instructions that adapt based on real-world inputs. For example, in an AR-based learning environment, an MLLM might generate contextual stories or educational content based on the objects and scenes detected in a classroom or outdoor setting \cite{vs2019dey}. This ability to blend the physical and digital worlds creates highly personalized experiences that can be applied in education, entertainment, and marketing.

In \textit{robotics} and \textit{embodied AI}, MLLMs like \textit{Scene-LLM} improve robots' abilities to navigate and interact with their environments. Robots equipped with scene understanding models can perform tasks in complex, unstructured environments, such as warehouses or hospitals. These systems use 3D scene representations to understand their surroundings, make decisions about where to move, and interact with objects based on their spatial relationships \cite{vs2024rao1}. In healthcare, for instance, robots might assist with patient care by navigating hospital environments and delivering medication based on their understanding of room layouts and equipment locations.

In \textit{content generation}, MLLMs have started to enable platforms that provide automatic captioning and summarization of images and videos. Applications such as digital marketing, social media content creation, and even journalism benefit from these models' ability to produce rich, engaging narratives from visual content \cite{vs2024zang}. This automated content creation reduces time and resources, while increasing personalization, allowing brands to engage more effectively with their audiences.

Future directions for MLLMs in these applications include improving real-time processing, enhancing model interpretability, and reducing the computational cost of deploying these large models in dynamic, resource-constrained environments. Addressing these challenges will be crucial for expanding MLLMs' use in more diverse, real-world scenarios \cite{yang2024viassist}.

\section{MLLM Applications in Content Creation and Editing}

\subsection{Introduction}

Multimodal Large Language Models (MLLMs) have significantly influenced the way content is created and edited, offering powerful tools for automating and enhancing tasks across various media formats. MLLMs excel in integrating text, image, video, and even audio data, allowing creators to generate, transform, and refine multimedia content. By leveraging these capabilities, MLLMs have become indispensable in fields such as multimedia storytelling, automated content generation, real-time editing, and collaborative creative workflows \cite{vs2024chang}. This section provides a detailed analysis of the technologies driving MLLMs and their specific applications in content creation and editing.

\subsection{Technologies Behind MLLM in Content Creation}

MLLMs are powered by several technological innovations, including transformers, Generative Adversarial Networks (GANs), and Vision-Language Models (VLMs). These technologies enable the models to understand and generate multimodal content, transforming industries such as entertainment, marketing, and journalism. 

The integration of GPT models for text generation with DALL-E and other image generation systems allows creators to produce entire multimedia pieces from simple prompts \cite{vs2024chang}. Advanced models like Video-LLaMA are pushing the boundaries by combining visual and textual inputs, enabling applications such as video editing and interactive storytelling. Self-supervised learning and multimodal training datasets enable these models to understand complex relationships between text, images, and videos, making them versatile tools for content creation \cite{vs2024song}.

\subsection{Applications in Content Creation and Editing}

The following are the key applications of MLLMs in the field of content creation and editing:

\begin{itemize}

\item \textbf{Multimodal Content Generation}: MLLMs allow artists and creators to generate images, text, and video content based on simple inputs. These models can interpret text prompts to create visual art, design characters, or craft environments for video games and films. This application is revolutionizing multimedia storytelling, where different types of content are seamlessly integrated to tell compelling narratives \cite{vs2024chang,vs2024schmidt}. MLLMs are also utilized in marketing and social media, where they analyze user preferences and trends to generate personalized content \cite{vs2020parde}.

\item \textbf{Real-Time Video and Image Editing}: MLLMs are revolutionizing video and image editing by enabling real-time modifications through natural language and visual inputs. Systems like ExpressEdit allow users to sketch over a video frame or provide verbal commands to alter specific elements in a scene \cite{vs2024tilekbay}. These models also support automated color correction, object tracking, and scene enhancements, making them invaluable in the film and advertising industries \cite{vs2024kubicek}.

\item \textbf{Automated Script and Article Writing}: MLLMs such as GPT-4 are highly effective in generating long-form text, including movie scripts, articles, and reports. By providing minimal input, users can rely on these models to produce well-structured and coherent content, freeing up time for more creative tasks \cite{vs2024eleftheriadis}. This application is especially valuable in journalism, where automated systems generate drafts that can be fine-tuned by editors \cite{vs2024anderson}.

\item \textbf{Collaborative Content Creation}: Collaborative content development tools powered by MLLMs allow teams to simultaneously work on different aspects of a project, from visual elements to text descriptions. Cloud-based platforms enable real-time collaboration, ensuring that all contributors are working on the most up-to-date version of the project \cite{vs2024santos}. In interactive design, tools like U-CREATE help creators develop augmented reality experiences and location-based services, streamlining the authoring process \cite{vs2024sauer}.

\item \textbf{Mobile Multimedia Editing}: Mobile applications integrating MLLMs have made content creation accessible to a broader audience. These apps allow users to edit and generate multimedia content on their devices using intuitive commands and gestures. By automating many technical aspects of editing, MLLM-powered mobile apps empower social media influencers and small businesses to create professional-grade content without requiring advanced editing skills \cite{vs2024jokela}.

\item \textbf{Content Repurposing and Multilingual Adaptation}: MLLMs are instrumental in repurposing content for different platforms and adapting it to various languages. Whether it's reformatting a blog post for social media or translating a promotional video into multiple languages, MLLMs maintain the original intent and adjust content to fit new formats and audiences \cite{vs2024obrenovic}. For global marketing campaigns, MLLMs can automate the localization of content, ensuring consistency across different regions while adapting to cultural nuances \cite{vs2024bateman}.

\item \textbf{Creative Personalization}: MLLMs are widely used for personalizing content based on user behavior and preferences. For example, personalized recommendations for entertainment platforms can be generated using MLLMs that analyze viewing habits, likes, and social trends \cite{vs2024bateman}. This application is particularly useful in e-commerce, where product descriptions and promotional materials are customized to reflect the tastes of individual users, improving engagement and conversion rates \cite{vs2024schmidt}.

\item \textbf{Dynamic Multimedia Creation}: MLLMs can dynamically generate multimedia presentations by combining textual, visual, and audio elements. This capability is invaluable in industries like education and training, where interactive and adaptive content is essential for engaging users. In educational technology, MLLMs assist in creating lesson plans, video tutorials, and interactive learning modules that adapt to the learner's pace and preferences \cite{vs2024anderson}.

\end{itemize}

MLLMs have ushered in a new era of content creation and editing by offering tools that automate and enhance creative processes. From multimodal content generation to collaborative editing, these models allow for the seamless integration of text, image, and video, transforming how multimedia content is produced. As MLLMs continue to evolve, we can expect further innovations in personalized content, real-time editing, and multilingual adaptation, opening new opportunities in industries ranging from entertainment to marketing \cite{vs2024chang,vs2024kubicek}.

\section{MLLM Applications in Cross-Modal Retrieval and Search}

\subsection{Introduction}

Cross-modal retrieval and search refers to retrieving relevant information from one modality (e.g., text) based on a query from another modality (e.g., image or audio). The growing availability of multimodal data, such as images, text, audio, and video, has made cross-modal retrieval a significant research area. Multimodal Large Language Models (MLLMs) have shown exceptional capability in addressing the complexities of cross-modal retrieval by understanding and relating different modalities. This section explores the technologies, methodologies, and applications of MLLMs in cross-modal retrieval and search, emphasizing recent advancements.

\subsection{Technological Foundations of Cross-Modal Retrieval}

MLLMs rely on a combination of multimodal transformers dual-encoder architectures, and self-supervised learning to manage the relationships between different types of data. For instance, Vision-Language Models (VLMs) such as CLIP and DALL-E have demonstrated their ability to map images to descriptive text, and vice versa, enabling efficient cross-modal retrieval. These models use contrastive learning to align visual and textual representations in a shared semantic space \cite{vs2024li,vs2015ranjan}. Other architectures, such as cross-attention mechanisms, help in precisely understanding the relationships between elements of different modalities.

Generative models like GPT-4V extend this capability by generating relevant outputs in one modality based on inputs from another. These models use large multimodal datasets to learn intricate correlations between visual, textual, and auditory data, enabling more nuanced retrieval results \cite{vs2024gomez}. 

\subsection{Applications of MLLMs in Cross-Modal Retrieval and Search}

\begin{itemize}

\item \textbf{Image-Text Retrieval} One of the most widespread applications of MLLMs in cross-modal retrieval is the search for images based on textual queries or vice versa. This has applications in areas such as e-commerce, where users can search for products using images or text descriptions. Models like CLIP map images and text into the same latent space, making it possible to retrieve images that semantically match the input text \cite{vs2024li}. Advanced models also allow retrieval of more abstract visual concepts, such as emotion or style, based on text queries.

\item \textbf{Video-Audio-Text Retrieval}: Cross-modal retrieval extends to the video and audio domains as well, with applications in multimedia search engines and content recommendation systems. For example, users can retrieve relevant video clips by providing a text description or even an audio snippet. Systems like Video-LLaMA and Speech2Text search can match video content with textual or spoken queries, allowing for accurate retrieval in large multimedia databases \cite{vs2014chen,vs2024gomez}. This has significant implications for media platforms, allowing users to discover video content based on both audio and text inputs.

\item \textbf{Generative Cross-Modal Retrieval}: Generative models are also pushing the boundaries of cross-modal retrieval by enabling systems to create new content from user queries. For example, DALL-E and similar models allow users to generate images based on detailed textual descriptions, while other systems are capable of generating music or video content from text-based prompts. This generative paradigm offers a new way of approaching content search and retrieval, where instead of finding existing content, users can generate what they need \cite{yin2023survey,vs2019muller}.

\item \textbf{Multi-Lingual and Cross-Lingual Retrieval}: MLLMs also excel at cross-lingual retrieval, where text or speech in one language can be used to retrieve relevant content in another language. Systems trained on multilingual datasets, such as Speech2Text or multilingual transformers, enable retrieval of multimedia content across languages without needing parallel translation. This is particularly useful in global applications, where users may search in one language and receive content from another \cite{vs2024gomez,vs2024li}. For instance, a user could search using an English query and retrieve relevant French or Spanish media files, promoting a more inclusive digital ecosystem.

\item \textbf{Cross-Modal Music Retrieval}: Cross-modal retrieval extends beyond text, images, and video into music as well. Researchers have developed systems that enable users to find musical pieces based on descriptions of melodies, moods, or even visual stimuli such as album covers or sheet music \cite{vs2019muller}. These systems can enhance music recommendation platforms, allowing users to find music across various modalities, leading to more immersive listening experiences.

\item \textbf{Lecture Video Retrieval} Cross-modal retrieval systems are becoming increasingly important in educational technology. For example, models like the Multi-modal Language Model (MLM) have been applied to index and retrieve lecture videos based on both spoken content and text on slides \cite{vs2014chen}. This allows students to quickly find relevant portions of lecture videos by searching with either keywords or topics, greatly improving the efficiency of content retrieval in educational platforms.

\item \textbf{Content-Based Image Retrieval in Medical Domains} In the medical domain, cross-modal retrieval is being used to link textual descriptions (e.g., symptoms or diagnoses) with relevant medical images such as X-rays or MRIs. MLLMs trained on multimodal medical datasets can assist healthcare professionals in retrieving and analyzing medical data based on both text and images, improving diagnostic accuracy and speeding up research workflows \cite{vs2024jiang,vs2018dorfer}.

\item \textbf{Interactive Search in AR/VR}: Augmented reality (AR) and virtual reality (VR) applications are using cross-modal retrieval to enable more interactive and immersive experiences. Users can search for virtual objects, spaces, or experiences by describing them with words or gestures, and MLLMs process these inputs to retrieve or generate corresponding virtual environments or objects. This is particularly useful in gaming, training simulations, and virtual tourism \cite{vs2024li}.

\end{itemize}

Despite the progress in cross-modal retrieval, several challenges remain. One major challenge is the ability to handle domain-specific data such as medical or legal information, where cross-modal models must accurately interpret highly specialized data. Another challenge is the scalability of cross-modal retrieval in real-time systems, where the latency of retrieval responses is critical for applications like video streaming and e-commerce \cite{vs2024chang,vs2024gomez}. Lastly, improving the accuracy and contextual relevance of cross-modal retrieval results, especially in complex scenarios involving abstract concepts or multilingual inputs, remains an ongoing area of research.

Future directions include the enhancement of personalized retrieval systems that adapt to user preferences and the exploration of cross-modal reasoning capabilities, where models not only retrieve content but also infer and reason across modalities \cite{vs2024yin,vs2024palmagomez}. Additionally, integrating real-time data processing for AR and VR applications could open new possibilities in immersive and interactive cross-modal experiences.

MLLMs have revolutionized cross-modal retrieval and search by offering more accurate, scalable, and flexible solutions for handling multimodal data. From enabling image-text searches to creating generative content from user queries, MLLMs are expanding the boundaries of how we interact with multimedia content. As these models continue to evolve, they promise to play an even more central role in a variety of fields, including healthcare, education, entertainment, and beyond.

\section{MLLMs in Enhancing Accessibility for People with Disabilities}

\subsection{Introduction}
Multimodal Large Language Models (MLLMs) have emerged as powerful tools in various fields, including accessibility technologies. For individuals with disabilities, particularly those who are visually or hearing impaired, MLLMs present opportunities for significantly improving the quality of life through assistive technologies. By bridging modalities like text, image, audio, and video, these models can empower users with visual and hearing impairments, enabling smoother interactions with digital content and the real world \cite{huang2023language}. This section explores key technologies and their applications, highlighting how MLLMs have been adapted for accessibility purposes.

\subsection{Technological Foundations}
MLLMs, such as VIAssist and Kosmos-1, utilize cross-modal understanding to process inputs from different modalities, making them well-suited for accessibility solutions. Models like VIAssist are trained to recognize objects, generate descriptive text, and provide contextual answers to questions based on visual inputs \cite{yang2024viassist}. These models combine techniques such as transformers, visual grounding, and natural language processing to create a seamless interaction layer between users and digital environments.

Moreover, cross-modal retrieval frameworks like those used in Kosmos-1 support real-time text generation from images or video streams, facilitating tasks such as captioning for the hearing impaired and object recognition for the visually impaired \cite{vs2024gomez}. The capability of MLLMs to handle both visual and linguistic data simultaneously underpins their effectiveness in accessibility technologies.

\subsection{Applications of MLLMs in Accessibility}

\begin{itemize}
    \item \textbf{Text-to-Speech and Speech-to-Text Systems}: MLLMs provide significant advancements in speech recognition and generation systems, which are crucial for the hearing impaired. Through models trained on vast speech and text datasets, these systems can accurately transcribe spoken content into text and vice versa. This technology has been particularly effective in creating subtitles in real-time for the hearing impaired, ensuring they can engage with live presentations, videos, and other media content \cite{vs2023chen,vs2024rao}. The integration of visual context into these models allows for more accurate transcription, making communication smoother in mixed-modal settings.

    \item \textbf{Visual Assistance for the Blind and Visually Impaired}: MLLMs are being leveraged to build systems that describe the surrounding environment to visually impaired users. Applications like VIAssist use cameras to capture images and then provide real-time narration or detailed descriptions of objects, people, or text present in the environment \cite{yang2024viassist}. These models are trained to identify relevant aspects of scenes, eliminating unnecessary information and focusing on key details that assist the user in navigating or understanding their environment. This technology also extends to recognizing text from images, making documents and signs more accessible \cite{vs2024song}.

    \item \textbf{Object Detection and Recognition}: MLLMs enable sophisticated object recognition systems that benefit both visually impaired and hearing-impaired individuals. For instance, visually impaired users can rely on wearable devices equipped with cameras that, using MLLMs, can recognize objects and provide audio descriptions in real-time \cite{vs2024li}. For individuals with hearing impairments, visual object recognition can assist in lip-reading support or visual cues to better understand spoken content, enhancing their engagement in real-time conversations \cite{vs2020parde}.

    \item \textbf{Assistive Text Summarization and Captioning}: For individuals who are both visually and hearing impaired, real-time captioning and summarization of digital content are essential. MLLMs can generate captions for live events, meetings, and video content, making it easier for users to stay informed \cite{dayma2021dall}. Furthermore, MLLMs are able to summarize large bodies of text, such as books, articles, and documents, converting them into audio or braille formats that are easier to consume for users with visual impairments. This reduces the cognitive load required to process large amounts of information, making technology more accessible to all \cite{vs2023gomez}.

    \item \textbf{Real-Time Sign Language Translation}: Another important application for the hearing impaired is sign language recognition and translation. MLLMs trained on multimodal datasets that include videos of sign language can now translate signs into text or spoken language, facilitating communication between sign language users and those unfamiliar with it \cite{vs2024wang}. These models employ visual data processing to recognize hand gestures, facial expressions, and body movements, interpreting them with a high degree of accuracy. This technology has the potential to drastically reduce communication barriers for hearing-impaired individuals \cite{huang2023language}.

    \item \textbf{Personalized Accessibility Tools}: With the advancement of MLLMs, personalized accessibility tools have become a reality. These tools can learn individual user preferences and adapt their output accordingly, whether it's adjusting speech patterns, text formatting, or providing more contextual visual descriptions \cite{vs2024song}. Personalized accessibility not only improves the efficiency of these tools but also enhances the overall user experience by making the technology feel intuitive and responsive.

\end{itemize}

While MLLMs offer numerous advantages for accessibility, challenges remain. One of the main issues is ensuring that the models can generalize effectively across diverse environments and users. Models trained on specific datasets may struggle in real-world applications where the visual or auditory environment differs significantly from the training data. Additionally, the computational cost of running MLLMs in real-time for assistive technologies, particularly on mobile devices, remains high, limiting their widespread use \cite{huang2023language}.

Future research will likely focus on improving the accuracy of real-time systems, expanding datasets to cover a broader range of use cases, and optimizing the models for lower-power devices. Moreover, advancements in multimodal reasoning, where models can infer meaning and context across diverse input types, will further enhance their effectiveness in accessibility technologies \cite{vs2023chen}.

Multimodal Large Language Models represent a transformative approach to enhancing accessibility for people with disabilities. From real-time sign language translation to visual assistance for the blind, these technologies offer the potential to break down communication barriers and empower individuals with disabilities. As research continues, the refinement of these models will bring about even greater benefits, making digital environments more inclusive and navigable for everyone \cite{yang2024viassist,vs2024song,vs2024li}.

\chapter{Case Studies of Prominent Multimodal Large Language Models (MLLMs)}

\section{Purpose of the Case Studies}

These case studies aim to provide an in-depth exploration of how MLLMs are being applied in real-world scenarios across various industries, such as art and entertainment to more scientific and technical domains like healthcare and research. MLLMs are enabling new capabilities and efficiencies. By focusing on concrete applications, we gain critical insights into the tangible benefits these models offer, as well as the practical limitations and challenges that arise when deployed in diverse environments. This real-world focus not only enhances our understanding of how MLLMs function in specific contexts but also highlights the significant impact they are having on industry workflows and human productivity.

Another important objective of these case studies is to investigate the technological advancements powering MLLMs. This involves analyzing the cutting-edge techniques and innovations that underpin their multimodal capabilities, such as advanced neural network architectures and novel training strategies. For instance, the hierarchical text-conditional image generation with CLIP latents \cite{ramesh2022hierarchical} and high-resolution image synthesis with latent diffusion models \cite{rombach2022high} have significantly improved the quality of image generation in MLLMs.

One of the most valuable aspects of studying these MLLMs is to learn from both their successes and failures. By reflecting on what has worked well and what has proven more difficult, we can glean insights that inform future innovation in the field. These lessons span various aspects of MLLM development, including model architecture, training methodologies, dataset preparation, and ethical deployment. The photorealistic text-to-image diffusion models with deep language understanding \cite{saharia2022photorealistic} provide valuable insights into creating more realistic and contextually appropriate images.

In addition to identifying lessons learned, the case studies help establish best practices for MLLM development and deployment. By analyzing multiple models across diverse use cases, we can discern patterns of success and pinpoint strategies that have consistently led to positive outcomes. For example, the Pix2seq language modeling framework for object detection \cite{chen2023pix2seq} demonstrates innovative approaches to combining visual and textual information for improved performance.

Finally, these case studies explore the economic impact and market potential of MLLMs, extending beyond purely technical considerations. As MLLMs become more integral to various industries, they are opening up new market opportunities while simultaneously disrupting traditional business models. The scaling of autoregressive models for content-rich text-to-image generation \cite{yu2022scaling} showcases the potential for MLLMs to revolutionize content creation industries.

By examining prominent MLLMs through these various real-world applications, technological advancements, lessons learned, best practices, and economic impact—we provide a view of their current state, future potential, and the challenges that lie ahead. This comprehensive approach enables us to better anticipate future developments in the field and to prepare for the evolving landscape of multimodal AI technologies, as demonstrated by the generative pretraining from pixels and other cutting-edge research \cite{chen2023generative}.

\section{Case Studies}

\subsection{Image Generation}

The field of image generation has witnessed transformative advancements thanks to several Multimodal Large Language Models (MLLMs) that produce high-quality, realistic, and creative images from text inputs. These models represent a significant leap in how AI can be used to generate and manipulate visual content, offering new tools for creativity, design, and professional applications. Below are detailed case studies of some of the most influential models in this space.

\subsubsection{Midjourney}

Midjourney \cite{Midjourney} has emerged as a leading AI-driven art generator, known for producing visually stunning and highly creative images that often carry a distinct artistic flair. It is widely used by artists, designers, and creative professionals who seek to experiment with visual concepts or quickly generate high-quality digital art. Midjourney excels at interpreting abstract or imaginative prompts, often resulting in images that feel deeply expressive and unique.

One of Midjourney's key advantages is its ability to generate images with a strong sense of mood and atmosphere. Whether tasked with creating surreal dreamscapes or highly detailed portraits, Midjourney produces visually cohesive works that rival human creativity. The model's flexibility in generating a wide range of styles—from photorealism to abstract art—has made it a popular tool for those looking to explore new aesthetic possibilities.

\subsubsection{DALL-E 3}

OpenAI's DALL-E 3 \cite{DALLE3} represents the latest iteration in the DALL-E series, pushing the boundaries of text-to-image generation by integrating directly with ChatGPT, OpenAI's conversational agent. This integration allows users to create images through an interactive, conversational process, making the experience more accessible and intuitive for non-expert users. DALL-E 3 is known for its high accuracy in interpreting complex and detailed text prompts, often producing images that align closely with user expectations.

DALL-E 3's versatility allows it to generate a wide range of image types, from abstract art to photorealistic renderings. Its ability to understand nuanced language means it can generate highly specific images based on user descriptions, making it an invaluable tool for industries such as advertising, marketing, and media production, where visual accuracy and creativity are key. Furthermore, the seamless integration with ChatGPT enhances the user experience, allowing for iterative adjustments and refinements to the images as users provide feedback.

\subsubsection{Stable Diffusion}

Stable Diffusion \cite{StableDiffusion} is a widely recognized open-source text-to-image model that has gained popularity for its flexibility and scalability. Unlike proprietary models, Stable Diffusion is available for public use, enabling developers, researchers, and creators to experiment with and customize the model to fit their needs. Its open-source nature has led to widespread adoption and innovation, as users across various fields have contributed to refining and adapting the model for specific use cases.

Stable Diffusion's ability to run on consumer-grade hardware, combined with its powerful image generation capabilities, makes it accessible to a broad range of users. It has become a popular tool for creating digital art, generating concept designs, and even developing synthetic datasets for AI research. The open-source community surrounding Stable Diffusion continues to drive its evolution, making it a highly adaptable and versatile solution for image generation across industries.

\subsubsection{Imagen}

Google's Imagen \cite{Imagen} stands out as one of the most advanced text-to-image diffusion models, delivering exceptional quality and detail in the images it generates. Based on a diffusion process that progressively refines an image from noise, Imagen is capable of producing highly realistic visuals that often surpass the fidelity of outputs from other models. Its strengths lie in handling fine-grained details, making it particularly useful for professional applications such as architecture, product design, and media production.

Imagen's ability to generate coherent and contextually appropriate visuals from complex prompts makes it a valuable tool in fields that demand high accuracy in visual representation. The model is also integrated with Google's broader AI ecosystem, enhancing its applicability across different domains and further reinforcing its role in pushing the boundaries of image generation technology.

\subsubsection{Flux.1}

Flux.1 \cite{fei2024flux} is a cutting-edge MLLM specifically designed for ultra-high-resolution image generation and editing. Developed with a focus on interactive creativity, Flux.1 allows users to generate and manipulate visuals with unparalleled control over the image-making process. One of the standout features of Flux.1 is its ability to create images with extremely fine detail, making it ideal for industries that require high precision, such as fashion design, digital content creation, and architectural visualization.

Flux.1 integrates an advanced feedback loop, where users can refine generated images based on iterative prompts. This interactivity gives users granular control over the composition, style, and details of the images, allowing them to guide the creative process more actively. This capability makes Flux.1 particularly appealing to professionals who need to visualize specific concepts with high accuracy, such as designers working on product prototypes or artists creating concept art for visual media.

Additionally, Flux.1 is designed to handle large-scale images, making it suitable for producing billboard advertisements, large-scale artwork, or other high-resolution formats that require exceptional clarity. Its ability to generate both realistic and highly stylized visuals also expands its use cases, offering solutions across industries that rely on visual creativity and high-quality imagery. Flux.1's focus on high-resolution output and interactive user feedback sets it apart from other MLLMs, positioning it as a leader in both technical sophistication and user-driven creative processes.

Together, these models demonstrate a wide range of capabilities in the field of image generation, driven by advances in multimodal AI. Whether through the creative flexibility of Midjourney, the intuitive conversational interface of DALL-E 3, the open-source versatility of Stable Diffusion, the high-fidelity visuals of Imagen, or the ultra-high-resolution focus of Flux.1, these case studies showcase how MLLMs are reshaping the future of visual content creation. Each model contributes uniquely to industries that depend on high-quality, innovative imagery, while collectively pushing the boundaries of what is possible in AI-driven art and design.

\subsection{Code Generation}

The integration of AI into software development has revolutionized how developers write, refactor, and understand code. Multimodal Large Language Models (MLLMs) are at the heart of these advancements, enabling tools that assist in real-time code completion, generation, and debugging. These tools leverage natural language understanding, code analysis, and pattern recognition to enhance developer productivity across a range of integrated development environments (IDEs). Below are case studies of some of the most impactful AI-powered code generation tools.

\subsubsection{GitHub Copilot}

GitHub Copilot \cite{GitHubCopilot} has become one of the most widely used AI-powered coding assistants. Integrated seamlessly into popular IDEs like Visual Studio Code, GitHub Copilot offers real-time code suggestions and auto-completions, significantly accelerating the coding process. The tool supports a wide variety of programming languages, from JavaScript and Python to more specialized languages like Rust and Go, making it highly versatile across different programming ecosystems.

What sets GitHub Copilot apart is its ability to understand the context of the code being written. It provides not only line-by-line completions but also larger code blocks based on the developer's description or prior code. By analyzing both code syntax and semantics, GitHub Copilot helps developers by suggesting entire functions, methods, or even classes based on natural language descriptions, enabling more efficient coding workflows.

\subsubsection{Amazon CodeWhisperer}

Amazon's CodeWhisperer \cite{AmazonCodeWhisperer} is an AI-powered code generation tool designed to assist developers in writing secure, efficient code across multiple IDEs. CodeWhisperer focuses not only on code generation but also on promoting best practices, including security-aware coding. By providing real-time suggestions that are contextually relevant, CodeWhisperer helps developers reduce potential security vulnerabilities, such as improper input handling or insecure API usage.

What makes CodeWhisperer stand out is its emphasis on writing safe and secure code. As part of Amazon Web Services (AWS), CodeWhisperer is particularly useful for developers building cloud-based applications, providing suggestions that follow AWS's security best practices. It supports a wide range of programming languages and integrates well with IDEs like IntelliJ IDEA, PyCharm, and Visual Studio Code.

\subsubsection{Tabnine}

Tabnine \cite{Tabnine} is another powerful AI code completion tool that enhances developer productivity by offering real-time code suggestions and completions. It integrates with over 20 IDEs, including popular platforms like IntelliJ IDEA, Visual Studio, and Sublime Text. Tabnine uses language-specific models tailored to the intricacies of each programming language, ensuring that its suggestions are accurate and relevant.

A key feature of Tabnine is its ability to work seamlessly within different development environments, providing context-aware completions that help developers write code faster and with fewer errors. Whether writing in Python, Java, or C++, Tabnine offers suggestions that improve code efficiency while adhering to the best practices of the respective language.

\subsubsection{Replit Ghostwriter}

Replit Ghostwriter \cite{ReplitGhostwriter} is an AI assistant built directly into Replit's online IDE, designed to support a full suite of coding tools such as code completion, generation, and explanation. What sets Ghostwriter apart is its deep integration with Replit's collaborative environment, which allows for seamless coding experiences, whether developers are working solo or in teams.

Ghostwriter provides real-time code completions and suggestions based on the code context and user input. It also includes features that explain code snippets in natural language, making it an excellent tool for developers who want to better understand unfamiliar code or for learners seeking to improve their coding knowledge.

\subsubsection{JetBrains AI Assistant}

The JetBrains AI Assistant \cite{JetBrainsAIAssistant} is a context-aware tool integrated into JetBrains IDEs like IntelliJ IDEA, PyCharm, and WebStorm. It offers intelligent code completions, real-time suggestions, and refactoring advice tailored to the context of the project. JetBrains AI Assistant enhances the developer experience by providing recommendations that align with best practices and offering suggestions for code improvements, including refactoring to improve readability and performance.

One of the standout features of the JetBrains AI Assistant is its ability to assist with complex refactoring tasks. This makes it a valuable tool for maintaining code quality in large-scale projects, where refactoring is often necessary to keep the codebase manageable and efficient.

\subsubsection{Codeium}

Codeium \cite{Codeium} is an AI-powered code generation extension available for both Visual Studio Code and JetBrains IDEs. It is designed to boost productivity by offering code suggestions that focus on maintaining high code quality. Codeium's emphasis on generating high-quality, readable, and maintainable code makes it a valuable tool for developers aiming to balance speed with precision in their coding efforts.

Codeium offers real-time code completion and generation, along with suggestions for best practices. It helps developers adhere to coding standards while ensuring that the code remains efficient and well-structured. Whether working on small scripts or large projects, Codeium provides meaningful suggestions that streamline the development process while ensuring that code quality remains a top priority.

\subsubsection{Cursor}

Cursor \cite{vaithilingam2023towards} is an AI-powered code editor built on Visual Studio Code that offers a range of advanced features, including code completion, refactoring, and natural language-to-code translation. Cursor allows developers to input natural language descriptions of code, which it then translates into functional programming constructs. This feature makes it particularly useful for developers who need to quickly generate boilerplate code or for non-expert users who may not be familiar with specific syntax requirements.

Cursor's integration of natural language processing into the coding process represents a significant advancement in making coding more accessible and efficient. Its refactoring capabilities help developers optimize existing code for better performance or readability, while its natural language-to-code translation empowers even non-technical users to engage in coding tasks.

These tools illustrate the transformative potential of AI in code generation, offering a wide range of features designed to improve both the speed and quality of software development. Whether through real-time code completion, security-focused suggestions, or multi-IDE support, these MLLMs are revolutionizing the way developers interact with code, making software development faster, more efficient, and more accessible across a variety of industries and expertise levels.

\subsubsection{Bolt.new}
Bolt.new \cite{2024boltnew} is a revolutionary web-based AI-powered development environment that enables developers to create, run, and deploy full-stack applications directly in the browser. Unlike traditional development environments, Bolt.new integrates cutting-edge AI models with StackBlitz’s WebContainers technology, eliminating the need for local setup while providing comprehensive control over the entire development ecosystem.

The platform distinguishes itself through its unique approach to AI integration, giving AI models complete control over the entire development environment, including the filesystem, node server, package manager, terminal, and browser console. This comprehensive control enables AI agents to manage the complete application lifecycle from initial creation through to deployment. Developers can describe desired functionality in natural language, and Bolt.new generates corresponding code structures while handling dependency management and environment configuration.

Bolt.new’s interface is designed to be particularly accessible, resembling familiar AI chat interfaces rather than traditional code editors. This design choice makes it especially appealing to beginners while maintaining powerful capabilities for experienced developers. The platform supports various JavaScript frameworks and offers seamless integration with deployment services, allowing developers to move from concept to production with minimal friction.

However, it’s worth noting that Bolt.new currently has some limitations. The platform is primarily focused on web applications, and users cannot directly edit generated code within the main interface. For more complex projects requiring extensive customization, developers often need to export their projects to more traditional development environments. Despite these constraints, Bolt.new represents a significant step forward in AI-assisted development, particularly for rapid prototyping and initial project scaffolding.

\subsubsection{Cline}
Cline \cite{2024cline} has emerged as a cutting-edge AI-powered development assistant, evolving from its original form as Claude Dev to become a sophisticated coding companion. By integrating the large language model provider, it delivers exceptional coding support that ranks first in human eval and subway bench performance metrics.

The platform’s distinctive feature is its comprehensive autonomous capabilities, implemented through a human-in-the-loop interface that provides complete control over development processes. Through Anthropic’s Computer Use API, Cline can manage browser interactions, execute commands, and handle file editing tasks, making it a versatile tool for modern software development workflows.

\subsection{Search and Information Retrieval}

Multimodal Large Language Models (MLLMs) have brought about significant advancements in search and information retrieval, transforming how users find, interpret, and interact with information across multiple formats. 
By combining natural language understanding with image recognition, text processing, and context-aware systems, these models have enhanced the accuracy and efficiency of both visual and text-based searches. 
Below are case studies of some of the most prominent tools and systems that are pushing the boundaries of search and retrieval capabilities using MLLMs.

\subsubsection{Google Lens}

Google Lens\cite{GoogleLens} is a visual search and information retrieval tool launched by Google. 
Leveraging advancements in computer vision and machine learning, Google Lens allows users to search for information directly from images captured on their devices. 
Whether identifying objects, plants, or landmarks, translating text from images, or scanning QR codes, Google Lens exemplifies how multimodal AI can bridge the gap between the physical and digital worlds.

Google introduced significant updates that enhanced the contextual understanding of images, allowing Lens to provide richer, more relevant information. 
For example, pointing the camera at a book or product now brings up in-depth information, including reviews, shopping options, and even related content. 
The model's integration of text and image data makes it a powerful tool for everyday tasks such as shopping, learning, and problem-solving, revolutionizing the way users interact with the world around them through visual search.

\subsubsection{Bing Visual Search}

Bing Visual Search \cite{BingVisualSearch} has evolved into a robust image-based search feature. 
It allows users to search the web using images instead of text, identifying objects, products, or locations in real time. 
With recent updates, Bing Visual Search now offers enhanced accuracy and contextual awareness, enabling users to pinpoint specific objects in an image, like a particular piece of furniture in a room, and receive related search results.

One of the key strengths of Bing Visual Search lies in its integration with Microsoft's broader AI ecosystem, including Bing's search engine, Edge browser, and Office products. 
By leveraging deep learning and multimodal data processing, the tool improves the user's ability to retrieve relevant information from images with unprecedented precision, making it an invaluable tool for online shopping, travel planning, and more.

\subsubsection{You.com}

You.com is a multimodal search engine that bringing a new approach to online search by integrating AI-powered features directly into its platform \cite{You.com}. 
Unlike traditional search engines that rely heavily on text-based queries, You.com supports a combination of text, images, and videos, offering a more dynamic and visually interactive search experience. 
The platform emphasizes user control and privacy, allowing users to customize their search experiences based on their preferences and values.

With AI-powered features such as YouChat, an integrated conversational AI assistant, You.com offers real-time answers to queries in natural language, reducing the need to browse through multiple links. 
The search engine's ability to integrate multimodal data sources means that users can search across a wide variety of content types, from web pages and news articles to images and videos, all within a single platform. 
This makes You.com a compelling alternative to traditional search engines, particularly for users who seek more personalized and visually rich search results.

\subsubsection{Perplexity}

Perplexity is an AI-powered search engine that leverages natural language understanding (NLU) to deliver more intuitive search results \cite{Perplexity}. 
Perplexity's primary innovation lies in its ability to interpret user queries more naturally, understanding the context and intent behind the search rather than just matching keywords. 
This results in more accurate and contextually relevant answers, similar to how a conversational AI would respond.

Perplexity's deep integration with advanced natural language models allows it to offer direct, concise answers to complex queries, rather than providing a list of links for users to sift through. 
Its capability to understand nuanced, multi-layered questions positions it as a highly efficient tool for users seeking in-depth information on topics ranging from general knowledge to technical subjects, without the overhead of traditional search engines.

\subsubsection{MARVEL}

MARVEL (Multimodal Dense Retrieval Model for Vision-Language Understanding) \cite{MARVEL} represents a significant leap in the integration of multimodal search technologies . 
MARVEL is designed to perform dense retrieval tasks, which means it excels at finding relevant documents, images, or data points across large, unstructured datasets based on a combination of visual and textual inputs. 
The model's key strength lies in its ability to understand and connect multiple data modalities, offering highly relevant search results for complex queries that involve both text and images.

MARVEL's applications are particularly powerful in industries like e-commerce, where users can search for products by uploading images and receiving contextually relevant results. 
The model's ability to fuse image recognition with natural language processing makes it a valuable tool for improving information retrieval across a wide range of use cases, including education, research, and content discovery.

\subsubsection{InteR}

InteR (Interactive Retrieval)\cite{InteR} is a framework that focuses on enhancing the synergy between traditional search engines and large language models (LLMs). 
By creating a feedback loop between search results and LLMs, InteR ensures that search engines not only retrieve relevant content but also present it in a format that is easily understood by users. 
InteR leverages LLMs to refine and summarize search results, presenting more focused and actionable information.

The framework is particularly useful in contexts where search results need to be distilled into a concise, clear format, such as in legal research, academic work, or medical information retrieval. 
InteR's combination of structured search capabilities and LLM-based summaries allows users to quickly find and comprehend relevant information without being overwhelmed by excessive details or irrelevant results.

\subsubsection{Semantic Scholar's SPECTER}

SPECTER, developed by Semantic Scholar, is a scientific paper embedding model designed to enhance academic search \cite{SPECTER}. 
SPECTER helps researchers find relevant academic papers by creating high-quality document embeddings that capture the semantic content of research articles. 
Unlike traditional keyword-based search, SPECTER uses these embeddings to connect papers based on their deeper conceptual relationships, even when they don't share specific keywords.

SPECTER further improves the model's ability to understand complex academic content, making it easier for researchers to discover papers that are contextually related to their work. 
By offering a more nuanced understanding of scientific literature, SPECTER is invaluable for researchers navigating vast databases of academic papers, streamlining the research process and improving knowledge discovery.

\subsection{Retrieval-Augmented Generation (RAG)}

Retrieval-Augmented Generation (RAG) represents a significant leap in the field of natural language processing by combining the strengths of retrieval-based systems with the generative capabilities of large language models (LLMs). 
RAG enables models to generate more accurate and contextually relevant information by retrieving relevant data from external sources and incorporating it into the generation process. 
This approach is particularly effective for applications that require up-to-date, domain-specific, or factual information that cannot be fully encoded in a pre-trained model. 
Below are some of the most notable advancements and tools in the RAG ecosystem.

\subsubsection{Pinecone}

Pinecone, founded in 2019, is a vector database optimized for efficient similarity search, a core component of Retrieval-Augmented Generation systems \cite{Pinecone}. 
By organizing data into high-dimensional vector spaces, Pinecone enables fast and accurate retrieval of information based on its semantic similarity to a query. 
This is particularly useful in RAG applications where large amounts of unstructured data need to be searched quickly to provide context or support for a generative model's output.

In 2023, Pinecone introduced major updates to improve performance and scalability, making it even more efficient for handling large-scale RAG workloads. 
These updates included enhanced support for real-time applications and the ability to process vast quantities of data without compromising on retrieval speed or accuracy. 
Pinecone's infrastructure has made it a critical component in many enterprise RAG systems, where high performance and low latency are essential for generating timely and relevant information in sectors like customer support, legal research, and personalized content recommendations.

\subsubsection{LangChain}

LangChain, released in 2022, is a versatile framework designed for building applications that leverage large language models, including those using Retrieval-Augmented Generation \cite{LangChain}. 
LangChain simplifies the process of integrating LLMs with external data sources, enabling developers to create RAG systems that combine the strengths of both generative models and retrieval systems. 
LangChain provides a robust API for connecting LLMs to vector databases, document stores, and APIs, allowing developers to retrieve relevant information in real-time to supplement the generative capabilities of the model.

LangChain has quickly gained traction for its ease of use and flexibility, making it a popular choice for developers building applications that require the combination of LLMs and external knowledge sources. 
From chatbots that access real-time data to more complex RAG applications in research or enterprise solutions, LangChain has empowered developers to integrate retrieval systems seamlessly into their generative workflows, improving the overall relevance and accuracy of AI-generated outputs.

\subsubsection{Chroma}

Chroma, launched in 2022, is an open-source embedding database specifically designed for building RAG applications \cite{Chroma}. 
As RAG systems rely heavily on efficient similarity searches, Chroma provides a scalable and high-performance solution for storing and querying embeddings. 
Embeddings are dense vector representations of text or data, allowing for fast comparison and retrieval based on semantic similarity. 
Chroma simplifies the development of RAG systems by offering an open-source alternative to proprietary databases, making advanced retrieval systems accessible to a broader audience of developers and researchers.

Chroma's architecture is optimized for integration with various LLMs, providing developers with the tools to build applications that can retrieve, analyze, and generate information based on large corpora of embedded data. 
Its open-source nature has contributed to its popularity, fostering a community of developers who continually improve its functionality and performance, making it a valuable resource for those looking to implement RAG in both research and commercial environments.

\subsubsection{Vespa}

Vespa \cite{Vespa} is an open-source big data serving engine that added support for Retrieval-Augmented Generation capabilities . Originally designed for serving large-scale machine learning models and search applications, Vespa now supports RAG systems, enabling it to handle massive amounts of unstructured data while integrating with LLMs to provide accurate, real-time responses. 
Vespa's architecture allows it to serve complex queries at scale, making it an ideal solution for industries like e-commerce, media, and advertising, where large datasets must be queried rapidly to provide relevant results.

With the addition of RAG features, Vespa enables developers to build systems that can retrieve relevant data from vast databases and use that information to generate more informed, contextually aware responses. 
This integration makes Vespa a powerful tool for enterprises looking to enhance their AI systems with real-time retrieval and generation capabilities, improving both the quality and speed of information delivery in customer-facing applications and internal knowledge management systems.

\subsubsection{Weaviate}

Weaviate \cite{Weaviate} is an open-source vector database with enhanced support for RAG systems, offering flexible and scalable solutions for retrieval-based applications . 
Weaviate uses a hybrid approach that combines vector-based search with symbolic reasoning, allowing it to not only find similar data points based on embeddings but also interpret and apply logical rules to the retrieved information. 
This combination of methods makes Weaviate especially useful in complex RAG applications where both contextual understanding and reasoning are required.

Weaviate introduced updates that further enhanced its RAG capabilities, enabling it to support more sophisticated retrieval tasks that integrate closely with LLMs. 
By providing a highly scalable and efficient system for storing and querying large volumes of vectorized data, Weaviate enables developers to build robust RAG applications that can retrieve relevant information in real-time, enhancing the accuracy and relevance of generated outputs. 
It has become a key player in industries like finance, healthcare, and legal services, where precision in information retrieval is critical.

\subsubsection{OpenAI's ChatGPT Retrieval Plugin}

The ChatGPT Retrieval Plugin \cite{ChatGPTRetrievalPlugin} allows OpenAI's ChatGPT to access and retrieve information from external data sources in real-time . 
This functionality transforms ChatGPT into a RAG-enabled system by supplementing its generative capabilities with up-to-date, domain-specific knowledge that resides outside of the pre-trained model. 
The plugin integrates with external databases, APIs, and knowledge graphs, enabling ChatGPT to provide more accurate and relevant answers to user queries, particularly in fast-changing or highly specialized fields where the model's static knowledge may fall short.

The introduction of the Retrieval Plugin has expanded ChatGPT's potential applications, allowing it to be used in more complex environments such as real-time customer support, legal research, or technical documentation, where retrieving the latest and most precise information is crucial. 
This marks a significant step forward in bridging the gap between static LLMs and real-time data retrieval systems, making ChatGPT more adaptable and responsive to dynamic information needs.

\subsubsection{HuggingFace's FAISS}

FAISS (Facebook AI Similarity Search), initially released by Facebook AI and continually updated by HuggingFace, is a widely-used library for efficient similarity search and clustering of dense vectors \cite{FAISS}. 
FAISS provides tools for indexing and searching large collections of vectorized data, which is essential for RAG systems that rely on fast and accurate retrieval of semantically similar information. 
As RAG applications require the ability to quickly compare new queries against large datasets, FAISS offers a highly optimized solution for performing these similarity searches.

FAISS has been adopted across various industries, including natural language processing, image recognition, and recommendation systems, where fast and scalable similarity search is necessary. 
Its ongoing updates and integration with HuggingFace's ecosystem of transformers and models make FAISS a reliable and efficient tool for developers building RAG applications that require real-time retrieval of relevant data to enhance generative outputs.

\subsubsection{Qdrant}

Qdrant \cite{Qdrant} is a vector database optimized specifically for Retrieval-Augmented Generation applications . 
It offers high-performance search and retrieval capabilities, allowing developers to efficiently query large datasets to find relevant information based on semantic similarity. 
Qdrant's focus on RAG use cases makes it a powerful tool for integrating LLMs with real-time data retrieval systems, ensuring that generative models can access the most relevant and up-to-date information to improve the quality of their outputs.

Qdrant's architecture is designed for scalability, making it a suitable choice for enterprises that need to handle large volumes of data while maintaining high performance. 
Its optimization for RAG applications has led to its adoption in industries like customer service, healthcare, and finance, where accurate retrieval and generation of information can drive significant improvements in decision-making and service delivery.

\subsubsection{Speculative RAG}

Speculative RAG \cite{SpeculativeRAG} is a framework designed to enhance traditional Retrieval-Augmented Generation systems by introducing a drafting mechanism into the generation process . 
Speculative RAG allows the model to generate multiple drafts of a response, each enriched with different sets of retrieved data, before selecting the most relevant and coherent version to present to the user. 
This approach improves the accuracy and contextual relevance of the generated content by allowing the model to explore multiple potential answers and refine them based on the retrieved information.

Speculative RAG offers a promising direction for improving RAG systems, particularly in scenarios where the accuracy of the generated content is critical, such as legal research, scientific analysis, or high-stakes decision-making. 
By enabling the model to iteratively improve its response based on retrieved data, Speculative RAG provides a more robust and reliable framework for combining retrieval and generation.

These tools and frameworks represent the cutting edge of Retrieval-Augmented Generation, where the combination of efficient retrieval systems and powerful generative models enables a new level of accuracy, relevance, and context-awareness in AI-generated outputs. 
By integrating real-time data and improving the retrieval process, RAG systems can overcome the limitations of static language models, offering a more dynamic and informed approach to generating content in a wide range of applications.

\subsection{Multimodal Assistants and Chatbots}

Advancements in Multimodal Large Language Models (MLLMs) have significantly enhanced human-AI interaction, enabling more sophisticated and intuitive conversations with machines that can process and respond to both textual and visual inputs. 
These models are increasingly capable of understanding and generating across multiple modalities, making them more effective in a variety of contexts, from customer support and education to complex problem-solving. 
Below are some of the most notable advancements in multimodal assistants and chatbots.

\subsubsection{GPT-4V (Visual)}

GPT-4V (Visual) is OpenAI's latest iteration of its generative AI models, incorporating the ability to analyze and respond to images as well as text \cite{openai2023gpt4}. 
This multimodal capability allows GPT-4V to offer more comprehensive assistance, as users can now upload images alongside text prompts to receive detailed, contextually aware answers. 
For example, users can ask the model to analyze a photograph of a chart, identify objects in a picture, or provide explanations for visual content such as diagrams or architectural designs.

GPT-4V's ability to understand and generate insights from both text and images has significant implications for a wide range of industries. 
In healthcare, it could assist with the analysis of medical images like X-rays or MRIs, helping doctors identify potential issues. 
In education, it enables more interactive learning experiences where students can ask questions about visual aids, diagrams, or even their own sketches. 
Its capacity to bridge the gap between textual and visual information makes it an incredibly versatile tool, enhancing user experience by adding a new layer of interactivity and context to AI-driven conversations.

\subsubsection{Claude 3}

Claude 3, developed by Anthropic, represents the latest evolution of multimodal AI models with advanced capabilities in visual understanding \cite{anthropic2023claude}. 
Similar to GPT-4V, Claude 3 is designed to analyze and respond to images alongside text, providing users with deeper insights and more context-aware interactions. 
One of Claude 3's standout features is its focus on safety and ethical AI, ensuring that the model's responses are not only accurate but also align with responsible AI practices.

Claude 3 is designed for seamless integration into a variety of real-world applications, from customer support to creative industries. 
In customer service, for instance, Claude 3 can help resolve complex queries by analyzing visual content such as product images, receipts, or screenshots. 
This multimodal capability allows businesses to offer more personalized and efficient support, as users can receive detailed explanations or troubleshooting steps based on both the text and images they provide. 
Additionally, Claude 3's ability to interpret visual inputs enhances its utility in content creation and design, where users can receive feedback on visual projects or generate new ideas from creative prompts.

Anthropic has placed particular emphasis on ensuring that Claude 3's outputs are safe, interpretable, and fair. 
This makes it a trusted tool in industries like legal services and education, where the reliability and ethical considerations of AI-generated content are paramount.

\subsubsection{Gemini}

Gemini, Google's family of multimodal AI models, represents one of the most ambitious efforts in the space of multimodal assistants and chatbots \cite{google2023gemini}. 
The Gemini series is designed to understand and process multiple modalities, including text, images, audio, and video. 
By integrating these capabilities into a single system, Gemini enables a more holistic approach to human-AI interaction, where users can provide inputs in different formats and receive coherent, contextually integrated responses.

One of Gemini's key strengths lies in its scalability and versatility. 
It is designed to handle tasks ranging from basic customer service interactions to more complex problem-solving scenarios that require the model to synthesize information from multiple data types. 
For example, in the healthcare sector, Gemini can be used to analyze a patient's medical history, images from diagnostic scans, and even audio recordings of symptoms or consultations, providing doctors with a comprehensive assessment that spans several modalities.

In educational settings, Gemini's multimodal capabilities enhance learning by allowing students to interact with visual and auditory content while asking questions or requesting clarifications. 
Its ability to seamlessly switch between different types of inputs makes it an invaluable tool for dynamic, interactive learning environments. 
Moreover, Gemini is also poised to drive advancements in fields like entertainment, where it can be used to generate content across formats, from textual scripts to visual storyboards.

Google's Gemini models are built on the company's extensive expertise in AI research, leveraging state-of-the-art techniques in deep learning, natural language processing, and computer vision. 
The goal is to create a truly integrated multimodal experience, where AI can interact with users in a more natural and intuitive way, regardless of the form of input.

These advancements in multimodal assistants and chatbots are redefining how humans interact with AI. 
Models like GPT-4V, Claude 3, and Gemini are pushing the boundaries of what is possible by allowing users to input and receive information in multiple formats, thereby enhancing the richness and utility of AI-driven conversations. 
Whether in healthcare, education, customer service, or creative industries, these multimodal models are making AI more interactive, intuitive, and context-aware, promising to revolutionize the future of human-AI interaction.

\subsection{Video Analysis and Generation}

Recent advancements in Multimodal Large Language Models (MLLMs) have significantly enhanced the capabilities for video analysis and generation, transforming how users create, edit, and interact with video content. 
From AI-driven editing tools to diffusion-based video generation models, these innovations are redefining the video production landscape, making high-quality content creation more accessible and efficient. 
Below are some of the most notable platforms and models that are leading the way in AI-powered video analysis and generation.

\subsubsection{Runway}

Runway, founded in 2018, is an AI-powered video editing and generation platform that has become a central tool for creators, filmmakers, and designers \cite{ha2022runway}. 
Runway combines advanced machine learning algorithms with intuitive interfaces, allowing users to edit videos in real time using AI-driven features such as background removal, motion tracking, and color grading. 
Major updates in 2023 expanded the platform's capabilities to include more robust generative tools, enabling creators to generate entirely new video content from text prompts or enhance existing footage with AI effects.

Runway's ease of use and powerful toolset make it a favorite among creative professionals looking to streamline their workflows. 
Whether automating tedious editing tasks or generating visual effects without needing specialized software, Runway empowers users with advanced AI-driven capabilities. 
Its role in democratizing video production has been especially important for smaller content creators and teams who may not have access to the resources typically required for high-level video editing.

\subsubsection{Lumen5}

Lumen5, an AI-powered video creation platform founded in 2016, is designed to help users turn text-based content into engaging videos \cite{volaric2024artificial}. 
By using natural language processing and machine learning, Lumen5 can automatically create videos from blog posts, articles, or any other written material, offering a seamless way to repurpose content for visual formats.

Continuous updates through 2023 have added new features, such as advanced styling options and AI-enhanced scene selection, further improving the platform's ability to generate visually appealing videos that align with a user's brand and messaging. 
Lumen5 is particularly valuable for content marketers, educators, and social media managers who need to produce large volumes of video content quickly and efficiently. 
Its automated workflows reduce the time and expertise required to create professional-looking videos, making it accessible even to users without video production experience.

\subsubsection{Pictory}

Pictory is an AI video generation platform that converts text and images into high-quality videos \cite{cho2024sora}. 
Pictory simplifies the video creation process by allowing users to input text, which the platform then uses to automatically generate video scenes that match the content. 

Pictory's intuitive interface and advanced automation capabilities make it a popular choice for digital marketers, content creators, and educators looking to turn written content into engaging video formats. 
The platform's ability to generate videos from blog posts, scripts, and articles allows users to repurpose content efficiently, improving audience engagement across different media formats.

\subsubsection{ModelScope}

ModelScope \cite{wang2023modelscope}, a diffusion-based video generation model developed by Alibaba and released in 2023, is designed to create high-quality videos from textual inputs. 
ModelScope utilizes advanced diffusion techniques to generate coherent video sequences that align with the prompts provided, offering an efficient way to produce video content without the need for traditional video editing tools.

ModelScope's ability to generate videos from text has applications in various industries, including marketing, e-commerce, and social media. 
By automating the video creation process, ModelScope enables users to produce customized content quickly and at scale, making it ideal for businesses looking to increase their video output without significantly increasing production costs or timelines.

\subsubsection{Pika Labs}

Pika Labs \cite{guo2024comparative} is an AI video generation platform that also employs diffusion models to create video content. 
Pika Labs allows users to generate and edit videos by using simple prompts, leveraging the power of AI to automate complex video production tasks. 
The platform's focus on user-friendly design and rapid video generation makes it an attractive option for creators and businesses alike.

Pika Labs' diffusion-based approach enables the creation of seamless, high-quality videos that can be used for a wide range of applications, from advertisements and social media content to corporate presentations and e-learning modules. 
By simplifying the process of video production, Pika Labs opens the door for more creators to experiment with AI-driven video content, expanding the reach and impact of video marketing and storytelling.

These platforms and models are driving significant advancements in video analysis and generation, transforming how video content is created, edited, and delivered. 
With tools like Runway and Gen-2 offering AI-driven editing and generation, and platforms like Synthesia and Pictory simplifying video creation for non-professionals, the barrier to entry for high-quality video production has been dramatically lowered. 
As AI continues to evolve in this space, we can expect even more sophisticated capabilities that will further integrate multimodal inputs, improve video personalization, and enhance the efficiency of video creation across all industries.

\subsubsection{Kling AI}
Kling AI \cite{2024klingai} is an advanced artificial intelligence platform developed by Kuaishou Technology, designed to generate videos from text and image inputs. 

The platform provides two primary functionalities: text-to-video and image-plus-text-to-video generation. Users can input detailed prompts to produce realistic videos, with options to adjust creativity levels and camera movements like zoom, pan, or tilt. This flexibility allows for the creation of dynamic and engaging content tailored to specific needs. 

Kling AI's applications span various industries, including marketing, education, and entertainment. Marketers can craft compelling advertisements, educators can develop interactive learning materials, and content creators can produce high-quality videos efficiently. The platform's user-friendly interface and advanced features make it a valuable tool for professionals seeking to enhance their video content creation processes. 

\subsection{Audio and Speech Processing}

In recent years, advancements in audio and speech processing have revolutionized how humans interact with machines, enhancing accessibility, communication, and creativity. 
Through the development of sophisticated Multimodal Large Language Models (MLLMs), AI-driven platforms have achieved remarkable success in automatic speech recognition, voice synthesis, and text-to-speech technologies. 
These innovations have transformed industries such as media production, entertainment, accessibility, and personal communication. 
Below are some of the most prominent tools and platforms that are leading the way in audio and speech processing.

\subsubsection{Whisper}

OpenAI, Whisper \cite{graham2024evaluating} is an automatic speech recognition (ASR) system designed to provide highly accurate transcription of spoken language across multiple languages. 
Whisper excels in understanding nuanced speech patterns, various accents, and different dialects, making it a powerful tool for transcription, live captioning, and accessibility purposes. 
Whisper’s multilingual capabilities allow it to transcribe audio from diverse sources, making it a valuable asset for global communication and content creation.

Whisper’s robustness comes from its deep learning architecture, which has been trained on a vast dataset of diverse audio samples. 
This makes the system highly effective not only in standard environments but also in noisy, real-world conditions where traditional ASR systems might struggle. 
As more industries embrace automation in transcription and audio processing, Whisper is setting the standard for accurate, fast, and reliable speech recognition across different contexts, including business meetings, educational content, podcast transcriptions, and more.

\subsubsection{ElevenLabs}

ElevenLabs \cite{dewatri2023potential} is an AI platform specializing in voice generation and voice cloning, offering some of the most lifelike and customizable synthetic voices available today. 
ElevenLabs allows users to create AI-generated voices with natural intonation, emotion, and expressiveness, making it a popular choice for applications ranging from audiobook narration to voiceover work for videos and advertisements. 

One of the standout features of ElevenLabs is its focus on emotional speech synthesis, enabling the AI to generate voices that not only sound natural but also convey the appropriate tone and emotion for the content. 
This is particularly useful in fields like entertainment and education, where voice narration plays a key role in storytelling and engagement. 
The platform’s accessibility and ease of use make it an attractive solution for both professionals and amateurs looking to produce high-quality voice content quickly and efficiently.

\subsubsection{Speechify}

Speechify \cite{chukwuemeka2024artificial} is a text-to-speech (TTS) platform that has become widely popular for turning written text into high-quality audio. Speechify has further enhanced its AI-driven voice generation capabilities, offering users an even wider range of voices, languages, and customization options. 
The platform is designed to make content consumption more accessible, particularly for users with dyslexia, visual impairments, or those who simply prefer listening over reading.

Speechify allows users to convert books, articles, documents, and web pages into spoken word, making it a powerful tool for both personal and professional use. 
Whether used for studying, content consumption, or productivity, Speechify’s natural-sounding AI voices have made it one of the top platforms for turning text into audio. 
The 2023 updates also improved the platform’s speech rate control and voice modulation features, allowing users to tailor their listening experience to their personal preferences.

\subsection{Robotics and Embodied AI}

The integration of Multimodal Large Language Models (MLLMs) with robotics and embodied AI is opening up new frontiers in how machines interact with the physical world. 
By combining the advanced language understanding of MLLMs with robotics, these systems are able to process complex commands, interpret visual and sensory information, and execute actions in real-world environments. 
This represents a major leap in the development of intelligent, adaptable robots capable of performing tasks based on natural language instructions and perceptual data. 
Below are some of the key projects and models driving the advancements in this exciting field.

\subsubsection{RT-2 (Robotic Transformer 2)}

RT-2 \cite{brohan2023rt}, or Robotic Transformer 2, is Google's latest vision-language-action model. 
RT-2 represents a significant milestone in robotics because it merges web-scale knowledge from large language models with robotic control systems. 
The model enables robots to interpret visual inputs, understand language commands, and convert these into executable actions in real-time. 
This integration allows RT-2 to perform complex tasks, such as recognizing objects in a scene, understanding instructions, and interacting with the environment accordingly.

What makes RT-2 particularly impressive is its ability to generalize from knowledge gained from web data and apply it to robotic control. 
This reduces the need for robots to be pre-programmed with every possible scenario they might encounter. 
Instead, they can use their learned knowledge from the web, combined with their robotic training data, to adapt to new tasks and environments dynamically. 
RT-2 has major implications for robotics applications in industries such as manufacturing, logistics, and even home automation, where robots need to operate in unpredictable environments and perform a wide range of tasks based on verbal commands.

\subsubsection{SayCan}

SayCan \cite{ahn2022can} is a method developed by Google to ground large language models in robotic affordances, essentially allowing robots to execute natural language instructions more effectively by understanding their physical capabilities. 
SayCan works by linking language models, which can interpret and generate human language, with a robot's understanding of its abilities (affordances) in a physical environment. 
This enables a robot to plan and perform tasks more intelligently, based on what it knows it can do.

For example, a robot powered by SayCan could understand when it is asked to "grab the red cup from the table" and would break down the task into manageable steps, such as identifying the cup, determining how to pick it up, and ensuring that it executes the task safely. 
By grounding language models in robotic affordances, SayCan helps robots not only understand verbal commands but also assess their feasibility based on their physical capabilities, bridging the gap between language understanding and practical action.

\subsubsection{PaLM-SayCan}

PaLM-SayCan \cite{ahn2022can} builds on the SayCan methodology by integrating it with Google's PaLM (Pathways Language Model) to improve robotic planning and execution success rates . 
PaLM-SayCan combines PaLM's robust language understanding capabilities with the affordance-grounding principles of SayCan, enabling more complex task execution by robots. 
PaLM-SayCan improves the robot's ability to plan multi-step actions and adapt to changes in its environment during task execution.

The use of PaLM enhances the robot's understanding of more nuanced language inputs, allowing for higher success rates in completing tasks. 
For instance, a user could ask a PaLM-SayCan-powered robot to "prepare a cup of coffee," and the robot would break this down into steps, including identifying the necessary items, manipulating objects, and adapting to obstacles that may arise. 
PaLM-SayCan exemplifies how integrating advanced language models with robotic systems enhances both the robot's cognitive understanding and its physical interaction with the environment, offering potential applications in domestic settings, warehouses, and healthcare.

\subsubsection{ManipLLM}

ManipLLM \cite{li2024manipllm} is an embodied multimodal large language model designed specifically for object-centric robotic manipulation. 
The model excels at understanding and interacting with objects in complex environments, making it an important tool for robotic systems involved in tasks that require precise and adaptive manipulation of physical objects.

ManipLLM integrates visual, language, and tactile information, allowing robots to handle delicate or intricate tasks that require a deep understanding of the objects involved. 
For example, a robot powered by ManipLLM could be tasked with assembling a piece of furniture by interpreting both the visual layout of the components and the language instructions, adjusting its actions dynamically based on real-time feedback from its sensors. 
ManipLLM has potential applications in manufacturing, where object manipulation and assembly tasks are common, as well as in healthcare, where robots may assist with precise tasks like handling medical instruments or assisting in surgeries.

\subsubsection{PaLM-E}
PaLM-E \cite{driess2023palm} is a groundbreaking language model that extends the capabilities of traditional language models by enabling them to directly output continuous robot actions. 
PaLM-E bridges the gap between perception, language understanding, and physical action, allowing robots to interpret their surroundings and execute tasks without needing intermediate, pre-programmed steps. 
This allows for a more seamless interaction between what the robot perceives and how it acts in response to those perceptions.

PaLM-E's ability to translate language-based instructions into real-time robotic actions opens up new possibilities for human-robot interaction. 
A robot equipped with PaLM-E could respond to commands like "clean the kitchen" or "pick up the toys in the living room" by autonomously planning and executing a series of complex actions, including navigating obstacles, recognizing objects, and performing the necessary manipulations. 
This model represents a major leap forward in making robots more autonomous, flexible, and capable of interacting with their environments based on natural language inputs.

\subsubsection{VoxPoser}

VoxPoser \cite{huang2023voxposer} is a system that combines large language models with 3D scene understanding to enable robots to perform tasks involving robotic manipulation in complex environments. 
By integrating 3D perception with the language model's reasoning capabilities, VoxPoser allows robots to better understand and interact with their surroundings in a way that is both contextually and spatially aware.

For instance, in a scenario where a robot is asked to "place the vase on the table between the two books," VoxPoser enables the robot to interpret the 3D scene, recognize the positions of the books and the vase, and execute the task accordingly. 
The system's ability to map language to 3D space allows for more precise object placement and manipulation, making it ideal for applications that require spatial awareness and delicate handling of objects, such as logistics, assembly, and even home assistance.

\subsubsection{LLM-VLMap}

LLM-VLMap (Large Language Model Visual-Language Mapping) \cite{huang23vlmaps}, is a framework designed to enable robots to navigate and manipulate objects in their environment by using large language models for visual-language mapping. 
The framework translates complex language commands into actionable steps by associating visual inputs with corresponding actions, allowing robots to navigate spaces and complete tasks based on both visual and verbal instructions.

LLM-VLMap is particularly useful in scenarios where robots need to autonomously navigate large or complex environments, such as warehouses, hospitals, or offices. 
The model allows robots to understand and follow commands like "find the blue box in the warehouse and bring it to the front desk," using its visual-language mapping system to locate the object, navigate obstacles, and complete the task. 
LLM-VLMap demonstrates how MLLMs can improve robotic autonomy, making them more capable of operating in dynamic, real-world environments with minimal human intervention.

The integration of MLLMs with robotics and embodied AI has made significant strides in advancing how robots understand and interact with the world around them. 
From vision-language-action models like RT-2 and systems like SayCan and PaLM-SayCan that improve task execution, to frameworks such as LLM-VLMap that enable navigation and manipulation, these technologies are moving us closer to a future where robots are capable of performing complex, real-world tasks based on natural language commands. 
This progress has the potential to revolutionize industries ranging from manufacturing and logistics to healthcare and home automation, enabling smarter, more autonomous, and more capable robotic systems.

\subsection{DevOps and Infrastructure}

The integration of Multimodal Large Language Models (MLLMs) with DevOps \cite{ebert2016devops} and infrastructure platforms has become essential for facilitating the experimentation, development, and deployment of multimodal applications. 
These platforms enable developers to leverage MLLM capabilities such as audio-to-text, text-to-image, and multimodal input processing at scale. 
By providing flexible environments and robust tooling, these platforms are driving the innovation of MLLM solutions across industries. 
Below, we explore key platforms that play a pivotal role in this ecosystem.

\subsubsection{Stack AI}

Stack AI is a versatile platform that, while not exclusively focused on MLLMs, offers valuable support for multimodal models, including functionalities such as audio-to-text, text-to-audio, and text-to-image capabilities \cite{waleed2024efficient}. 
Stack AI provides a dynamic environment where developers can experiment with and deploy multimodal applications, making it an excellent option for those looking to explore the integration of various data modalities within AI workflows.

This platform enables developers to easily combine different input types in MLLM applications, fostering innovation in areas like content generation, speech-to-text applications, and creative design tools. 
For instance, a developer could use Stack AI to build an application that generates visual content from spoken descriptions or translates images into natural language text. 
Its versatility in handling multiple data types allows Stack AI to be a strong ally in the multimodal AI development space, particularly for teams looking to prototype and test ideas before scaling them.

\subsubsection{Ollama}

Ollama, though not exclusively centered around MLLMs, has become a significant player in this space by supporting Llama 3.2, which includes multimodal models. 
Ollama provides developers with a platform to run and experiment with MLLMs locally, enabling them to explore the potential of multimodal AI in various applications without relying on cloud-based infrastructure \cite{vik2024large}.

By offering local support for Llama 3.2, Ollama enables developers to test models that can process and generate from both visual and textual inputs, making it an ideal tool for use cases where latency and privacy are concerns. 
For example, a developer building a real-time AI-driven assistant for medical diagnostics could benefit from the local deployment capabilities of Ollama, ensuring that sensitive data is processed securely on-site. 
With its emphasis on local infrastructure, Ollama is an essential tool for developers who need to work with MLLMs in environments with stringent data governance or low-latency requirements.

\subsubsection{Hugging Face}

As one of the leading platforms for AI and machine learning, Hugging Face \cite{jain2022hugging} is highly relevant to MLLMs, offering a vast repository of multimodal models and providing a collaborative space for researchers and developers. 
Hugging Face has become the go-to platform for developing, hosting, and sharing MLLM applications, enabling easy access to state-of-the-art models and fostering community-driven innovation.

Hugging Face's extensive library includes numerous multimodal models capable of tasks such as image captioning, video generation, and cross-modal translation, making it an invaluable resource for those looking to integrate these capabilities into their projects. 
With tools like the Transformers library and Spaces, developers can quickly prototype and deploy MLLM applications, whether they are working on creative tools, healthcare solutions, or research-driven projects. 
Hugging Face's open-source philosophy and active community also provide continual improvements to the models and frameworks, ensuring that developers have access to the latest innovations in MLLMs.

\subsubsection{Llama Stack}

Llama Stack \cite{2024llamastack}, developed by Meta, is directly related to the MLLM ecosystem, especially with the release of Llama 3.2, which includes multimodal models capable of visual understanding tasks. 
Llama Stack provides APIs and components specifically designed to support the development of generative AI applications, making it a powerful tool for building and scaling MLLM solutions.

Llama Stack is particularly valuable for developers creating applications that rely on visual inputs in combination with natural language understanding. 
For example, a developer could use Llama Stack to build an AI that not only analyzes images for object recognition but also generates detailed descriptions or suggests actions based on the visual data. 
By providing a seamless integration between visual and language processing, Llama Stack is poised to play a crucial role in the future of MLLM-driven applications in fields like e-commerce, robotics, and virtual reality.

\subsubsection{LangChain}

LangChain is a framework that is highly relevant to the MLLM space, as it supports multimodal inputs, enabling the seamless integration of text, audio, image, and other data types into AI models \cite{chase2022langchain}. 
LangChain simplifies the process of working with multimodal data by offering developers functionality to pass different modalities directly to models, making it a valuable tool for MLLM development.

LangChain's ability to handle multimodal data makes it ideal for building applications where diverse input types must be processed simultaneously, such as intelligent assistants capable of understanding both spoken commands and visual cues. 
For example, a developer could use LangChain to build a chatbot that processes both images and text, allowing users to ask questions about a picture and receive detailed explanations in return. 
LangChain's versatility in handling various input types expands the possibilities for MLLM applications in fields like customer service, creative design, and interactive storytelling.

\subsection{The Evolution of Multimodal Large Language Models in Real-World Applications}
The case studies presented in this chapter provide a comprehensive view of how Multimodal Large Language Models (MLLMs) are transforming industries by integrating diverse data types, improving human-AI interaction, and revolutionizing content creation, search, code generation, and robotics. These innovations, driven by cutting-edge neural network architectures and advanced training methods, are making AI systems more dynamic, responsive, and capable of understanding and generating outputs across various modalities like text, audio, images, and video.

From creative industries leveraging tools like Midjourney \cite{Midjourney} and DALL-E 3 \cite{DALLE3} to generate high-quality visuals, to enterprises using GitHub Copilot \cite{GitHubCopilot} and Amazon CodeWhisperer \cite{AmazonCodeWhisperer} for AI-driven code generation, MLLMs have proven to be invaluable across a range of real-world applications. The advancements in video generation tools such as Runway \cite{ha2022runway} has lowered the barrier to entry for video production, while audio and speech processing platforms are reshaping accessibility and personalized content.

The integration of MLLMs with DevOps and infrastructure platforms, such as Hugging Face, LangChain \cite{chase2022langchain}, and Ollama, has further accelerated innovation, providing developers with robust tools to experiment, develop, and scale MLLM-based applications. These platforms make it easier to deploy MLLMs across various sectors, enabling more dynamic, real-time interactions and capabilities in areas like customer service, robotics, and creative workflows.

Moreover, the rise of Retrieval-Augmented Generation (RAG) systems—seen through the development of tools like Pinecone \cite{johnson2021pinecone}—represents a significant shift in AI, where models no longer rely solely on pre-trained data but actively retrieve real-time information to generate contextually relevant outputs \cite{lewis2020retrieval}. This blend of retrieval and generation enhances the accuracy and utility of AI in critical applications such as legal research, personalized recommendations, and technical support.

In robotics and embodied AI, advancements showcase how MLLMs are enabling robots to perform complex, real-world tasks by interpreting both visual inputs and natural language commands. These breakthroughs mark a significant leap forward in creating adaptable, autonomous robotic systems capable of interacting seamlessly with their environment, offering new possibilities in automation for industries ranging from manufacturing to healthcare.

In conclusion, MLLMs are not only pushing the boundaries of what AI can achieve but also unlocking new opportunities for businesses, creators, and developers across various fields. The integration of multimodal capabilities into AI systems is ushering in an era where machines can understand, generate, and interact with the world in ways that closely mimic human cognition and creativity. As these technologies continue to evolve, we can expect MLLMs to play an increasingly central role in shaping the future of artificial intelligence, driving innovation, efficiency, and enhanced user experiences across all sectors.

\chapter{Challenges and Limitations of Multimodal Large Language Models}

\section{Introduction}
Multimodal Large Language Models (MLLMs) represent a significant advancement in artificial intelligence, marking a paradigm shift from traditional unimodal approaches to more comprehensive systems capable of processing and generating content across various modalities, including text, images, audio, and video. These sophisticated models have demonstrated remarkable capabilities in tasks such as image captioning, visual question answering, and cross-modal retrieval, pushing the boundaries of what artificial intelligence can achieve. However, the development and deployment of MLLMs are fraught with significant challenges that span technical, architectural, and ethical domains. This chapter provides a comprehensive exploration of these challenges, examining the complex landscape of multimodal AI and its implications for future research and applications.

The emergence of MLLMs represents a natural evolution in the field of artificial intelligence, addressing the fundamental limitation of traditional language models that operate solely within the textual domain. By incorporating multiple modalities, these models better reflect human cognitive processes, which naturally integrate information from various sensory inputs. This advancement has opened new possibilities in human-computer interaction, content understanding, and generation, while simultaneously introducing novel challenges that require innovative solutions.

\section{Model Architecture and Scalability}

\subsection{Designing Efficient Multimodal Architectures}
The design of MLLM architectures presents unique challenges due to the need to process and integrate information from multiple modalities effectively. Unlike traditional language models, which operate within a single modality, MLLMs must handle diverse input types while maintaining coherent internal representations and generating meaningful outputs. This complexity necessitates careful consideration of architectural choices and their implications for model performance, efficiency, and scalability.

\subsubsection{Cross-modal Attention Mechanisms}
Attention mechanisms are crucial for MLLMs to capture relationships between different modalities. These mechanisms serve as the foundation for understanding complex interactions between various types of input, enabling models to focus on relevant information across modalities. However, designing efficient and effective cross-modal attention remains challenging:

\begin{itemize}
    \item \textbf{Computational Complexity}: Traditional attention mechanisms scale quadratically with input size, which becomes problematic for multimodal inputs. This scaling challenge is particularly acute when dealing with high-dimensional inputs such as images or video sequences combined with text. Recent work on efficient attention, such as Performer \cite{choromanski2021rethinking}, offers promise but requires adaptation for multimodal settings. These adaptations must balance computational efficiency with the ability to capture nuanced cross-modal relationships.

    \item \textbf{Modality-specific Biases}: Attention weights may be biased towards certain modalities, leading to suboptimal integration. This bias can result in models that overemphasize one modality while neglecting important information from others. Kim et al. \cite{kim2021vilt} proposed ViLT, which uses a single transformer for both vision and language, but balancing attention across modalities remains an open problem. Addressing these biases requires careful architectural design and training strategies that ensure equal representation and importance across all modalities.
    
    \item \textbf{Long-range Dependencies}: Capturing long-range dependencies across modalities is crucial but computationally expensive. These dependencies are essential for understanding complex relationships between different parts of multimodal inputs, such as connecting visual elements with their textual descriptions or understanding temporal relationships in video content. Techniques like Longformer \cite{beltagy2020longformer} could be adapted for multimodal contexts but require careful design to handle cross-modal interactions effectively.
\end{itemize}

\subsubsection{Modality-specific vs. Unified Encoders}
The choice between separate encoders for each modality and a unified encoder for all modalities presents significant trade-offs that must be carefully considered in the design of MLLMs:

\begin{itemize}
    \item \textbf{Separate Encoders}: Models like CLIP \cite{radford2021learning} use separate encoders for images and text, allowing for modality-specific pre-training. This approach enables specialized processing of each modality and can leverage existing pre-trained models. However, this approach may struggle with fine-grained cross-modal reasoning due to the potential semantic gap between different encoder spaces. The challenge lies in effectively bridging these separate representations while maintaining the benefits of specialized processing.
    
    \item \textbf{Unified Encoders}: Models like DALL-E \cite{ramesh2021zero} process both text and images in a single transformer, potentially allowing for better cross-modal integration. This unified approach can facilitate more natural interactions between modalities and potentially lead to emergent cross-modal understanding. However, this approach may sacrifice modality-specific optimizations that could benefit individual tasks. The key challenge is designing architectures that can effectively handle the diverse characteristics of different modalities within a single framework.
    
    \item \textbf{Hybrid Approaches}: Recent work by \cite{lu2022unified} on Unified-IO proposes a hybrid approach, using modality-specific tokenizers followed by a shared transformer. This promises a balance between specialization and integration but introduces additional complexity. These approaches attempt to combine the benefits of both separate and unified encoders while minimizing their respective drawbacks. The challenge lies in managing the increased architectural complexity while maintaining efficient training and inference.

\end{itemize}

\subsubsection{Scaling Laws for Multimodal Models}
Understanding how MLLM performance scales with model size and dataset characteristics is crucial for efficient development. This understanding guides resource allocation and architectural decisions in the development of increasingly capable models:

\begin{itemize}
    \item \textbf{Modality-specific Scaling}: Work by \cite{kaplan2020scaling} on language model scaling laws needs extension to multimodal settings. Different modalities may exhibit distinct scaling characteristics due to their inherent properties and computational requirements. Preliminary studies suggest that different modalities may have different optimal scaling relationships, necessitating careful consideration of how to allocate model capacity across modalities.
    
    \item \textbf{Cross-modal Scaling}: The relationship between model size and cross-modal performance is not well understood. This relationship is particularly complex due to the interactions between different modalities and the potential for emergent capabilities as models scale. Recent work by on scaling vision-language models provides initial insights, but more comprehensive studies are needed to fully understand the scaling dynamics of multimodal systems.
    
    \item \textbf{Dataset Scaling}: The impact of dataset size and quality on MLLM performance across different tasks and modalities requires further investigation. The quality and diversity of training data play crucial roles in model performance, but the relationships between dataset characteristics and model capabilities are not yet fully understood. Work by highlights the need for diverse and high-quality multimodal datasets, emphasizing the importance of careful data curation and scaling strategies.
\end{itemize}

\subsection{Computational Efficiency and Latency}
The computational demands of MLLMs present significant challenges for both training and inference. As these models grow in complexity and capability, managing their computational requirements becomes increasingly critical for practical applications. The intersection of multiple modalities introduces unique computational challenges that exceed those of traditional unimodal models, requiring innovative solutions across various aspects of model design and deployment.

\subsubsection{Inference Optimization}
Real-time applications of MLLMs require low-latency inference, which is challenging due to the models' size and complexity. The ability to process multiple modalities simultaneously while maintaining acceptable response times is crucial for practical applications, from real-time video analysis to interactive conversational agents. This challenge becomes particularly acute when dealing with resource-constrained environments or when scaling to serve multiple users:

\begin{itemize}
    \item \textbf{Model Compression}: Techniques like quantization and pruning \cite{ganesh2021compressing} need adaptation for multimodal settings. These compression methods must be carefully calibrated to preserve both modality-specific features and cross-modal relationships. Balancing compression across modalities while maintaining cross-modal performance is particularly challenging, as different modalities may have varying sensitivities to compression. For example, visual features might be more robust to quantization than text embeddings, requiring modality-specific compression strategies.
    
    \item \textbf{Hardware-aware Design}: Designing MLLMs with specific hardware accelerators in mind can significantly improve efficiency. This approach requires deep understanding of both the computational patterns of multimodal processing and the capabilities of modern hardware architectures. Work on hardware-aware transformers by could be extended to multimodal architectures, considering the unique processing requirements of different modalities. This includes optimizing memory access patterns, reducing communication overhead between processing units, and leveraging specialized hardware features for different types of operations.
    
    \item \textbf{Adaptive Computation}: Techniques like conditional computation \cite{bengio2013estimating} could allow MLLMs to allocate computational resources dynamically based on input complexity across modalities. This approach enables more efficient processing by adjusting the computational depth and width of the model based on the complexity of inputs in each modality. For instance, a simple image might require less processing than a complex scene, or a short text prompt might need fewer layers than a lengthy document. Implementing such adaptive mechanisms while maintaining model coherence and performance presents both opportunities and challenges.
\end{itemize}

\subsubsection{Training Efficiency}
The computational demands of training MLLMs are substantial and require innovative approaches. The challenge of training these models efficiently is compounded by the need to process and integrate information from multiple modalities simultaneously, often with varying data distributions and learning dynamics:

\begin{itemize}
    \item \textbf{Efficient Optimization}: Techniques like large batch training and gradient accumulation need careful adaptation for multimodal data to ensure stable and efficient training. The heterogeneous nature of multimodal data introduces additional complexity to optimization procedures, requiring careful consideration of learning rates, batch sizes, and gradient scaling across different modalities. Maintaining stable training dynamics while maximizing hardware utilization becomes particularly challenging when dealing with inputs of varying sizes and computational requirements across modalities.
    
    \item \textbf{Curriculum Learning}: Designing effective curricula for multimodal learning is challenging but could significantly speed up training. Work by on multi-modal curriculum learning provides initial directions, suggesting ways to structure the learning process from simple to complex examples across modalities. The challenge lies in determining appropriate difficulty metrics for different modalities and designing curricula that promote effective cross-modal learning while maintaining training efficiency. This includes considering how to balance modality-specific progression with cross-modal integration tasks.
    
    \item \textbf{Pre-training Strategies}: Developing efficient pre-training objectives that capture cross-modal relationships without requiring excessive computation is an ongoing challenge. Recent work on contrastive learning in multimodal settings \cite{jia2021scaling} shows promise but requires further investigation. The design of pre-training objectives must balance the need for learning robust modality-specific representations with the development of meaningful cross-modal associations. This includes considering how to effectively leverage large-scale multimodal datasets while managing computational constraints.
\end{itemize}

\subsubsection{Memory Management}
Handling multiple modalities simultaneously poses significant memory challenges. The diverse nature of multimodal data, combined with the need to maintain multiple types of representations and their interactions, creates unique memory management requirements:

\begin{itemize}
    \item \textbf{Gradient Checkpointing}: Techniques like gradient checkpointing \cite{chen2016training} need adaptation for multimodal architectures to balance memory usage and computational overhead. The challenge lies in determining optimal checkpointing strategies that consider the varying computational costs and memory requirements of different modalities. This includes deciding which intermediate activations to store versus recompute, taking into account the specific characteristics of different types of layers and cross-modal interactions.
    
    \item \textbf{Attention Memory Optimization}: Developing memory-efficient attention mechanisms is crucial for scalable multimodal processing. Recent work on linear attention \cite{katharopoulos2020transformers} and kernel-based methods \cite{choromanski2021rethinking} could be extended to multimodal settings. These approaches must be adapted to handle the unique challenges of cross-modal attention, where attention patterns may vary significantly between different types of inputs. The optimization must consider both intra-modal and cross-modal attention mechanisms, potentially employing different strategies for different types of interactions.
    
    \item \textbf{Mixed Precision Training}: Utilizing mixed precision training \cite{micikevicius2018mixed} in multimodal contexts requires careful consideration of numerical stability across different modalities and operations. Different modalities may have varying sensitivities to numerical precision, necessitating modality-specific precision strategies. This includes determining appropriate precision levels for different types of operations and managing the transition between precision levels while maintaining model stability and performance.
\end{itemize}

\section{Cross-modal Learning and Representation}

\subsection{Alignment of Different Modalities}
Creating unified representations that effectively capture information across modalities is a central challenge in MLLM development. The goal is to develop representations that can meaningfully capture and integrate information from different modalities while preserving the unique characteristics and relationships within each modality. This challenge is fundamental to enabling sophisticated cross-modal reasoning and generation capabilities.

\subsubsection{Joint Embedding Spaces}
Developing techniques for learning aligned embeddings across modalities is crucial for effective multimodal reasoning. The creation of these joint embedding spaces must balance the need for modality-specific information preservation with the ability to perform meaningful cross-modal operations:

\begin{itemize}
    \item \textbf{Contrastive Learning}: Methods like CLIP \cite{radford2021learning} use contrastive learning to align visual and textual representations through a self-supervised learning approach. While these methods have shown remarkable success in image-text alignment, extending these approaches to more modalities and fine-grained alignments remains challenging. The challenge includes designing appropriate contrastive objectives that can handle multiple modalities simultaneously and capture fine-grained semantic relationships across modalities.
    
    \item \textbf{Cross-modal Autoencoders}: Techniques like multimodal autoencoders \cite{ngiam2011multimodal} aim to learn shared representations through reconstruction objectives. These approaches attempt to find common latent spaces that can capture the essential information from each modality while enabling cross-modal generation and translation. Balancing modality-specific and shared information in these models is an ongoing research direction, requiring careful consideration of architecture design and training objectives.
    
    \item \textbf{Optimal Transport}: Recent work by chen2020optimal uses optimal transport theory to align cross-modal embeddings, providing a principled framework for learning alignments between different modality spaces. Scaling these approaches to large-scale MLLMs and multiple modalities presents both opportunities and challenges. The mathematical foundations of optimal transport offer promising directions for achieving more precise and theoretically grounded cross-modal alignments, but computational scalability and adaptation to multiple modalities remain significant challenges.

\end{itemize}

\subsubsection{Temporal Alignment in Video-Text Models}
Handling temporal aspects in multimodal data, particularly for video understanding, presents unique challenges that go beyond the complexity of static multimodal content. The dynamic nature of video data, combined with the need to align and understand relationships across different modalities over time, introduces significant computational and modeling challenges. This temporal dimension adds layers of complexity to the already challenging task of multimodal integration.

\begin{itemize}
    \item \textbf{Long-term Dependencies}: Capturing long-term dependencies across modalities in video understanding tasks is computationally challenging and requires sophisticated architectural solutions. These dependencies can span seconds, minutes, or even longer periods, making it difficult to maintain relevant context over time. Approaches like hierarchical transformers \cite{liu2021video} show promise but require further development for multimodal settings. These architectures attempt to build representations at multiple temporal scales, from fine-grained frame-level features to high-level semantic concepts that span longer durations. The challenge lies in effectively combining these hierarchical representations while maintaining computational efficiency and meaningful cross-modal relationships.
    
    \item \textbf{Asynchronous Events}: Dealing with asynchronous events across modalities (e.g., delayed narration in videos) requires sophisticated temporal modeling that can handle complex temporal relationships. This challenge is particularly evident in real-world scenarios where different modalities may not be perfectly synchronized or may have varying temporal granularity. Recent work on temporal attention mechanisms provides a starting point for addressing this challenge, offering ways to learn flexible temporal alignments between modalities. These mechanisms must be capable of handling varying temporal scales and maintaining coherent cross-modal understanding despite temporal misalignments.
    
    \item \textbf{Efficient Video Processing}: Processing high-resolution video data in MLLMs is computationally intensive, requiring careful consideration of resource utilization and efficiency. The challenge is compounded by the need to process multiple frames while maintaining temporal coherence and cross-modal relationships. Techniques like dynamic sparse attention \cite{child2019generating} could be adapted for efficient video processing in multimodal contexts, allowing models to focus computational resources on the most relevant temporal and spatial regions. This includes developing methods for adaptive frame sampling, temporal pooling, and efficient feature extraction that preserve important temporal dynamics while reducing computational overhead.
\end{itemize}

\subsection{Transfer Learning and Generalization}
Enabling effective knowledge transfer between modalities and tasks is crucial for the development of versatile MLLMs. The ability to leverage knowledge across different modalities and adapt to new tasks efficiently represents a fundamental challenge in multimodal learning. This capability is essential for creating models that can generalize effectively and adapt to new situations with minimal additional training.

\subsubsection{Cross-modal Transfer}
Facilitating knowledge transfer between modalities presents several challenges that must be addressed to create truly adaptive and generalizable multimodal systems:

\begin{itemize}
    \item \textbf{Zero-shot Cross-modal Transfer}: Enabling MLLMs to perform tasks in one modality based on knowledge from another without specific training examples is a significant challenge that requires sophisticated architectural and training approaches. Work by on frozen language models for visual learning provides insights into potential approaches, but generalizing this approach to multiple modalities and tasks remains an open problem. The challenge lies in creating representations that can effectively bridge different modalities while maintaining the specific characteristics and requirements of each modality. This includes developing methods for abstract reasoning that can translate concepts learned in one modality to meaningful applications in another.
    
    \item \textbf{Few-shot Learning}: Developing few-shot learning techniques that effectively leverage knowledge across modalities is crucial for adaptive MLLMs that can quickly learn from limited examples. Recent work on meta-learning in multimodal contexts \cite{pahde2021multimodal} shows promise but requires further investigation for large-scale models. The challenge involves creating learning algorithms that can effectively utilize prior knowledge across modalities to accelerate learning in new situations. This includes developing methods for efficient adaptation that can leverage cross-modal relationships while maintaining model stability and performance.
    
    \item \textbf{Negative Transfer}: Preventing negative transfer, where learning in one modality degrades performance in another, is a significant challenge that requires careful consideration of learning dynamics and knowledge representation. Techniques like gradient surgery could be adapted for multimodal settings to mitigate negative transfer by identifying and preventing harmful parameter updates. This includes developing methods to detect and prevent interference between modalities while maintaining beneficial knowledge transfer.
\end{itemize}

\subsubsection{Domain Adaptation in Multimodal Settings}
MLLMs often struggle when faced with domain shifts, particularly when these shifts occur differently across modalities. The challenge of domain adaptation becomes more complex in multimodal settings due to the need to handle shifts in multiple modalities simultaneously while maintaining cross-modal relationships:

\begin{itemize}
    \item \textbf{Cross-modal Domain Adaptation}: Developing techniques that can adapt to domain shifts in multiple modalities simultaneously is challenging due to the complex interactions between modalities and the need to maintain coherent cross-modal relationships. Recent work on multi-source domain adaptation \cite{peng2019moment} provides a foundation for addressing these challenges, but extending these approaches to large-scale MLLMs remains an open problem. This includes developing methods that can effectively handle varying degrees of domain shift across different modalities while maintaining model performance and cross-modal understanding.
    
    \item \textbf{Unsupervised Multimodal Adaptation}: Creating unsupervised domain adaptation techniques for multimodal data is crucial for real-world deployments where labeled data in target domains may be scarce or unavailable. Approaches like MUDA show promise in addressing this challenge but require scaling to more complex multimodal scenarios. The challenge involves developing methods that can effectively leverage unlabeled data across modalities to adapt to new domains while maintaining model performance and reliability.
    
    \item \textbf{Continual Adaptation}: Enabling MLLMs to continuously adapt to changing domains across modalities without forgetting previously learned knowledge is a significant challenge that requires sophisticated approaches to memory and learning. Techniques like elastic weight consolidation \cite{kirkpatrick2017overcoming} need careful adaptation for multimodal continual learning scenarios. This includes developing methods for selective parameter updates that can preserve important knowledge while allowing for adaptation to new domains and tasks.
\end{itemize}

\section{Model Robustness and Reliability}

\subsection{Adversarial Robustness}
MLLMs are vulnerable to adversarial attacks, particularly those that exploit the interaction between modalities. The multimodal nature of these models introduces new attack surfaces and vulnerabilities that must be carefully addressed to ensure reliable and secure deployment.

\subsubsection{Cross-modal Adversarial Attacks}
Developing robust MLLMs requires addressing various types of adversarial attacks that can exploit vulnerabilities in cross-modal processing and integration:

\begin{itemize}
    \item \textbf{Multimodal Adversarial Examples}: Creating defense mechanisms against adversarial examples that span multiple modalities is challenging due to the complex interactions between different types of inputs and the potential for attacks to exploit cross-modal dependencies. Recent work by on attacking audio-visual models highlights the complexity of cross-modal adversarial attacks and the need for sophisticated defense mechanisms. This includes developing methods that can detect and mitigate attacks that target multiple modalities simultaneously or exploit inconsistencies in cross-modal processing.
    
    \item \textbf{Certified Robustness}: Extending certified robustness techniques to multimodal settings is an open problem that requires new theoretical frameworks and practical implementations. Approaches like randomized smoothing cohen2019certified need adaptation to handle the complexities of multiple input modalities and their interactions. The challenge involves developing certification methods that can provide meaningful guarantees about model behavior across different modalities and types of inputs.
    
    \item \textbf{Transferability of Attacks}: Understanding and mitigating the transferability of adversarial examples across modalities and model architectures is crucial for developing robust MLLMs. Work by naseer2019cross on cross-modal transferability provides initial insights into how attacks can transfer between modalities, but more comprehensive studies are needed for MLLMs. This includes investigating how adversarial perturbations in one modality can affect processing in other modalities and developing defense mechanisms that can handle these complex attack scenarios.

\end{itemize}

\subsubsection{Robustness to Input Perturbations}
Ensuring consistent performance under various input conditions is crucial for reliable MLLM deployment in real-world applications. The challenge of maintaining robust performance becomes particularly complex in multimodal settings, where perturbations can affect different modalities independently or in combination. Understanding and addressing these challenges is essential for developing MLLMs that can operate reliably in diverse and unpredictable environments.

\begin{itemize}
    \item \textbf{Visual Robustness}: Developing models robust to visual noise, occlusions, and transformations is challenging and requires sophisticated approaches to maintain performance across a wide range of visual conditions. This challenge is particularly acute in real-world scenarios where lighting conditions, camera angles, and image quality can vary significantly. Techniques like adversarial training \cite{madry2018towards} need adaptation for multimodal contexts to improve visual robustness without compromising performance on clean data. This includes developing methods that can maintain cross-modal understanding even when visual inputs are degraded or partially obscured, while also ensuring that defensive mechanisms don't interfere with the model's ability to extract meaningful features from clean inputs.
    
    \item \textbf{Linguistic Variations}: Addressing robustness to linguistic variations, including typos, dialects, and non-standard language use, is crucial for creating MLLMs that can effectively serve diverse user populations. This challenge becomes particularly important in multilingual and multicultural contexts where language use can vary significantly from standard forms. Recent work on text perturbation strategies \cite{tan2020mind} could be extended to multimodal settings, providing ways to systematically evaluate and improve robustness to linguistic variations while maintaining cross-modal understanding. This includes developing methods that can handle variations in text while preserving semantic relationships with other modalities.
    
    \item \textbf{Cross-modal Consistency}: Ensuring consistent outputs when information across modalities is perturbed or conflicting presents unique challenges that require careful consideration of how different modalities interact and influence each other. The challenge involves developing methods that can maintain coherent outputs even when different modalities provide contradictory or noisy information. Developing evaluation metrics and training objectives for cross-modal consistency is an active area of research, requiring new approaches to quantifying and optimizing the alignment between different modalities under various perturbation scenarios.
\end{itemize}

\subsection{Handling Missing or Noisy Modalities}
Real-world applications of MLLMs often involve scenarios where some modalities are missing or corrupted, making robust handling of incomplete or degraded inputs essential for practical deployment. This challenge is particularly relevant in applications where sensor failures, network issues, or other technical limitations may result in missing or degraded modalities.

\subsubsection{Graceful Degradation}
Developing MLLMs that maintain reasonable performance with partial or noisy inputs is crucial for ensuring reliable operation in real-world conditions. The ability to gracefully handle degraded inputs while maintaining as much functionality as possible represents a key challenge in multimodal system design:

\begin{itemize}  
    \item \textbf{Modality Imputation}: Developing methods to infer or reconstruct missing modalities could improve robustness by providing substitute inputs when original modalities are unavailable. Recent work on cross-modal generation \cite{ramesh2021zero} provides a foundation for addressing this challenge, but adapting these techniques for real-time inference in MLLMs is an open problem. This includes developing efficient methods for generating high-quality imputations that maintain semantic consistency with available modalities while being computationally feasible for real-time applications.
\end{itemize}

\subsubsection{Uncertainty Quantification}
Reliable deployment of MLLMs requires accurate uncertainty estimation, particularly in multimodal contexts where different sources of uncertainty can interact in complex ways. Understanding and quantifying uncertainty is crucial for making informed decisions about model outputs and identifying situations where additional information or human intervention may be needed:

\begin{itemize}
    \item \textbf{Calibration Techniques}: Developing calibration methods for multimodal outputs is challenging due to the diverse nature of different modalities and the need to maintain consistent calibration across various types of outputs. Recent work on temperature scaling \cite{guo2017calibration} needs extension to handle multi-modal outputs effectively. This includes developing methods that can provide well-calibrated uncertainty estimates across different modalities while accounting for their unique characteristics and potential interactions.
    
    \item \textbf{Bayesian MLLMs}: Exploring Bayesian approaches to uncertainty quantification in MLLMs is a promising direction for providing principled uncertainty estimates in multimodal settings. Techniques like variational inference \cite{blei2017variational} need adaptation to handle the complexities of multimodal architectures, including the development of appropriate prior distributions and efficient inference methods that can scale to large multimodal models. This includes addressing challenges related to computational efficiency and the handling of different types of uncertainty across modalities.
    
    \item \textbf{Out-of-distribution Detection}: Identifying out-of-distribution inputs in multimodal settings is crucial for safe deployment, as it enables models to recognize situations where their predictions may be unreliable. Recent work on contrastive training for OOD detection \cite{tack2020csi} could be extended to multimodal scenarios, providing ways to identify unusual or potentially problematic inputs across different modalities. This includes developing methods that can effectively detect out-of-distribution samples while considering the joint distribution of multiple modalities.
\end{itemize}

\section{Interpretability and Explainability (Continued)}

\subsection{Visualizing Cross-modal Attention}
Understanding how MLLMs attend to and integrate information from different modalities is crucial for interpretability and trust in these systems. The visualization of attention mechanisms in multimodal contexts presents unique challenges due to the complex interactions between different modalities and the need to represent these interactions in an interpretable manner. This understanding is essential not only for model development and debugging but also for building user trust and enabling effective human oversight.

\subsubsection{Attention Map Analysis}
Analyzing attention patterns in MLLMs presents unique challenges in multimodal contexts, requiring sophisticated approaches to visualize and interpret how models integrate information across different modalities:

\begin{itemize}
    \item \textbf{Multi-head Attention Visualization}: Techniques like attention rollout \cite{abnar2020quantifying} need adaptation for multimodal scenarios to capture complex cross-modal interactions effectively. The challenge lies in developing visualization methods that can meaningfully represent attention patterns across different modalities while maintaining interpretability. This includes addressing questions of how to visualize attention between different types of tokens (e.g., text tokens, image patches, audio segments) and how to represent the hierarchical nature of attention in deep networks. The visualization must be both technically accurate and intuitively understandable to humans, potentially requiring different levels of abstraction for different audiences.
    
    \item \textbf{Temporal Attention Analysis}: For video-based MLLMs, visualizing attention over time presents additional challenges that require consideration of both spatial and temporal dimensions. Work by on temporal attention could be extended to multi-modal temporal data, providing insights into how models integrate information across time and modalities. This includes developing methods to visualize how attention patterns evolve over time, how different modalities influence each other temporally, and how the model maintains coherence across longer sequences. The challenge involves creating visualizations that can effectively represent these complex temporal relationships while remaining comprehensible to human observers.
    
    \item \textbf{Cross-modal Attention Flows}: Developing methods to visualize how information flows between modalities through attention mechanisms is an open challenge that requires innovative approaches to representation and visualization. Techniques like attention flow \cite{abnar2020quantifying} could be adapted for cross-modal settings, providing insights into how information is integrated across modalities. This includes developing methods to track and visualize how information from one modality influences the processing of others, how different modalities contribute to final predictions, and how attention patterns reflect the model's understanding of relationships between modalities.
\end{itemize}

\subsubsection{Feature Attribution Methods}
Extending feature attribution techniques to multimodal settings presents unique challenges that require careful consideration of how to attribute importance across different types of inputs while maintaining consistency and interpretability:

\begin{itemize}
    \item \textbf{Gradient-based Methods}: Techniques like Integrated Gradients \cite{sundararajan2017axiomatic} need careful adaptation to handle multiple input modalities consistently while providing meaningful attributions. The challenge involves developing methods that can appropriately scale and compare gradients across different types of inputs, accounting for the different characteristics and scales of each modality. This includes addressing questions of how to normalize attributions across modalities, how to handle interactions between modalities, and how to present these attributions in a way that is meaningful to human observers. The development of these methods must consider both the technical accuracy of the attributions and their practical utility for understanding model behavior.
    
    \item \textbf{Perturbation-based Methods}: Methods like LIME \cite{ribeiro2016should} require extension to generate meaningful perturbations across different modalities while maintaining semantic coherence. The challenge lies in developing perturbation strategies that are appropriate for each modality while considering cross-modal dependencies and constraints. This includes determining how to generate realistic perturbations that preserve semantic relationships between modalities, how to sample perturbations effectively in high-dimensional multimodal spaces, and how to aggregate results across different types of perturbations. The development of these methods must balance the need for comprehensive exploration of the input space with computational feasibility and interpretability of results.
    
    \item \textbf{Unified Attribution Frameworks}: Developing frameworks that provide consistent attributions across modalities is crucial for understanding how different inputs contribute to model decisions. Recent work on unified saliency maps \cite{rebuffi2020saliency} provides a starting point but requires further development for complex MLLMs. This includes creating methods that can meaningfully compare and combine attributions across different modalities, handling challenges related to different scales and characteristics of different input types, and developing presentation methods that can effectively communicate these unified attributions to users. The framework must be both theoretically sound and practically useful for understanding model behavior in real-world applications.
\end{itemize}

\section{Challenges and Future Directions in Multimodal Large Language Models}

The rapid advancement of artificial intelligence has ushered in a new era of language models, with Multimodal Large Language Models (MLLMs) emerging as a frontier technology that promises to revolutionize how machines understand and interact with the world. These sophisticated systems, capable of processing and generating content across various modalities such as text, images, audio, and video, represent a significant leap forward in AI capabilities. However, with great power comes great responsibility, and the development of MLLMs presents a complex landscape of challenges that researchers and practitioners must navigate.

\subsection{Concept-based Explanations}
As we push the boundaries of what MLLMs can achieve, it becomes increasingly crucial to move beyond low-level feature attribution and towards higher-level concept-based explanations. This shift is not merely an academic exercise but a fundamental requirement for making MLLMs more interpretable, trustworthy, and ultimately more useful in real-world applications.

\subsubsection{Multimodal Concept Discovery}
The identification of interpretable concepts across modalities stands as a cornerstone challenge in enhancing the explainability of MLLMs. This area of research holds immense potential for bridging the gap between machine reasoning and human understanding.

\begin{itemize}
    \item \textbf{Unsupervised Concept Discovery}: While techniques like TCAV \cite{kim2018interpretability} have shown promise in single-modality scenarios, there is a pressing need to extend these approaches to discover concepts that span multiple modalities. This extension is non-trivial, as it requires algorithms capable of identifying abstract concepts that manifest differently across diverse data types. For instance, the concept of "joy" might be expressed through positive words in text, upbeat melodies in audio, and smiling faces in images. Developing methods that can autonomously discover such cross-modal concepts would significantly enhance our ability to interpret MLLM decision-making processes.
    
    \item \textbf{Cross-modal Concept Alignment}: The challenge of aligning concepts across modalities represents a fundamental hurdle in multimodal understanding. This task involves developing sophisticated algorithms that can recognize when different modalities are expressing the same underlying concept, even when the surface-level representations are vastly different. For example, aligning the visual concept of a "cozy home" with its textual descriptions and associated sounds requires a deep understanding of both the individual modalities and their interrelationships. Solving this challenge could lead to MLLMs that can more seamlessly translate concepts between modalities, greatly enhancing their versatility and applicability in diverse scenarios.
    
    \item \textbf{Hierarchical Concept Learning}: The development of frameworks for learning hierarchical concept structures that integrate information across modalities represents a frontier in MLLM research. Such frameworks would need to capture not only the relationships between concepts within a single modality but also how these hierarchies interact and align across different modalities. This approach could lead to more nuanced and context-aware interpretations of multimodal data, allowing MLLMs to reason about complex scenarios with a level of abstraction closer to human cognition. For instance, a hierarchical concept structure might relate high-level concepts like "transportation" to more specific concepts like "cars" and "bicycles" across visual, textual, and auditory modalities, enabling richer and more coherent multimodal reasoning.
\end{itemize}

\subsubsection{Compositional Explanations}
The complexity of multimodal reasoning demands explanatory approaches that are inherently compositional, capable of breaking down complex decisions into understandable components while preserving the richness of cross-modal interactions.

\begin{itemize}
    \item \textbf{Neuro-symbolic Methods}: The integration of symbolic reasoning with neural networks, as exemplified by the work of \cite{mao2019neuro}, holds significant promise for providing more interpretable explanations in multimodal contexts. These hybrid approaches aim to combine the flexibility and learning capabilities of neural networks with the transparency and logical rigor of symbolic systems. In the context of MLLMs, neuro-symbolic methods could enable the generation of explanations that are both data-driven and logically structured, potentially offering insights into how the model combines information across modalities to arrive at its conclusions. For example, a neuro-symbolic MLLM might explain its classification of a scene as "dangerous" by providing a logical chain of reasoning that incorporates visual cues (e.g., presence of smoke), textual context (e.g., news reports of a fire), and audio information (e.g., sound of sirens).

    \item \textbf{Program Synthesis}: Techniques for synthesizing programs that explain model decisions, such as those explored by \cite{ellis2018learning}, represent a powerful approach to generating interpretable explanations. Extending these methods to multimodal reasoning tasks presents both challenges and opportunities. The goal would be to generate executable programs that can recreate the MLLM's decision-making process in a human-readable format. This approach could be particularly powerful for explaining complex multimodal interactions, as it would allow for step-by-step tracing of how information from different modalities is combined and processed. For instance, a synthesized program might explain how an MLLM determines the mood of a movie scene by detailing the steps it takes to analyze the visual composition, dialogue sentiment, and musical score, and how it weighs and combines these factors.

    \item \textbf{Natural Language Explanations}: The generation of coherent natural language explanations that integrate information from multiple modalities remains one of the most significant challenges in making MLLMs interpretable. This task requires not only the ability to reason across modalities but also the capacity to translate that reasoning into clear, concise, and contextually appropriate language. The difficulty lies in capturing the nuances of multimodal interactions without oversimplifying or losing critical information. Advances in this area could lead to MLLMs that can provide human-like explanations for their decisions, greatly enhancing their usefulness in fields such as education, healthcare, and decision support systems. For example, an advanced MLLM might explain its diagnosis of a medical condition by referencing specific visual features from medical imaging, relevant passages from the patient's medical history, and audio cues from recorded patient interviews, all synthesized into a coherent narrative that a healthcare professional can easily understand and verify.
\end{itemize}

\section{Evaluation and Benchmarking}

As the capabilities of MLLMs continue to expand, the development of robust evaluation methodologies and comprehensive benchmarks becomes increasingly critical. These tools are essential not only for measuring progress in the field but also for identifying limitations, biases, and areas for improvement in MLLM systems.

\subsection{Comprehensive Multimodal Benchmarks}
The creation of benchmarks that effectively evaluate the capabilities and limitations of MLLMs is a cornerstone challenge in advancing the field. These benchmarks must be carefully designed to capture the full spectrum of MLLM abilities while also probing for potential weaknesses and biases.

\subsubsection{Task Diversity}
Ensuring that benchmarks cover a wide range of multimodal tasks is essential for comprehensively evaluating MLLM performance:

\begin{itemize}
    \item \textbf{Cross-modal Reasoning}: The development of tasks that require complex reasoning across modalities represents a frontier in MLLM evaluation. While datasets like CLEVR \cite{johnson2017clevr} have set a high standard for visual reasoning tasks, extending this approach to truly multimodal scenarios presents significant challenges. Such tasks might involve, for example, answering questions about a scene that require integrating information from visual, textual, and auditory inputs. For instance, a cross-modal reasoning task might present a video clip of a busy street scene along with a textual description and ambient audio, then ask questions that require the MLLM to synthesize information across all three modalities to infer complex relationships or predict outcomes.

    \item \textbf{Open-ended Generation}: Creating evaluation protocols for open-ended multimodal generation tasks presents unique challenges, particularly in assessing creativity and coherence across modalities. These tasks might include generating a story with accompanying illustrations, creating a multimedia presentation on a given topic, or composing music with lyrics that match a provided image. The difficulty lies not only in generating content that is coherent within each modality but also in ensuring that the generated elements are semantically aligned and enhance each other across modalities. Evaluation metrics for such tasks must go beyond traditional measures of quality for individual modalities and consider the holistic impact and coherence of the multimodal output.

    \item \textbf{Long-form Understanding}: Designing benchmarks for long-form multimodal content understanding, such as video story comprehension or multimedia document analysis, addresses a critical gap in current evaluation frameworks. These tasks require MLLMs to maintain context and track complex narratives or arguments across extended multimodal inputs. For example, a benchmark might involve summarizing a lengthy documentary film, requiring the model to integrate visual cues, spoken dialogue, background music, and on-screen text over an extended period. Such tasks test not only the model's ability to process individual modalities but also its capacity to synthesize information over time and across modalities to form coherent, high-level understandings.
\end{itemize}

\subsubsection{Fairness and Representation}
Ensuring that benchmark datasets are inclusive and unbiased is an ongoing challenge that requires continuous attention and innovation:

\begin{itemize}
    \item \textbf{Cultural Diversity}: Developing strategies for creating culturally diverse multimodal datasets that represent a wide range of global perspectives is crucial for ensuring that MLLMs can perform effectively across different cultural contexts. This challenge involves not only collecting data from diverse sources but also ensuring that the tasks and evaluation criteria are culturally sensitive and relevant. For instance, a truly diverse benchmark might include tasks that require understanding cultural nuances in gestures, idioms, or social cues across different societies, testing the MLLM's ability to navigate complex cultural landscapes.

    \item \textbf{Intersectionality}: Designing benchmarks that assess model performance across intersectional categories, considering multiple demographic factors simultaneously, is essential for understanding how MLLMs perform for diverse user groups. This approach recognizes that individuals' experiences and identities are shaped by the intersection of various factors such as race, gender, age, and socioeconomic status. Benchmarks incorporating intersectionality might, for example, evaluate an MLLM's ability to understand and generate content relevant to older women from minority ethnic backgrounds, ensuring that the model's performance is robust across diverse intersectional identities.

    \item \textbf{Bias Detection}: Creating tools and metrics for identifying and quantifying biases in multimodal datasets and model outputs is a critical component of responsible MLLM development. This challenge involves developing sophisticated analytical techniques that can detect subtle biases across different modalities and their interactions. For example, a bias detection tool might analyze whether an MLLM consistently associates certain visual characteristics with particular personality traits in generated text descriptions, or whether it shows preferences for certain types of voices when generating audio content to match text or images.
\end{itemize}

\subsection{Metrics for Multimodal Performance}
The development of appropriate metrics to evaluate MLLM performance is crucial for meaningful progress in the field. These metrics must capture not only the quality of outputs in individual modalities but also the coherence and effectiveness of multimodal integration.

\subsubsection{Cross-modal Coherence Metrics}
Evaluating the consistency and coherence of MLLMs across modalities presents a complex challenge that requires innovative approaches:

\begin{itemize}
    \item \textbf{Semantic Alignment Measures}: Developing metrics that assess the semantic alignment between generated content in different modalities is essential for ensuring that MLLMs produce coherent multimodal outputs. These measures must go beyond surface-level similarity to capture deep semantic relationships. For instance, a semantic alignment metric might evaluate how well the emotional tone of generated text matches the mood conveyed by an accompanying generated image or musical piece. This could involve developing new embedding techniques that can represent semantic content across modalities in a comparable space, allowing for quantitative assessment of alignment.

    \item \textbf{Perceptual Similarity Metrics}: Creating metrics that correlate with human judgments of cross-modal similarity and coherence is crucial for developing MLLMs that produce outputs that are not only technically correct but also intuitively coherent to human users. This challenge involves bridging the gap between computational measures and human perception. Approaches might include developing large-scale human evaluation datasets to train machine learning models that can predict human judgments of multimodal coherence, or creating novel perceptual models that simulate human cross-modal processing.

    \item \textbf{Temporal Coherence Measures}: For video-based tasks, developing metrics that evaluate coherence over time across modalities is particularly challenging. These metrics must capture not only the moment-to-moment alignment of different modalities but also the overall narrative or thematic coherence across an extended temporal sequence. This might involve developing new techniques for analyzing the temporal dynamics of multimodal content, such as methods for tracking the evolution of themes or emotions across visual, auditory, and textual components of a video over time.
\end{itemize}

\subsubsection{Compositional Generalization Metrics}
Assessing the ability of MLLMs to combine concepts across modalities in novel ways is crucial for understanding their potential for creative and flexible multimodal reasoning:

\begin{itemize}
    \item \textbf{Systematic Generalization}: Designing evaluation protocols that test for systematic generalization in multimodal contexts, similar to SCAN \cite{lake2018generalization} but extended to multiple modalities, is essential for ensuring that MLLMs can apply learned concepts and relationships to novel situations. These protocols might involve creating carefully constructed test sets that require the model to apply known concepts in new multimodal combinations. For example, a test might assess whether a model that has learned to associate certain visual textures with tactile descriptions can generate appropriate cross-modal content for entirely new texture-description pairs.

    \item \textbf{Few-shot Composition}: Developing metrics to evaluate few-shot compositional abilities across modalities addresses the important challenge of assessing how well MLLMs can quickly adapt to new multimodal tasks with minimal examples. This is particularly relevant for real-world applications where the ability to quickly learn and apply new multimodal concepts is crucial. Metrics in this area might evaluate how effectively a model can learn to generate appropriate audio given a new combination of visual and textual inputs after seeing only a few examples, testing the model's ability to rapidly compose learned unimodal concepts into novel multimodal outputs.

    \item \textbf{Out-of-distribution Composition}: Creating benchmarks that assess compositional generalization to novel combinations of modalities or concepts is crucial for understanding the robustness and flexibility of MLLMs. These benchmarks would test the model's ability to handle inputs or tasks that fall outside the distribution of its training data, particularly in terms of how different modalities are combined. For instance, a benchmark might evaluate how well a model trained on image-caption pairs can handle tasks involving image-audio-text triads, assessing its ability to compose learned bimodal relationships into novel trimodal outputs.
\end{itemize}

\section{Conclusion}

The development of Multimodal Large Language Models represents a frontier of challenges and opportunities in artificial intelligence that promises to reshape our understanding of machine learning and its applications. The journey towards creating MLLMs that can seamlessly integrate and reason across diverse modalities—text, images, audio, and beyond—is fraught with complex technical, ethical, and philosophical questions that demand interdisciplinary collaboration and innovative thinking.

As we have explored in this thesis, the challenges span a wide range of areas, from the fundamental architecture designs that enable efficient multimodal processing to the intricate task of making these complex systems interpretable and explainable. The need for concept-based explanations that bridge the gap between low-level feature attribution and high-level human understanding is particularly pressing, as it holds the key to making MLLMs not just powerful, but also trustworthy and usable in critical real-world applications.

The development of comprehensive evaluation frameworks and benchmarks stands out as a crucial enabler for progress in the field. As MLLMs grow in complexity and capability, our ability to accurately assess their performance, identify their limitations, and ensure their fairness becomes increasingly important. The challenges of creating truly representative and unbiased datasets, developing metrics that can capture the nuances of multimodal coherence and compositional generalization, and designing tasks that probe the full spectrum of MLLM abilities are formidable but essential to overcome.

Moreover, the ethical implications of developing such powerful AI systems cannot be overstated. As MLLMs become more capable of understanding and generating human-like multimodal content, questions of privacy, consent, and the potential for misuse become increasingly urgent. The research community must remain vigilant and proactive in addressing these concerns, ensuring that the development of MLLMs is guided by strong ethical principles and a commitment to beneficial AI.

The path forward requires a delicate balance between pushing the boundaries of what's technically possible and carefully considering the implications and limitations of these advancements. It calls for collaboration across disciplines—machine learning, computer vision, natural language processing, cognitive science, ethics, and beyond—to tackle these multifaceted challenges. By fostering such interdisciplinary efforts, we can work towards creating MLLMs that are not only more powerful and efficient but also more interpretable, fair, and aligned with human values.

As we continue to make strides in this exciting field, it's crucial to maintain a perspective that balances ambition with responsibility. The potential applications of MLLMs are vast and transformative, spanning areas such as healthcare, education, creative industries, and scientific research. In healthcare, for instance, MLLMs could revolutionize diagnosis and treatment planning by integrating patient data across multiple modalities—medical imaging, textual records, and even real-time physiological data. In education, these models could create personalized learning experiences that adapt to individual students' needs, presenting information in the most effective combination of modalities for each learner.

However, realizing these potentials requires overcoming significant challenges:

\begin{itemize}
    \item \textbf{Scalability and Efficiency}: As MLLMs grow in complexity and capability, ensuring their scalability and efficiency becomes increasingly crucial. Future research must focus on developing architectures and training paradigms that can handle the immense computational demands of multimodal processing while remaining accessible and deployable in real-world scenarios.

    \item \textbf{Continual Learning and Adaptation}: Developing MLLMs that can continuously learn and adapt to new information and modalities without forgetting previously acquired knowledge represents a significant challenge. This capability is crucial for creating systems that remain relevant and effective in dynamic, real-world environments.

    \item \textbf{Ethical AI and Governance}: As MLLMs become more powerful and pervasive, establishing robust ethical guidelines and governance frameworks for their development and deployment becomes imperative. This includes addressing issues of bias, fairness, privacy, and the potential socioeconomic impacts of these technologies.

    \item \textbf{Human-AI Collaboration}: Exploring ways to effectively integrate MLLMs into human workflows and decision-making processes presents both technical and social challenges. Developing intuitive interfaces and interaction paradigms that leverage the strengths of both human intelligence and AI capabilities is a critical area for future research.
\end{itemize}

The future of MLLMs is inextricably linked to our broader understanding of intelligence and cognition. As these models become more sophisticated, they not only serve as powerful tools but also as computational models that can inform our understanding of human multimodal processing and reasoning. This reciprocal relationship between AI development and cognitive science offers exciting possibilities for advancing both fields.

Furthermore, the development of MLLMs has the potential to democratize access to powerful AI capabilities, enabling individuals and organizations across various domains to leverage multimodal AI for innovation and problem-solving. However, this democratization must be accompanied by efforts to ensure equitable access and to mitigate the risk of exacerbating existing digital divides.

In conclusion, the field of Multimodal Large Language Models stands at a thrilling juncture, poised to redefine the boundaries of artificial intelligence and its impact on society. The challenges we face are formidable, spanning technical, ethical, and philosophical domains. Yet, these challenges also present unprecedented opportunities for innovation and discovery. By fostering interdisciplinary collaboration, maintaining a commitment to ethical and responsible development, and continually pushing the boundaries of what's possible, we can work towards creating AI systems that truly understand and interact with the world in all its multimodal complexity.

As we move forward, it is crucial to remember that the goal is not just to create more powerful AI systems, but to develop technologies that enhance human capabilities, foster understanding, and contribute positively to society. The journey ahead is long and complex, but the potential rewards—in terms of scientific advancement, technological innovation, and societal benefit—are immense. It is a journey that will require the collective efforts of researchers, practitioners, policymakers, and society at large, working together to shape a future where Multimodal Large Language Models serve as powerful tools for human empowerment and progress.

\chapter{Ethical Considerations and Responsible AI}

As Multimodal Large Language Models (MLLMs) continue to advance and shape the AI landscape, capable of processing and generating content across various modalities such as text, images, and audio, it is crucial to address the ethical implications and challenges that arise from their development and deployment to ensure responsible AI practices \cite{konidena2024ethical}. 

One of the primary concerns in MLLM is bias mitigation. It refers to systematic errors or unfair preferences in the model's outputs that can reinforce or amplify societal prejudices and stereotypes. These biases can manifest in various forms, including gender, racial, or cultural biases, and they pose ethical challenges in the deployment and use of LLMs across different applications \cite{peng2024securing}. Researchers and developers must implement comprehensive bias mitigation strategies \cite{zhang2023mitigating}. These include ensuring diverse and representative training datasets, conducting regular bias \cite{boix2022machine} audits across different modalities \cite{pymetrics2022audit}, and developing bias-aware fine-tuning techniques \cite{kim2024domain}. Additionally, interdisciplinary collaboration with experts from fields such as ethics, sociology, and psychology can provide valuable insights into identifying and addressing potential biases \cite{aquino2023practical}.

Privacy and data protection present another significant challenge in the realm of MLLMs. As these models process and generate increasingly complex and potentially sensitive information, robust measures must be put in place to protect individual privacy \cite{he2024emerged, friha2024llm}. This includes implementing advanced data anonymization techniques, exploring decentralized training methods like federated learning, and applying differential privacy approaches. Furthermore, clear protocols for obtaining consent and managing data rights must be established to ensure ethical handling of personal information used in training these models \cite{mccoy2023ethical}.

The potential for misuse of MLLMs is a pressing concern that requires proactive safeguards. As these models become more sophisticated, there is a risk they could be used to generate harmful or deceptive content across multiple modalities \cite{chen2024trustworthy}. To mitigate this risk, developers must implement advanced content filtering mechanisms, establish use case restrictions for high-risk domains, and develop techniques for watermarking and provenance tracking of MLLM-generated content. Creating comprehensive ethical use guidelines for both developers and end-users is also crucial in promoting responsible utilization of these powerful tools.

Ensuring fairness and equitable access to MLLM technology is vital to prevent the exacerbation of existing digital divides \cite{ray2023chatgpt}. This includes developing strategies to make MLLMs accessible across diverse languages and cultural contexts, as well as integrating features that make these systems usable for people with disabilities. Additionally, considering the environmental impact of MLLM development and deployment is crucial, with efforts needed to improve energy efficiency, track carbon footprints, and promote the use of renewable energy sources in AI infrastructure.

The governance and regulation of MLLMs require collaborative efforts between policymakers, industry leaders, and academics to develop appropriate frameworks that balance innovation with ethical considerations. This may involve establishing independent ethics committees to oversee MLLM research and development projects, as well as working towards global standards and agreements on the ethical development and use of these technologies \cite{rosenstrauch2023artificial}.

Finally, it is essential to consider the long-term societal impact of MLLMs. This includes funding interdisciplinary research on how these technologies affect cognition, social interactions, and cultural evolution. Developing educational programs to improve public understanding of MLLMs and their implications is crucial for fostering informed societal engagement with these technologies. Additionally, strategies must be developed to address potential workforce disruptions, including programs for reskilling and upskilling in response to the changing landscape of work in the age of advanced AI.

By thoroughly addressing these ethical considerations, we can work towards ensuring that the development and deployment of Multimodal Large Language Models contribute positively to society while minimizing potential risks and negative impacts. This requires ongoing vigilance, collaboration, and a commitment to ethical principles as these technologies continue to evolve and integrate into various aspects of our lives.

\section{Bias Mitigation Strategies}

One of the most pressing ethical concerns surrounding MLLMs is the presence of biases in both the training data and the resulting model outputs \cite{xu2024survey}. This issue is complex and multifaceted, requiring a comprehensive approach to address effectively. Let's explore this topic in more depth, examining the nature of these biases, their potential impacts, and strategies for mitigation.

Biases in MLLMs can manifest in various ways, often reflecting and amplifying existing societal prejudices. These biases may be related to race, gender, age, socioeconomic status, cultural background, or other demographic factors. For instance, an MLLM might generate images that reinforce gender stereotypes or produce text that uses racially insensitive language \cite{basta2022gender}. In multimodal systems, these biases can be particularly insidious as they may appear across different modalities, creating a compounded effect \cite{magesh2024hallucination}.

The consequences of biased MLLMs can be severe and far-reaching. When deployed in real-world applications, these systems can lead to discriminatory outcomes in areas such as hiring, lending, criminal justice, and healthcare. Moreover, biased MLLMs can shape public perception and reinforce harmful stereotypes, potentially exacerbating social inequalities.

To mitigate these biases, researchers and developers must employ a range of strategies:

\begin{itemize}
\item Diverse and representative data collection \cite{cegin2024effects}: Ensuring that training datasets include a wide range of perspectives, experiences, and demographic representations is crucial. This involves not only collecting data from diverse sources but also carefully curating and balancing the dataset to avoid overrepresentation of certain groups or viewpoints.

\item Bias detection and measurement \cite{lin2024investigating}: Developing robust techniques to identify and quantify biases in both training data and model outputs is essential. This may involve creating specialized test sets designed to probe for specific types of biases across different modalities.

\item Algorithmic debiasing \cite{owens2024multi}: Implementing techniques to reduce biases during the training process, such as adversarial debiasing or reweighting examples, can help mitigate the problem at its source.

\item Regular auditing and monitoring \cite{patil2024review}: Conducting ongoing assessments of MLLM outputs to detect emerging biases or unintended consequences is crucial, especially as these models are deployed in diverse real-world contexts.

\item Interdisciplinary collaboration \cite{jiao2024navigating}: Engaging experts from fields such as sociology, psychology, and ethics can provide valuable insights into the nature and impact of biases, as well as strategies for mitigation.

\end{itemize}

\textbf{Bias Mitigation Strategies} 
\begin{figure}
    \centering
    \includegraphics[width=0.7\linewidth]{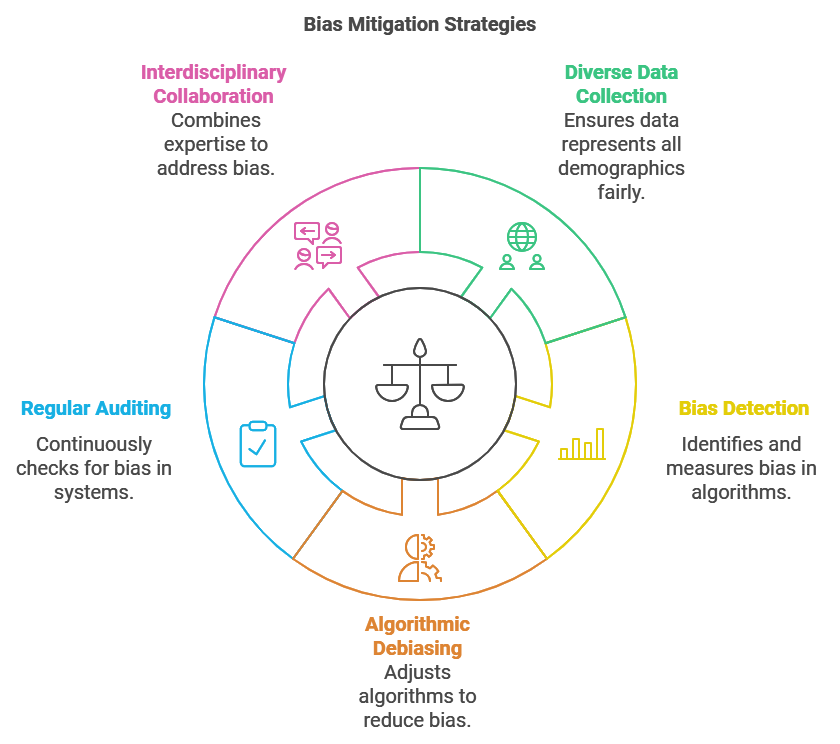}
    \caption{Bias Mitigation Strategies}
    \label{fig:Fine-tuning}
\end{figure}

It's important to note that bias mitigation in MLLMs is an ongoing process rather than a one-time fix. As these systems continue to evolve and be applied in new contexts, vigilance and continuous improvement in bias detection and mitigation strategies will be necessary.

Moreover, addressing biases in MLLMs also raises broader questions about the role of AI in society and the ethical responsibilities of those developing and deploying these technologies. It underscores the need for ongoing dialogue between technologists, policymakers, and the public to ensure that the development of MLLMs aligns with societal values and promotes fairness and equality.:

\subsection{Identifying and Measuring Bias}

The first step in addressing bias is to identify and quantify its presence in both the training data and the outputs generated by the model. This process typically involves analyzing the data for biased patterns or imbalances that may relate to sensitive attributes, such as race, gender, age, sexual orientation, or cultural background. These biases can arise from the ways data is collected, labeled, or represented, often reflecting real-world societal inequalities \cite{poulain2024bias}.

To assess these biases, researchers and developers use various fairness metrics. One common metric is demographic parity, which checks whether outcomes are distributed equally across different demographic groups. Another key metric is equalized odds, which assesses whether the model's error rates (false positives and false negatives) are consistent across groups. Other measures, such as disparate impact, calibration within groups, and individual fairness, offer additional perspectives for identifying and quantifying bias \cite{chen2023ai}.

These fairness metrics are crucial for revealing disparities in the model’s performance across different population segments. For example, a model might show better accuracy for one gender over another or make more errors in predicting outcomes for certain racial groups. Identifying these disparities allows researchers to apply targeted interventions, such as rebalancing the training data, applying algorithmic adjustments (e.g., bias mitigation techniques), or using post-processing corrections to improve fairness. In short, the goal is to ensure that machine learning models are both effective and equitable in their applications, serving all users fairly and minimizing harmful biases \cite{mehrabi2021survey}.

\subsection{Bias Mitigation Techniques}

Once biases have been identified, several techniques can be employed to mitigate their impact \cite{tripathi2024insaaf, lee2024life}:

\begin{itemize}
    \item \textbf{Adversarial Debiasing}: This approach involves training the MLLM with an adversarial objective, where a separate model attempts to predict sensitive attributes from the main model's outputs. By penalizing the main model for allowing the adversary to make accurate predictions, the MLLM is encouraged to learn more fair and unbiased representations.
    
    \item \textbf{Data Augmentation}: Increasing the representation of underrepresented groups in the training data can help reduce bias. Techniques such as oversampling minority classes, generating synthetic examples, or re-weighting instances can create a more balanced dataset, ensuring that the MLLM is exposed to a diverse range of perspectives and experiences.
    
    \item \textbf{Post-processing Techniques}: After the MLLM has been trained, post-processing methods can be applied to adjust its outputs and ensure fairness across different demographic groups. For example, calibration techniques can be used to equalize the model's performance across sensitive attributes, while threshold optimization can help balance the trade-off between fairness and accuracy.
\end{itemize}

\subsection{Challenges and Considerations}

While significant progress has been made in developing techniques to mitigate biases in machine learning models, several obstacles persist. One such challenge is the inherent tension between achieving complete fairness across all demographic groups and maintaining optimal model performance. In certain cases, eliminating bias completely may lead to a reduction in the model's overall accuracy or predictive power. Therefore, developers and researchers must carefully weigh the trade-offs between fairness and performance, considering the specific context and application of the model.

Furthermore, the complexity of multimodal data, which encompasses both textual and visual information, presents an additional hurdle in identifying and mitigating biases. Biases can manifest in subtle ways within both modalities, making their detection and rectification a complex task. For example, a model might exhibit bias in its interpretation of visual cues or in its understanding of language nuances, potentially leading to discriminatory or unfair outcomes. Addressing biases in multimodal data requires a comprehensive approach that considers the interplay between different modalities and employs sophisticated techniques to ensure fairness across all dimensions.

\section{Privacy and Data Protection}

Massive Language and Learning Models require extensive datasets to achieve their full potential, and these datasets often contain a wealth of sensitive personal information. This can range from personal images and medical records to social media posts, financial details, and location data. The inclusion of such data creates significant concerns about privacy and data protection, making it crucial for those developing and deploying these models to adopt rigorous ethical practices.

One primary concern is the unintentional leakage of sensitive data. These models can sometimes memorize and reproduce parts of their training data, leading to the potential exposure of private information during interactions with users. For instance, a model trained on medical records might inadvertently generate content containing personal health information, violating confidentiality and privacy regulations \cite{brown2022does, yao2024survey, pan2020privacy}.

Another major challenge lies in obtaining proper consent for data usage. In many instances, the data used to train these models might be scraped from public sources without the explicit consent of the individuals involved. This creates ethical dilemmas, especially when individuals are unaware that their data is being used to train AI systems. Ensuring transparent data collection practices and obtaining informed consent is vital for maintaining trust and complying with data protection regulations \cite{weidinger2021ethical,brown2022does,zhang2024right,weidinger2022taxonomy}.

The ethical principle of data minimization encourages developers to collect and store only the data strictly necessary for their specific task. By limiting the amount of sensitive information incorporated into these models, developers reduce the potential for harm and align their practices with privacy regulations emphasizing the necessity and proportionality of data collection. The ethical responsibility of protecting user privacy in the development and deployment of these models is complex and multifaceted. It involves navigating challenges related to data consent, preventing data leaks, anonymizing sensitive information, and adhering to strict regulations. Developers and deployers must prioritize privacy at every stage of the model lifecycle to ensure that the sensitive information of individuals is respected and protected. Only through the integration of strong privacy safeguards into the very fabric of their development can we guarantee that these powerful technologies are used ethically and responsibly \cite{sanderson2023ai,phattanaviroj2024data,kibriya2024privacy}.

\subsection{Privacy-Preserving Techniques}

To safeguard user privacy, several techniques can be employed in the MLLM development process:

\begin{itemize}
    \item \textbf{Differential Privacy}: By introducing carefully calibrated noise into the training data or the model's outputs, differential privacy helps prevent the leakage of sensitive information about individual data points. This allows MLLMs to learn useful patterns from the data while providing strong privacy guarantees \cite{singh2024whispered,charles2024fine}.
    
    \item \textbf{Federated Learning}: Instead of centralizing all training data in a single location, federated learning enables MLLMs to be trained collaboratively across multiple decentralized devices or institutions. Each participant keeps their raw data locally, only sharing model updates with the central server. This approach is particularly valuable in domains such as healthcare, where data sharing is restricted by privacy regulations \cite{kuang2024federatedscope}.
    
    \item \textbf{Data Minimization and Anonymization}: Collecting and retaining only the minimum amount of data necessary for the specific task at hand reduces the risk of privacy breaches. Additionally, techniques such as data anonymization, where personally identifiable information is removed or obfuscated, can help protect user privacy while still allowing MLLMs to learn from the data \cite{wiest2024anonymizing, li2024llm}.
\end{itemize}


\begin{figure}
    \centering
    \includegraphics[width=0.7\linewidth]{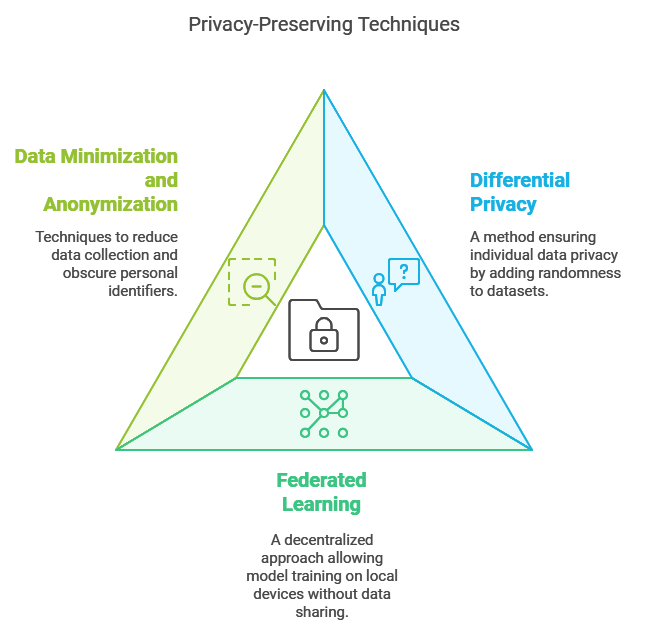}
    \caption{Privacy-Preserving Techniques}
    \label{Privacy-Preserving}
\end{figure}

\section{Conclusion}
In conclusion, the rapid advancement of Multimodal Large Language Models (MLLMs) heralds a new era of possibilities across various domains, from creative arts to scientific research and beyond. However, this progress also brings forth a complex web of ethical considerations that we must address with utmost care and responsibility.

Mitigating biases, protecting privacy, preventing misuse, ensuring transparency, and upholding accountability are all critical pillars in the responsible development and deployment of MLLMs. By integrating these ethical considerations into every stage of the AI lifecycle, we can navigate the challenges and complexities of this emerging technology landscape.

As MLLMs continue to evolve, ongoing collaboration between researchers, developers, policymakers, and the public will be essential to ensure these powerful tools are used for the betterment of society. By proactively addressing ethical concerns, fostering transparency, and upholding principles of fairness and accountability, we can harness the potential of MLLMs to create a future where AI serves as a force for good, empowering individuals, communities, and societies across the globe.

\chapter{Conclusion}

As we conclude our exploration of Multimodal Large Language Models (MLLMs) and their transformative impact on the field of artificial intelligence, it is crucial to reflect on the advancements they have enabled, the potential societal implications they bring forth, and the responsibility we bear in ensuring their ethical development and deployment.

\section{Recap of MLLMs' Impact on AI Research and Applications}

Multimodal Large Language Models (MLLMs) have profoundly transformed the landscape of AI research and applications, ushering in a new era of capabilities that span multiple modalities. This section provides a comprehensive overview of their impact, highlighting key advancements, applications, and critical considerations.

\subsection{Advancements in AI Capabilities}

MLLMs have significantly expanded the horizons of AI systems by enabling them to process and understand multiple modalities simultaneously. This breakthrough has led to remarkable progress in various tasks:

\begin{itemize}
    \item \textbf{Visual Question Answering (VQA):} MLLMs have revolutionized VQA tasks by:
        \begin{itemize}
            \item Interpreting complex visual scenes and providing accurate responses to natural language queries about image content
            \item Leveraging attention mechanisms and dynamic embeddings to focus on relevant image parts, improving performance by up to 9\% on standard datasets
            \item Excelling in cross-modal reasoning, enabling the modeling of relationships between objects in images and words in questions
            \item Utilizing graph-based approaches to represent and reason about scene structures, achieving high accuracies (e.g., 96.3\% on the GQA dataset)
            \item Addressing challenges such as complex reasoning, context understanding, and handling of imperfect inputs (e.g., blurry images)
            \item Integrating speech recognition and text-to-speech capabilities to enhance accessibility, particularly for visually impaired users
        \end{itemize}
    \item \textbf{Image Captioning:} These models generate detailed, context-aware descriptions of images, bridging the gap between visual perception and linguistic expression.
    \item \textbf{Cross-Modal Retrieval:} MLLMs excel at finding relevant images based on text queries and vice versa, enhancing search capabilities across modalities.
    \item \textbf{Multimodal Translation:} The ability to translate between different modalities, such as converting text to images or describing videos in textual form, has opened up new possibilities for content creation and accessibility.
\end{itemize}

\subsection{Unified Representations and Transfer Learning}

A key innovation in MLLM development has been the creation of unified representations for multimodal data. This advancement allows MLLMs to:

\begin{itemize}
    \item Align and understand relationships between various types of content
    \item Transfer knowledge across modalities, enhancing generalization capabilities
    \item Perform zero-shot and few-shot learning on new tasks with minimal additional training
\end{itemize}

Models such as CLIP, DALL-E, and GPT-4 with vision capabilities have demonstrated remarkable versatility and scalability. Their ability to generalize to new tasks with minimal retraining has made them invaluable tools in both research and practical applications.

\subsection{Applications Across Diverse Domains}

The impact of MLLMs extends across numerous fields, revolutionizing processes and enabling new possibilities:

\subsubsection{Creative Industries and Content Creation}
\begin{itemize}
    \item \textbf{Art Generation:} Tools like Midjourney and DALL-E 2 allow users to create unique artwork from text descriptions, democratizing artistic creation.
    \item \textbf{Video Production:} MLLMs assist in various stages of video production, from generating storyboards to creating short video clips based on text prompts.
    \item \textbf{Music Composition:} Models like MusicLM can generate original music based on text descriptions or even hummed melodies, expanding the boundaries of musical creativity.
\end{itemize}

\subsubsection{Healthcare}
\begin{itemize}
    \item \textbf{Medical Imaging Analysis:} MLLMs enhance diagnostic capabilities by interpreting X-rays, MRIs, and CT scans, potentially improving accuracy and efficiency in healthcare.
    \item \textbf{Drug Discovery:} These models analyze molecular structures and predict potential drug candidates, accelerating the pharmaceutical research process.
    \item \textbf{Personalized Treatment Plans:} By integrating patient data from multiple sources, MLLMs can suggest tailored treatment options, moving towards more personalized medicine.
\end{itemize}

\subsubsection{E-commerce and Retail}
\begin{itemize}
    \item \textbf{Visual Search:} Customers can find products by uploading images, enhancing the shopping experience and product discovery.
    \item \textbf{Virtual Try-On:} AR-powered systems allow users to virtually try on clothes or visualize furniture in their homes, reducing return rates and improving customer satisfaction.
    \item \textbf{Personalized Recommendations:} MLLMs analyze user behavior across text and visual data to provide more accurate and contextually relevant product suggestions.
\end{itemize}

\subsubsection{Autonomous Systems and Robotics}
\begin{itemize}
    \item \textbf{Self-Driving Cars:} MLLMs help vehicles understand their environment by integrating visual, textual (road signs), and sensor data, enhancing safety and navigation capabilities.
    \item \textbf{Robotic Manipulation:} Robots equipped with MLLMs can better understand and interact with their surroundings using multimodal inputs, improving dexterity and adaptability.
    \item \textbf{Drone Navigation:} MLLMs enable drones to navigate complex environments using visual and sensor data, expanding their potential applications in various industries.
\end{itemize}

\subsection{Critical Considerations}

While the advancements brought by MLLMs are significant, it is crucial to approach their development and deployment with a critical perspective:

\begin{itemize}
    \item \textbf{Ethical Concerns:} The potential for bias in MLLMs, particularly in sensitive applications like healthcare and autonomous systems, necessitates ongoing research into fairness and bias mitigation techniques.
    \item \textbf{Computational Resources:} The immense computational power required to train and run MLLMs raises questions about environmental impact and accessibility.
    \item \textbf{Data Privacy:} The vast amounts of multimodal data used to train these models present significant privacy concerns that must be addressed.
    \item \textbf{Interpretability:} As MLLMs become more complex, ensuring their decision-making processes are interpretable and explainable becomes increasingly challenging yet crucial for trust and accountability.
    \item \textbf{Potential for Misuse:} The ability of MLLMs to generate realistic content across modalities raises concerns about deepfakes and misinformation, requiring robust safeguards and detection methods.
\end{itemize}

In conclusion, while MLLMs have undeniably revolutionized AI research and applications, their development and deployment must be approached with careful consideration of both their potential benefits and risks. As these models continue to evolve, ongoing research, ethical guidelines, and regulatory frameworks will be essential to ensure their responsible and beneficial integration into society.

\section{Potential Societal Implications}

While MLLMs offer immense potential benefits, they also raise important societal questions that demand careful consideration. Ethical concerns surrounding bias and fairness, data privacy, and job displacement must be addressed to ensure that these technologies do not perpetuate inequalities or cause unintended harm.

The ability of MLLMs to generate realistic content also poses risks for the creation and spread of disinformation and deepfakes, which could be weaponized to manipulate public opinion or defame individuals. Additionally, the potential misuse of MLLMs in autonomous systems, such as drones or surveillance technologies, raises serious security and privacy concerns that require international regulation and oversight.

However, MLLMs also have the potential to bring about positive societal impacts. They can be transformative in fields like healthcare and education, assisting in medical diagnosis, personalized learning, and cultural preservation. Multilingual and cross-cultural MLLMs offer the possibility of promoting lesser-known languages and cultures, providing tools for digital communication and education in underrepresented communities.

\section{Call to Action for Responsible Development and Use}

As we stand at the precipice of an AI-driven future, it is imperative that we commit to the responsible development and deployment of MLLMs. This requires a concerted effort from researchers, industry leaders, policymakers, and the public to ensure that these technologies are created and used in an ethical, transparent, and accountable manner.

Developers and organizations must prioritize bias mitigation by actively identifying and addressing biases in MLLMs through diverse training datasets, fairness metrics, and adversarial debiasing techniques. Transparency in model development, including clear documentation of training data, model architectures, and decision-making processes, is essential for building trust and accountability.

Collaboration between industry and academia is crucial for advancing the capabilities of MLLMs while ensuring their responsible development. Engaging with the public to educate them about the risks and benefits of these technologies will foster trust and ensure that MLLMs serve the interests of society as a whole.

As we move forward, it is essential to integrate ethical considerations into every stage of AI development, from dataset creation to model deployment and monitoring. By designing MLLMs with ethics in mind and considering their environmental impact, we can work towards building a sustainable future for AI that benefits all of humanity.

The journey ahead is filled with both promise and challenges. It is up to us to navigate this path with wisdom, foresight, and a steadfast commitment to the responsible development and use of Multimodal Large Language Models. By doing so, we can unlock their transformative potential while ensuring that they serve as a force for good in our rapidly evolving world.

\bibliography{references}

\bibliographystyle{unsrt}

\end{document}